%% file: main.tex
\documentclass{article} \usepackage{iclr2019_conference,times}

\definecolor{mydarkblue}{rgb}{0,0.08,0.45}
\usepackage[colorlinks=true,
    linkcolor=mydarkblue,
    citecolor=mydarkblue,
    filecolor=mydarkblue,
    urlcolor=mydarkblue]{hyperref}

\usepackage[utf8]{inputenc} \usepackage[T1]{fontenc}    \usepackage{url}            \usepackage{booktabs}       \usepackage{amsfonts,amsthm}       \usepackage{nicefrac}       \usepackage{microtype}      \usepackage{graphicx}
\usepackage{blindtext}
\usepackage{xcolor, color, colortbl}
\usepackage{commath}
\usepackage{mdframed}
\usepackage{enumerate}
\usepackage{bold-extra}
\usepackage{nccmath}
\usepackage{caption}
\usepackage{subcaption}
\usepackage{dsfont}
\usepackage{amsmath, amsthm, amssymb}
\usepackage{enumitem}
\usepackage{booktabs}
\usepackage{multirow}
\usepackage{rotating}
\usepackage{mathtools}
\usepackage{empheq}
\usepackage{bm}
\usepackage{algorithm,algorithmic}
\usepackage{wrapfig}
\input{math_commands.tex}
\input{defs.tex}

\title{A Variational Inequality Perspective on \\ Generative Adversarial Networks}

\author{Gauthier Gidel$^{1,}$\thanks{Equal contribution, correspondence to \texttt{firstname.lastname@umontreal.ca}.}
\And
Hugo Berard$^{1,3,*}$
\And
Gaëtan Vignoud$^1$
\And
Pascal Vincent$^{1,2,3}$
\And
Simon Lacoste-Julien$^{1,2}$
\And \\[-8mm]
$^1$Mila \& DIRO, University of Montreal \,   $^2$Canada CIFAR AI Chair\\
$^3$Facebook Artificial Intelligence Research
}

\iclrfinalcopy \begin{document}

\maketitle

\begin{abstract}
Generative adversarial networks (GANs) form a generative modeling approach known for producing appealing samples, but they are notably difficult to train.
One common way to tackle this issue has been to propose new formulations of the GAN objective. Yet, surprisingly few studies have looked at optimization methods designed for this adversarial training.
In this work, we cast GAN optimization problems in the general variational inequality framework.
Tapping into the mathematical programming literature, we counter some
common misconceptions about the difficulties of saddle point optimization
and propose to extend techniques designed for variational
inequalities to the training of GANs. We apply \emph{averaging}, \emph{extrapolation} and a computationally cheaper variant that we call \emph{extrapolation from the past} to the stochastic gradient method (SGD) and Adam. \end{abstract}

\section{Introduction} \vspace{-1mm}
\label{sec:introduction}
Generative adversarial networks (GANs) \citep{goodfellow2014generative} form a generative modeling approach known for producing realistic natural images~\citep{karras2017progressive} as well as high quality super-resolution~\citep{ledig2017photo} and style transfer~\citep{zhu2017unpaired}. Nevertheless, GANs are also known to be difficult to train, often displaying an unstable behavior~\citep{goodfellow2016nips}. Much recent work has tried to tackle these training difficulties, usually by proposing new formulations of the GAN objective~\citep{nowozin_f-gan:_2016,arjovsky2017wasserstein}. Each of these formulations can be understood as a two-player game, in the sense of game theory~\citep{von1944theory},
and can be addressed as a variational inequality problem (VIP)~\citep{harker1990finite}, a framework that encompasses traditional saddle point optimization algorithms~\citep{korpelevich1976extragradient}.

Solving such GAN games is traditionally approached by running variants of stochastic gradient descent (SGD) initially developed for optimizing supervised neural network objectives. Yet it is known that for some games~\citep[\S8.2]{goodfellow2016nips} SGD exhibits oscillatory behavior and fails to converge.
This oscillatory behavior,  which does not arise from stochasticity, highlights a fundamental problem: while a direct application of basic gradient descent is an appropriate method for regular minimization problems, it is \emph{not} a sound optimization algorithm for the kind of two-player games of GANs. This constitutes a fundamental issue for GAN training, and calls for the use of more principled methods with more reassuring convergence guarantees.

\paragraph{Contributions.}
We point out that multi-player games can be cast as \emph{variational inequality problems} (VIPs) and consequently the same applies to any GAN formulation posed as a minimax or non-zero-sum game.
We present two techniques from this literature, namely \emph{averaging} and \emph{extrapolation}, widely used to solve VIPs but which have not been explored in the context of GANs before.\footnote{The preprints for \citep{mertikopoulos2018mirror} and~\citep{yazici2018unusual}, which respectively explored extrapolation and averaging for GANs, appeared after our initial preprint. See also the related work section \S\ref{sec:related_work}.}

We extend standard GAN training methods such as SGD or Adam into variants that incorporate these techniques (Alg.~\ref{alg:Extra-Adam} is new).
We also explain that the oscillations of basic SGD for GAN training previously noticed ~\citep{goodfellow2016nips} can be explained by standard variational inequality optimization results and we illustrate how \emph{averaging} and \emph{extrapolation} can fix this issue.

We study a variant of extragradient that we call \emph{extrapolation from the past} originally introduced by~\citet{popov1980modification}. It only requires one gradient computation per update compared to \emph{extrapolation}, which needs to compute the gradient twice.
We \emph{prove its convergence} for strongly monotone operators and in the stochastic VIP setting.

Finally, we test these techniques in the context of GAN training. We observe a $4$-$6\%$ improvement over~\citet{miyato2018spectral} on the inception score and the Fréchet inception distance on the CIFAR-10 dataset using a WGAN-GP~\citep{gulrajani2017improved} and a ResNet generator.\footnote{Code available at \url{https://gauthiergidel.github.io/projects/vip-gan.html}.}

\paragraph{Outline.}  	\label {par:outline_}
 	\S\ref{sec:background} presents the background on GAN and optimization, and shows how to cast this optimization as a VIP. \S\ref{sec:optim} presents standard techniques and \emph{extrapolation from the past} to optimize variational inequalities in a batch setting.
\S\ref{sec:gradient_method} considers these methods in the \emph{stochastic} setting, yielding three corresponding variants of SGD, and provides their respective convergence rates.
\S\ref{sec:combining} develops how to combine these techniques with already existing algorithms.
\S\ref{sec:related_work} discusses the related work and \S\ref{sec:experiments} presents experimental results.

\section{GAN optimization as a variational inequality problem} \label{sec:background}
\subsection{GAN formulations} \label{sub:gan_formulation}
The purpose of generative modeling is to generate samples from a distribution $q_\vtheta$ that matches best the true distribution $p$ of the data. The generative adversarial network training strategy can be understood as a \emph{game} between two players called \emph{generator} and \emph{discriminator}. The former produces a sample that the latter has to classify between real or fake data. The final goal is to build a generator able to produce sufficiently realistic samples to fool the discriminator.

In the original GAN paper~\citep{goodfellow2014generative}, the GAN objective is formulated as a \emph{zero-sum game} where the cost function of the discriminator $D_\vphi$ is given by the negative log-likelihood of the binary classification task between real or fake data generated from $q_\vtheta$ by the generator,
\begin{equation}\label{eq:minimax_gan}
  \min_{\vtheta} \max_{\vphi} \LL(\vtheta,\vphi)
  \quad \text{where} \quad
    \LL(\vtheta,\vphi) \defas - \!\!\underset{\rvx \sim p}{\mathbb E} \!\![ \log D_\vphi(\rvx)] -  \!\!\underset{\rvx' \sim q_\vtheta}{\mathbb E} \!\![\log (1- D_\vphi(\rvx'))] \,.
\end{equation}
However~\citet{goodfellow2014generative} recommends to use in practice a second formulation, called \emph{non-saturating GAN}. This formulation is a \emph{non-zero-sum game} where the aim is to jointly minimize:
\begin{equation}\label{eq:non_saturating_objective}
  \LL_G(\vtheta,\vphi) \defas - \!\!\underset{\rvx' \sim q_{\vtheta}}{\mathbb E} \!\! \log D_\vphi(\rvx')
  \;\; \text{and} \;\;
   \LL_D(\vtheta,\vphi) \defas - \!\underset{\rvx \sim p}{\mathbb E} \log D_\vphi(\rvx) \,-  \!\!\underset{\rvx' \sim q_\vtheta}{\mathbb E} \!\!\log (1- D_\vphi(\rvx')) \,.
\end{equation}
The dynamics of this formulation has the same \emph{stationary points} as the zero-sum one~\eqref{eq:minimax_gan} but is claimed to provide ``much stronger gradients early in learning''~\citep{goodfellow2014generative} .

\subsection{Equilibrium} \label{sub:equilibrium}
The minimax formulation~\eqref{eq:minimax_gan} is theoretically convenient because a large literature on games studies this problem and provides guarantees on the existence of equilibria. Nevertheless, practical considerations lead the GAN literature to consider a different objective for each player as formulated in~\eqref{eq:non_saturating_objective}. In that case, the \emph{two-player game problem}~\citep{von1944theory} consists in finding the following \emph{Nash equilibrium}:
\begin{equation} \label{eq:two_player_games}
  \vtheta^* \in \argmin_{\vtheta \in \Theta}\LL_G(\vtheta,\vphi^*)
  \quad \text{and} \quad
  \vphi^* \in \argmin_{\vphi \in \Phi} \LL_D(\vtheta^*,\vphi) \,.
\end{equation}
Only when $\LL_G = - \LL_D$ is the game called a \emph{zero-sum game} and~\eqref{eq:two_player_games} can be formulated as a minimax problem.
One important point to notice is that the two optimization problems in~\eqref{eq:two_player_games} are \emph{coupled} and have to be considered \emph{jointly} from an optimization point of view.

Standard GAN objectives are non-convex (i.e. each cost function is non-convex), and thus such (pure) equilibria may not exist. As far as we know, not much is known about the existence of these equilibria for non-convex losses (see~\citet{heusel2017gans} and references therein for some results). In our theoretical analysis in \S\ref{sec:gradient_method}, our assumptions (monotonicity~\eqref{eq:monotonicity} of the operator and convexity of the constraint set) imply the existence of an equilibrium.

In this paper, we focus on ways to optimize these games, assuming that an equilibrium exists.
As is often standard in non-convex optimization, we also focus on finding points satisfying the necessary \emph{stationary conditions}. As we mentioned previously, one difficulty that emerges in the optimization of such games is that the two different cost functions of~\eqref{eq:two_player_games} have to be minimized jointly in $\vtheta$ and~$\vphi$.
Fortunately, the optimization literature has for a long time studied so-called \emph{variational inequality problems}, which generalize the stationary conditions for two-player game problems.

\subsection{Variational inequality problem formulation} \label{sub:variational_inequality}

We first consider the local necessary conditions that characterize the solution of the \emph{smooth} two-player game~\eqref{eq:two_player_games}, defining \emph{stationary points}, which will motivate the definition of a variational inequality.
In the unconstrained setting, a \emph{stationary point} is a couple $(\vtheta^*,\vphi^*)$ with zero gradient:
\begin{equation}
  \|\nabla_\vtheta \LL_G(\vtheta^*,\vphi^*)\| = \| \nabla_\vphi \LL_D(\vtheta^*,\vphi^*) \| = 0 \,.
\end{equation}
When constraints are present,\footnote{An example of constraint for GANs is to clip the parameters of the discriminator~\citep{arjovsky2017wasserstein}.} a \emph{stationary point} $(\vtheta^*,\vphi^*)$ is such that the directional derivative of each cost function is non-negative in any feasible direction (i.e. there is no feasible descent direction):
\begin{equation} \label{eq:SP_stationnary_conditions}
\nabla_\vtheta \LL_G(\vtheta^*,\vphi^*)^\top \! (\vtheta - \vtheta^*) \geq 0
\quad \text{and} \quad
\nabla_\vphi \LL_D(\vtheta^*,\vphi^*)^\top \!(\vphi - \vphi^*) \geq 0 \,,
\; \; \forall \,(\vtheta,\vphi) \in \Theta \times \Phi.
\end{equation}
Defining $\vomega \defas (\vtheta,\vphi),\,\vomega^* \defas (\vtheta^*,\vphi^*) ,\,\Omega \defas \Theta\times \Phi$, Eq.~\eqref{eq:SP_stationnary_conditions} can be compactly formulated as:
\begin{equation}\label{eq:VI_stationary_cond}
   F(\vomega^*)^\top (\vomega - \vomega^*) \geq 0 \, , \; \;\forall \vomega \in \Omega
   \quad \text{where} \quad
F(\vomega) \defas \begin{bmatrix}
\nabla_\vtheta  \LL_G(\vtheta,\vphi) &
\nabla_\vphi  \LL_D(\vtheta,\vphi)
\end{bmatrix}^\top \,.
\end{equation}
These stationary conditions can be generalized to any continuous vector field: let $\Omega \subset \RR^d$ and $F : \Omega \to \RR^d$ be a continuous mapping. The \emph{variational inequality problem}~\citep{harker1990finite} (depending on $F$ and $\Omega$) is:
\begin{equation}\label{eq:VI_problem_weak} \tag{VIP}
  \text{find} \; \vomega^* \in \Omega \quad \text{such that} \quad F(\vomega^*)^\top (\vomega - \vomega^*) \geq 0 \, , \; \;\forall \vomega \in \Omega\,.
\end{equation}
We call \emph{optimal set} the set $\Omega^*$ of $\vomega \in \Omega$ verifying~\eqref{eq:VI_problem_weak}. The intuition behind it is that any $\vomega^* \in \Omega^*$ is a \emph{fixed point} of the \emph{constrained} dynamic of $F$ (constrained to $\Omega$).

We have thus showed that both saddle point optimization and non-zero sum game optimization, which encompass the large majority of GAN variants proposed in the literature, can be cast as VIPs.
In the next section, we turn to suitable optimization techniques for such problems.

\section{Optimization of Variational Inequalities (batch setting)} \label{sec:optim}

\begin{figure}
\vspace*{-1cm}
\hspace*{-3mm}
\begin{minipage}{0.41\textwidth}
 \centering
 \includegraphics[width=\textwidth]{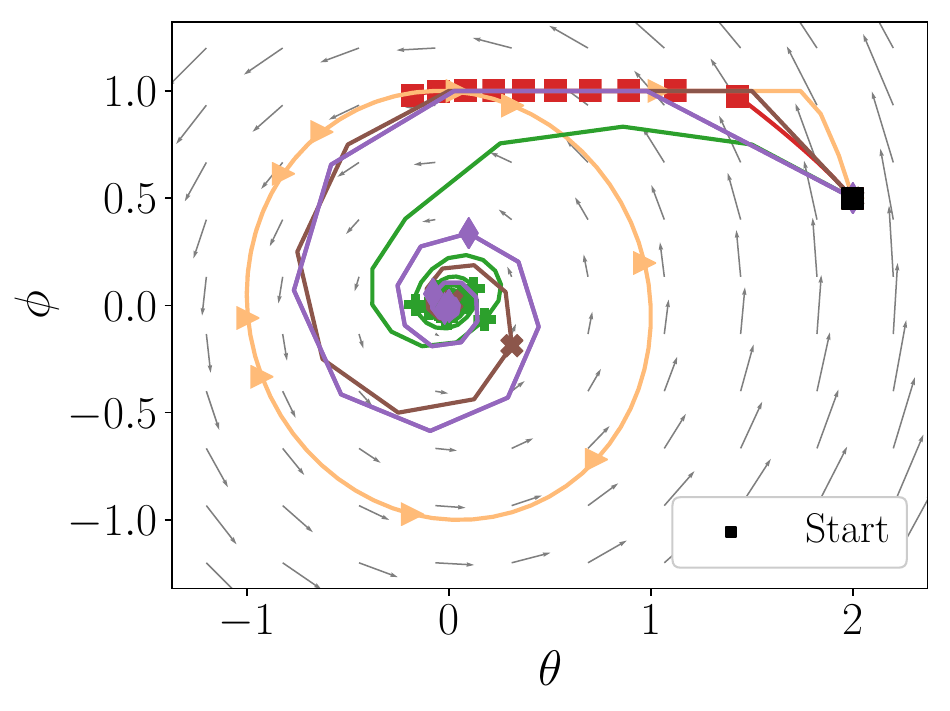}
\end{minipage}
\;
\begin{minipage}{0.57\textwidth}
 \centering
 \vspace{-3mm}
 \includegraphics[width=1.03\textwidth, height= .18\textwidth]{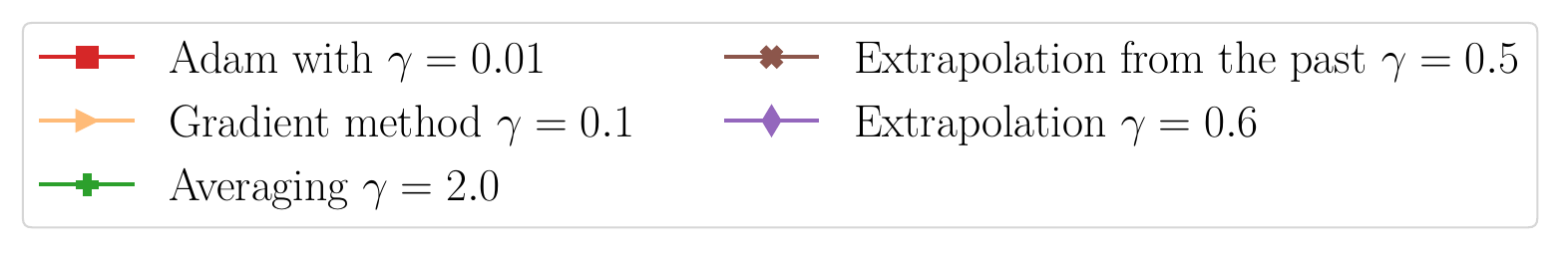}
\vspace{-6mm}
\captionof{figure}{
\small
Comparison of the basic gradient method (as well as Adam) with the techniques presented in \S\ref{sec:optim} on the optimization of~\eqref{eq:bilinear_wgan}. Only the algorithms advocated in this paper (Averaging, Extrapolation and Extrapolation from the past) converge quickly to the solution. Each marker represents 20 iterations. We compare these algorithms on a non-convex objective in \S\ref{sub:non_convex_gan}.
}
\label{fig:bilinear_toy}
\end{minipage}
\end{figure}

Let us begin by looking at techniques that were developed in the optimization literature to solve VIPs. We present the intuitions behind them as well as their performance on a simple bilinear problem (see Fig.~\ref{fig:bilinear_toy}). Our goal is to provide mathematical insights on \emph{averaging} (\S\ref{sub:averaging}) and \emph{extrapolation} (\S\ref{sub:extrapolation}) and propose a novel variant of the extrapolation technique that we called \emph{extrapolation from the past} (\S\ref{sub:extrapolation_from_the_past}).
We consider the batch setting, i.e., the operator
$F(\vomega)$ defined in Eq.~\ref{eq:VI_stationary_cond} yields an exact full gradient.
We present extensions of these techniques to the stochastic setting later in \S\ref{sec:gradient_method}.

The two standard methods studied in the VIP literature
are the \emph{gradient method}~\citep{bruck_weak_1977}
and the \emph{extragradient method}~\citep{korpelevich1976extragradient}.
The iterates of the basic gradient method are given by $\vomega_{t+1} = P_\Omega[\vomega_t - \eta F(\vomega_t)]$ where $P_\Omega[\cdot]$ is the \emph{projection onto the constraint set} (if constraints are present) associated to~\eqref{eq:VI_problem_weak}. These iterates are known to converge linearly under an additional assumption on the operator\footnote{Strong monotonicity, a generalization of strong convexity. See \S\ref{app:definitions}.}~\citep{chen1997convergence}, but oscillate for a bilinear operator as shown in Fig.~\ref{fig:bilinear_toy}.
On the other hand, the \emph{uniform average} of these iterates converge for any bounded monotone operator with a $O(1/\sqrt{t})$ rate~\citep{nedic_subgradient_2009}, motivating the presentation of \emph{averaging} in \S\ref{sub:averaging}.
By contrast, the \emph{extragradient method} (extrapolated gradient) does not require any averaging to converge for monotone operators (in the batch setting), and can even converge at the faster $O(1/t)$ rate~\citep{nesterov2007dual}. The idea of this method is to compute a lookahead step (see intuition on \emph{extrapolation} in \S\ref{sub:extrapolation})
in order to compute a more stable direction to follow.

\subsection{Averaging} \label{sub:averaging}

More generally, we consider a \emph{weighted averaging} scheme with weights $\rho_t\geq 0$. This \emph{weighted averaging} scheme have been proposed for the first time for (batch) VIP by~\citet{bruck_weak_1977},
\begin{equation}\label{eq:averaging_scheme}
 	\bar \vomega_T \defas \frac{\sum_{t=0}^{T-1} \rho_t \vomega_t}{S_T} \, ,
 	\quad S_T \defas \sum_{t=0}^{T-1} \rho_t \,.
\end{equation}
Averaging schemes can be efficiently implemented in an online fashion noticing that,
\begin{equation}\label{eq:online_averaging}
	\bar \vomega_T = (1 - \tilde \rho_T)  \bar \vomega_{T-1} + \tilde \rho_T \vomega_T
  \quad \text{where} \quad
  0 \leq \tilde \rho_T \leq 1 \,.
  \end{equation}
For instance, setting $\tilde \rho_T = \frac{1}{T}$ yields \emph{uniform averaging} ($\rho_t =1$) and $\tilde \rho_t = 1 - \beta <1$ yields \emph{geometric averaging}, also known as \emph{exponential moving averaging} ($\rho_t = \beta^{T-t} ,\, 1\leq t \leq T$).
Averaging is experimentally compared with the other techniques presented in this section in Fig.~\ref{fig:bilinear_toy}.

In order to illustrate how averaging tackles the oscillatory behavior in game optimization, we consider a toy example where the discriminator and the generator are linear: $D_\vphi(\rvx)= \vphi^T\rvx$ and $G_\vtheta(\rvz)=\vtheta \rvz$ (implicitly defining $q_\vtheta$). By substituting these expressions in the WGAN objective,\footnote{Wasserstein GAN (WGAN) proposed by~\citet{arjovsky2017wasserstein} boils down to the following minimax formulation:
$\min_{\vtheta \in \Theta} \max_{\vphi \in \Phi, ||D_\vphi||_L \leq 1} \EE_{\rvx \sim p}[D_\vphi(\rvx)] - \EE_{\rvx' \sim q_{\vtheta}}[D_\vphi(\rvx')]$. } we get the following bilinear objective:
\vspace{-1mm}
\begin{equation} \label{eq:bilinear_wgan}
 \min_{\vtheta \in \Theta} \max_{\vphi \in \Phi, ||\vphi|| \leq 1} \vphi^T\EE[\rvx] - \vphi^T\vtheta\EE[\rvz] \,.
\end{equation}
A similar task was presented by \citet{nagarajan_gradient_2017} where they consider a quadratic discriminator instead of a linear one, and show that gradient descent is not necessarily asymptotically stable.
The bilinear objective has been extensively used~\citep{goodfellow2016nips,mescheder2018convergence,yadav_stabilizing_2017,daskalakis2017training} to highlight the difficulties of gradient descent for saddle point optimization. Yet, ways to cope with this issue have been proposed decades ago in the context of mathematical programming.
For illustrating the properties of the methods of interest, we will study their behavior in the rest of \S\ref{sec:optim} on a simple \emph{unconstrained} unidimensional version of Eq.~\ref{eq:bilinear_wgan} (this behavior can be generalized to general multidimensional bilinear examples, see \S\ref{sub:generalization_to_general_unconstrained_bilinear_objective}):
\vspace*{-1mm}
\begin{equation}
	\label{eq:simple_bilienar}
	\min_{\theta \in \R} \max_{\phi \in \R} \; \;\theta \cdot \phi
	\qquad \text{and} \qquad
	(\theta^*,\phi^*) = (0,0) \,.
\end{equation}
The operator associated with this minimax game is $F(\theta,\phi) = (\phi, - \theta)$.
There are several ways to compute the discrete updates of this dynamics. The two most common ones are the \emph{simultaneous} and the \emph{alternating} gradient update rules,
\begin{equation}
\label{eq:update_rules_bilinear}
	\text{Simultaneous update:} \;\;
  \left\{\begin{aligned}
	\theta_{t+1} &=  \theta_t - \eta \phi_t \\
	\phi_{t+1} & = \phi_t +  \eta \theta_{t}
	\end{aligned} \right.
	\,,
  \quad
  \text{Alternating update:} \;\;
  \left\{\begin{aligned}
  \theta_{t+1} &=  \theta_t - \eta \phi_t \\
  \phi_{t+1} & = \phi_t +  \eta \theta_{t+1}
  \end{aligned} \right.
  .
\end{equation}
Interestingly, these two choices give rise to completely different behaviors. The norm of the \emph{simultaneous} updates diverges geometrically, whereas the alternating iterates are bounded but do not converge to the equilibrium. As a consequence, their respective uniform average have a different behavior, as highlighted in the following proposition (proof in \S\ref{sub:explicit_method} and generalization in \S\ref{sub:generalization_to_general_unconstrained_bilinear_objective}):
\newcommand{\propOne}{
  The \emph{simultaneous} iterates diverge geometrically and the \emph{alternating} iterates defined in~\eqref{eq:update_rules_bilinear} are bounded but do not converge to 0 as
\begin{equation}
      \emph{Simultaneous:} \;\; \theta_{t+1}^2+\phi_{t+1}^2 = (1+\eta^2) (\theta_t^2+\phi_t^2) \,,
      \quad
       \emph{Alternating:} \;\;\theta_{t}^2 + \phi_t^2 = \Theta (\theta_0^2 + \phi_0^2)
\end{equation}
where $u_t = \Theta(v_t) \Leftrightarrow \exists \alpha,\beta, t_0 >0 \text{ such that } \forall t \geq t_0,  \alpha v_t \leq u_t \leq \beta v_t$.

The uniform average $(\bar \theta_t,\bar \phi_t) \defas \frac{1}{t} \sum_{s=0}^{t-1}( \theta_{s},\phi_s)$ of the \emph{simultaneous} updates (resp. the \emph{alternating updates}) diverges (resp. converges to 0) as,
\begin{equation}
  \emph{Simultaneous:} \;\,  \bar \theta_t^2 + \bar \phi_t^2 = \Theta\left(\frac{\theta_0^2 + \phi_0^2}{\eta^2 t^2} (1+\eta^2)^t \right) \! ,
  \;\;
   \emph{Alternating:} \;\,
  \bar \theta_t^2 + \bar \phi_t^2 = \Theta \left(\frac{\theta_0^2 + \phi_0^2}{\eta^2 t^2}\right) .
\end{equation}
}
\begin{proposition}
\label{prop:explicit}
\propOne
\end{proposition}
This sublinear convergence result, proved in \S\ref{sec:explicit_implicit_and_extragradient_methods_on_unconstrained_bilinear_games}, underlines the benefits of averaging when the sequence of iterates is bounded (i.e. for \emph{alternating} update rule).
When the sequence of iterates is not bounded (i.e. for \emph{simultaneous} updates) averaging fails to ensure convergence. This theorem also shows how \emph{alternating} updates may have better convergence properties than \emph{simultaneous} updates.

\subsection{Extrapolation} \label{sub:extrapolation}

Another technique used in the variational inequality literature to prevent oscillations is \emph{extrapolation}.
This concept is anterior to the extragradient method since \citet{korpelevich1976extragradient} mentions that the idea of \emph{extrapolated} ``prices'' to give ``stability'' had been already formulated by \citet[Chap. II]{polyak1963gradient}. The idea behind this technique is to compute the gradient at an (extrapolated) point different from the current point from which the update is performed, stabilizing the dynamics:
\begin{align}
	\text{Compute extrapolated point:} \; \;&\vomega_{t+1/2} = P_\Omega[\vomega_t - \eta F(\vomega_t)] \, , \label{eq:extrapolation}   \\
	\text{Perform update step:} \quad  &\vomega_{t+1} = P_\Omega[\vomega_t - \eta F(\vomega_{t+1/2})] \label{eq:update_step_extrapolation} \,.
\end{align}
Note that, even in the \emph{unconstrained case}, this method is intrinsically different from Nesterov's momentum\footnote{\citet[\S7.2]{sutskever2013training} showed the equivalence between ``standard momentum'' and Nesterov's formulation.}~\citep[Eq.~2.2.9]{yurii1983introductory} because of this lookahead step for the gradient computation:
\begin{equation}
  \text{Nesterov's method:} \quad\vomega_{t+1/2} = \vomega_t - \eta F(\vomega_t) \,,
  \qquad
  \vomega_{t+1} = \vomega_{t+1/2} + \beta (\vomega_{t+1/2} - \vomega_t)  \,.
\end{equation}
Nesterov's method does not converge when trying to optimize~\eqref{eq:simple_bilienar}.
One intuition of why \emph{extrapolation} has better convergence properties than the standard gradient method comes from Euler's integration framework.
Indeed, to first order, we have $\vomega_{t+1/2} \approx \vomega_{t+1} + o(\eta)$ and consequently, the update step~\eqref{eq:update_step_extrapolation} can be interpreted as a first order approximation to an \emph{implicit method} step:
\begin{equation}\label{eq:implicit_step}
 	\text{Implicit step:} \quad \vomega_{t+1} = \vomega_t - \eta F(\vomega_{t+1})
 	\,.
 \end{equation}
\emph{Implicit methods} are known to be more stable and to benefit from better convergence properties~\citep{atkinson2003introduction} than \emph{explicit methods}, e.g., in \S\ref{app:extragradient_method} we show that~\eqref{eq:implicit_step} on~\eqref{eq:simple_bilienar} converges \emph{for any} $\eta$. Though, they are usually not practical since they require to solve a potentially non-linear system at each step.
Going back to the simplified WGAN toy example~\eqref{eq:simple_bilienar} from \S\ref{sub:averaging}, we get the following update rules:
\begin{equation}
\label{eq:update_implicit_extra}
	\text{Implicit:} \; \; \left\{\begin{aligned}
	\theta_{t+1} &=  \theta_t - \eta \phi_{t+1} \\
	\phi_{t+1} & = \phi_t +  \eta \theta_{t+1}
	\end{aligned} \right. \,,
	\qquad
	\text{Extrapolation:}
	\left\{\begin{aligned}
	\theta_{t+1} &=  \theta_t - \eta (\phi_{t} + \eta \theta_t) \\
	\phi_{t+1} & = \phi_t +  \eta (\theta_{t} - \eta \phi_t)
	\end{aligned} \right. \,.
\end{equation}
In the following proposition, we see that for $\eta<1$, the respective convergence rates of the \emph{implicit method} and \emph{extrapolation} are highly similar. Keeping in mind that the latter has the major advantage of being more practical, this proposition clearly underlines the benefits of \emph{extrapolation}. Note that Prop.~\ref{prop:explicit} and~\ref{prop:implicit_extra} generalize to general unconstrained bilinear game (more details and proof in  \S\ref{sub:generalization_to_general_unconstrained_bilinear_objective}),
\begin{proposition}\label{prop:implicit_extra}
The squared norm of the iterates $N_t^2 \defas \theta_t^2 + \phi_t^2$, where the update rule of $\theta_t$ and $\phi_t$ are defined in~\eqref{eq:update_implicit_extra}, decreases geometrically for any $\eta <1$ as,
\begin{equation}
\emph{Implicit:} \;\; N_{t+1}^2 = \big(1 -\eta^2 + \eta^4 + \mathcal O (\eta^6)\big)N_t^2 \, ,\quad\;
\emph{Extrapolation:} \;\; N_{t+1}^2 = (1-\eta^2+\eta^4)N_t^2 \,.
\end{equation}
\end{proposition}

\subsection{Extrapolation from the past} \label{sub:extrapolation_from_the_past}
One issue with extrapolation is that the algorithm ``wastes'' a gradient~\eqref{eq:extrapolation}. Indeed we need to compute the gradient at two different positions for every single update of the parameters.
\citep{popov1980modification} proposed a similar technique that only requires a single gradient computation per update. The idea is to store and re-use the extrapolated gradient for the extrapolation:
\begin{align}
  \text{Extrapolation from the past:} \; \;&\vomega_{t+1/2} = P_\Omega[\vomega_t - \eta F(\vomega_{t-1/2})] \quad \label{eq:extrapolation_past}  \\
  \text{Perform update step:} \quad  &\vomega_{t+1} = P_\Omega[\vomega_t - \eta F(\vomega_{t+1/2})]\;\; \text{and store:} \; \;F(\vomega_{t+1/2}) \label{eq:update_past}
\end{align}
A similar update scheme was proposed by \citet[Alg. 1]{chiang2012online} in the context of online convex optimization and generalized by~\citet{rakhlin2013online} for general online learning.
Without projection, \eqref{eq:extrapolation_past} and~\eqref{eq:update_past} reduce to the optimistic mirror descent described by~\citet{daskalakis2017training}:  
\vspace{-3mm}
\begin{equation}\label{eq:optimistic} 
\text{Optimistic mirror descent (OMD):}
\quad \vomega_{t+1/2} = \vomega_{t-1/2} - 2\eta F(\vomega_{t-1/2}) + \eta F(\vomega_{t-3/2})
\end{equation}
OMD was proposed with similar motivation as ours, namely tackling oscillations due to the game formulation in GAN training, but with an online learning perspective.
Using the VIP point of view, we are able to prove a linear convergence rate for \emph{extrapolation from the past} (see details and proof of Theorem~\ref{thm:linear_con} in \S\ref{app:extrapolation_from_the_past}). We also provide results on the averaged iterate for a stochastic version in \S\ref{sec:gradient_method}. 
In comparison to the convergence results from~\citet{daskalakis2017training} that hold for a bilinear objective, we provide a faster convergence rate (linear vs sublinear) on the last iterate for a general (strongly monotone) operator $F$ and any projection on a convex $\Omega$. One thing to notice is that the operator of a bilinear objective is \emph{not} strongly monotone, but in that case one can use the standard extrapolation method~\eqref{eq:extrapolation} which converges linearly for an unconstrained bilinear game~\citep[Cor.~3.3]{tseng1995linear}.
\newcommand{\theoremLinearConvPast}{
  If $F$ is $\mu$-strongly monotone (see \S\ref{app:definitions} for the definition of strong monotonicity) and $L$-Lipschitz, then the updates \eqref{eq:extrapolation_past} and~\eqref{eq:update_past} with $\eta = \frac{1}{4L}$ provide linearly converging iterates,
  \begin{equation}
    \|\vomega_{t}-\vomega^*\|_2^2
  \leq  \left( 1-  \frac{\mu}{4L} \right)^t\|\vomega_0-\vomega^*\|_2^2\, , \quad \forall t \geq 0
  \,.
\end{equation}
}
\begin{theorem}[Linear convergence of \emph{extrapolation from the past}]\label{thm:linear_con}
  \theoremLinearConvPast
\end{theorem}

\section{Optimization of VIP with stochastic gradients} \label{sec:gradient_method}

\begin{figure}
\vspace*{-5mm}
\begin{minipage}{0.31\linewidth}
\begin{algorithm}[H]
    \caption{AvgSGD}\label{alg:AvgSGD}
    \begin{algorithmic}
    	\STATE Let $\vomega_0 \in \Omega$\\[1mm]
      \FOR{$t=0 \ldots T-1$}
      \STATE $\xi_t \sim P$ \hfill \emph{(mini-batch)}
      \STATE $\bm{d}_t \leftarrow  F( \vomega_{t},\xi_{t})$ \\[1.5mm]
      \STATE $\vomega_{t+1} \leftarrow P_\Omega[\vomega_t - \eta_t \bm{d}_t]$ \\[1mm]
      \ENDFOR
      \STATE Return $\bar \vomega_{T} \leftarrow \frac{\sum_{t=0}^{T-1}\eta_t \vomega_{t}}{\sum_{t=0}^{T-1}\eta_t}$\\[1mm]
    \end{algorithmic}
  \end{algorithm}
\end{minipage}
\begin{minipage}{0.33\linewidth}
   \begin{algorithm}[H]
    \caption{AvgExtraSGD}\label{alg:AvgExtraSGD}
    \begin{algorithmic}
      \FOR{$t=0 \ldots T-1$}
                        	  \STATE $\xi_t,\xi_t' \sim P$ \hfill \emph{(mini-batches)}
      \STATE $\bm{d}_t \leftarrow  F( \vomega_{t},\xi_{t})$
      \STATE $\vomega'_{t} \leftarrow P_\Omega[\vomega_t - \eta_t \bm{d}_t]$
      \STATE $\bm{d}'_t \leftarrow  F( \vomega'_{t},\xi'_{t})$
            \STATE $\vomega_{t+1} \leftarrow P_\Omega[\vomega_t - \eta_t \bm{d}'_t]$       \label{algline:AvgExtraSGD_update_step}
      \ENDFOR
      \STATE Return $\bar \vomega_{T} \leftarrow \frac{\sum_{t=0}^{T-1}\eta_t \vomega'_{t}}{\sum_{t=0}^{T-1}\eta_t}$
    \end{algorithmic}
  \end{algorithm}
\end{minipage}
\begin{minipage}{0.34\linewidth}
  \begin{algorithm}[H]
   \caption{AvgPastExtraSGD}\label{alg:AvgPastExtraSGD}
   \begin{algorithmic}
      \STATE Let $\vomega_0 \in \Omega$ \\[-1mm]
     \FOR{$t=0 \ldots T-1$}
	 \STATE $\xi_t \sim P$ \hfill \emph{(mini-batch)}
                    \STATE $\vomega'_{t} \leftarrow P_\Omega[\vomega_t - \eta_t \bm{d}_{t-1}]$
     \STATE $\bm{d}_t \leftarrow  F( \vomega'_{t},\xi_{t})$
     \STATE $\vomega_{t+1} \leftarrow P_\Omega[\vomega_t - \eta_t \bm{d}_t]$
     \ENDFOR
     \STATE Return $\bar \vomega_{T} \leftarrow \frac{\sum_{t=0}^{T-1}\eta_t \vomega'_{t}}{\sum_{t=0}^{T-1}\eta_t}$
   \end{algorithmic}
 \end{algorithm}
\end{minipage}
\caption{Three variants of SGD computing $T$ updates, using the techniques introduced in \S\ref{sec:optim}.
\label{alg:SGDvariants}
}
\vspace*{-3mm}
\end{figure}
In this section, we consider extensions of the techniques presented in \S\ref{sec:optim} to the context of a \emph{stochastic} operator, i.e., we no longer have access to the exact gradient $F(\vomega)$ but to an unbiased \emph{stochastic} estimate of it, $F(\vomega,\xi)$, where $\xi\sim P$ and $F(\vomega) := \EE_{\xi \sim P}[F(\vomega,\xi)]$.
It is motivated by GAN training where we only have access to a finite sample estimate of the expected gradient, computed on a mini-batch.
For GANs, $\xi$ is a mini-batch of points coming from the true data distribution $p$ and the generator distribution $q_\vtheta$.

For our analysis, we require at least one of the two following assumptions on the stochastic operator:
\begin{assumption}\label{assum:var_bounded}
Bounded variance by $\sigma^2$: $\EE_\xi[\|F(\vomega)-F(\vomega,\xi)\|^2]\leq\sigma^2 \,, \;\; \forall \vomega \in \Omega\,.$
\end{assumption}
\begin{assumption}\label{assum:sub_bounded}
Bounded expected squared norm by $M^2$:
 $\EE_\xi[\|F(\vomega,\xi)\|^2]\leq M^2, \;\forall \vomega \in \Omega.$ \vspace*{-2mm}
\end{assumption}
Assump.~\ref{assum:var_bounded} is standard in stochastic variational analysis, while Assump.~\ref{assum:sub_bounded} is a stronger assumption sometimes made in stochastic convex optimization. To illustrate how strong Assump.~\ref{assum:sub_bounded} is, note that it does not hold for an unconstrained bilinear objective like in our example~\eqref{eq:simple_bilienar} in \S\ref{sec:optim}. It is thus mainly reasonable for bounded constraint sets.
Note that in practice we have $\sigma \ll M$.

We now present and analyze three algorithms that are variants of SGD that are appropriate to solve~\eqref{eq:VI_problem_weak}. The first one Alg.~\ref{alg:AvgSGD} (AvgSGD) is the stochastic extension of the gradient method for solving~\eqref{eq:VI_problem_weak}; Alg.~\ref{alg:AvgExtraSGD} (AvgExtraSGD) uses \emph{extrapolation} and Alg.~\ref{alg:AvgPastExtraSGD} (AvgPastExtraSGD) uses \emph{extrapolation from the past}. A fourth variant that re-use the mini-batch for the extrapolation step (ReExtraSGD, Alg.~\ref{alg:ReAvgExtraSGD}) is described in \S\ref{sec:another_extension_of_sem}. These four algorithms return an \emph{average} of the iterates (typical in stochastic setting).
The proofs of the theorems presented in this section are in \S\ref{app:proof_of_thm_thm:}.

To handle constraints such as parameter clipping~\citep{arjovsky2017wasserstein}, we gave a \emph{projected} version of these algorithms, where $P_\Omega[\vomega']$ denotes the projection of~$\vomega'$ onto~$\Omega$ (see \S\ref{app:definitions}).
Note that when $\Omega = \RR^d$, the projection is the identity mapping (unconstrained setting).
In order to prove the convergence of these four algorithms, we will assume that $F$ is monotone:
\begin{equation} \label{eq:monotonicity}
  (F(\vomega) - F(\vomega'))^\top (\vomega - \vomega') \geq 0 \quad \forall\, \vomega,\vomega' \in \Omega \,.
\end{equation}
If $F$ can be written as~\eqref{eq:VI_stationary_cond}, it implies that the cost functions are convex.\footnote{The convexity of the cost functions in~\eqref{eq:two_player_games} is a necessary condition (not sufficient) for the operator to be monotone. In the context of a zero-sum game, the convexity of the cost functions is a sufficient condition.} Note however that general GANs parametrized with neural networks lead to non-monotone VIPs.
\begin{assumption}\label{assum:monotone_bounded}
  $F$ is \emph{monotone} and $\Omega$ is a compact convex set, such that $\max_{\vomega,\vomega' \in \Omega} \|\vomega - \vomega'\|^2 \leq R^2$.
  \vspace*{-6mm}
\end{assumption}
In that setting the quantity $g(\vomega^*) := \max_{\vomega \in \Omega} F( \vomega)^\top(\vomega^* - \vomega)$ is well defined and is equal to 0 if and only if $\vomega^*$ is a solution of~\eqref{eq:VI_problem_weak}. Moreover, if we are optimizing a \emph{zero-sum game}, we have $\vomega = (\vtheta,\vphi), \, \Omega = \Theta\times\Phi $ and $F(\vtheta,\vphi) = [\nabla_\vtheta \LL(\vtheta,\vphi)\, \; -\!\nabla_\vphi \LL(\vtheta,\vphi)]^\top$. Hence, the quantity $h(\vtheta^*,\vphi^*) := \max_{\vphi \in \Phi} \LL(\vtheta^*,\vphi) - \min_{\vtheta \in \Theta} \LL(\vtheta,\vphi^*)$ is well defined and equal to 0 if and only if $(\vtheta^*,\vphi^*)$ is a \emph{Nash equilibrium} of the game. The two functions $g$ and $h$ are called \emph{merit functions} (more details on the concept of \emph{merit functions} in \S\ref{sec:more_merit_functions}). In the following, we call,
\begin{equation}\label{eq:merit_main}
	\Err(\vomega) \defas
	\left\{
	\begin{array}
	{cl}
	\!\!\!
	\underset{(\vtheta',\vphi') \in \Omega}{\max} \LL(\vtheta,\vphi') -\LL(\vtheta',\vphi) &
	\text{if} \;\; F(\vtheta,\vphi) = [\nabla_\vtheta \LL(\vtheta,\vphi) \,\, -\!\nabla_\vphi \LL(\vtheta,\vphi)]^\top \!\! \\
   \!\!\!\underset{\vomega' \in \Omega}{\max} F( \vomega')^\top( \vomega - \vomega') &
  \text{otherwise.}
	\end{array}
	\right.
\end{equation}
\paragraph{Averaging.} Alg.~\ref{alg:AvgSGD} (AvgSGD) presents the stochastic gradient method with \emph{averaging}, which reduces to the standard (simultaneous) SGD updates for the two-player games used in the GAN literature, but returning an \emph{average} of the iterates.
\begin{theorem}\label{thm:SGM}
 Under Assump.~\ref{assum:var_bounded}, \ref{assum:sub_bounded} and~\ref{assum:monotone_bounded}, SGD with averaging (Alg.~\ref{alg:AvgSGD}) with a constant step-size gives,
 \begin{equation}\label{eq:thm_1_variance_term}
  \EE[\Err(\bar \vomega_T)] \leq \frac{R^2}{2\eta T} + \eta \frac{M^2 +\sigma^2}{2}  \quad \text{where} \quad \bar \vomega_T \defas \frac{1}{T}\sum_{t=0}^{T-1} \vomega_t \,, \quad \forall T \geq 1 \,.
\vspace*{-2mm}
\end{equation}
\end{theorem}
Thm.~\ref{thm:SGM} uses a similar proof as~\citep{nemirovski2009robust}.
The constant term $\eta (M^2+\sigma^2)/2$ in~\eqref{eq:thm_1_variance_term} is called the \emph{variance term}.
This type of bound is standard in stochastic optimization. We also provide in \S\ref{app:proof_of_thm_thm:} a similar $\tilde{O}(1/\sqrt{t})$ rate with an extra log factor when $\eta_t = \frac{\eta}{\sqrt{t}}$.
We show that this variance term is smaller than the one of \emph{SGD with prediction method}~\citep{yadav_stabilizing_2017} in \S\ref{app:SGDPreComparison}.
\paragraph{Extrapolations.} Alg.~\ref{alg:AvgExtraSGD} (AvgExtraSGD) adds an extrapolation step compared to Alg.~\ref{alg:AvgSGD} in order to reduce the oscillations due to the game between the two players.
A theoretical consequence is that it has a smaller variance term than~\eqref{eq:thm_1_variance_term}.
As discussed previously, Assump.~\ref{assum:sub_bounded} made in Thm.~\ref{thm:SGM} for the convergence of Alg.~\ref{alg:AvgSGD} is very strong in the unbounded setting.
One advantage of SGD with \emph{extrapolation} is that Thm.~\ref{thm:SEGM} does not require this assumption.

\begin{theorem}\citep[Thm.~1]{juditsky2011solving} \label{thm:SEGM}
Under Assump.~\ref{assum:var_bounded} and~\ref{assum:monotone_bounded}, if $\EE_\xi[F]$ is $L$-Lipschitz, then SGD with \emph{extrapolation and averaging} (Alg.~\ref{alg:AvgExtraSGD}) using a constant step-size $\eta \leq \frac{1}{\sqrt{3}L}$ gives,
\begin{equation}\label{eq:thm_2_variance_term}
  \EE[\Err(\bar \vomega_T)] \leq \frac{R^2}{\eta T} + \frac{7}{2} \eta \sigma^2
  \quad \text{where} \quad
  \bar \vomega_T \defas \frac{1}{T}\sum_{t=0}^{T-1} \vomega'_t \,,\quad \forall T\geq 1\,.
\vspace*{-2mm}
\end{equation}
\end{theorem}

Since in practice $\sigma \ll M$, the variance term in~\eqref{eq:thm_2_variance_term} is significantly smaller than the one in~\eqref{eq:thm_1_variance_term}.
To summarize, SGD with \emph{extrapolation} provides better convergence guarantees but requires two gradient computations and samples per iteration. This motivates our new method, Alg.~\ref{alg:AvgPastExtraSGD} (AvgPastExtraSGD) which uses \emph{extrapolation from the past} and achieves \emph{the best of both worlds} (in theory).

 \begin{theorem}\label{thm:new_AvgExtraSGD}
Under Assump.~\ref{assum:var_bounded} and~\ref{assum:monotone_bounded}, if $\EE_\xi[F]$ is $L$-Lipschitz then SGD with \emph{extrapolation from the past} using a constant step-size $\eta \leq \frac{1}{2\sqrt{3}L}$, gives that the averaged iterates converge as,
\begin{equation}\label{eq:thm_3_variance_term}
  \EE[\Err(\bar \vomega_T)] \leq \frac{R^2}{\eta T} + \frac{13}{2} \eta \sigma^2
  \quad \text{where} \quad
  \bar \vomega_T \defas \frac{1}{T}\sum_{t=0}^{T-1} \vomega'_t \, \quad \forall T \geq 1 \,.
\vspace*{-2mm}
\end{equation}
\end{theorem}
The bound is similar to the one provided in Thm.~\ref{thm:SEGM} but each iteration of Alg.~\ref{alg:AvgPastExtraSGD} is computationally half the cost of an iteration of Alg.~\ref{alg:AvgExtraSGD}.

\section{Combining the techniques with established algorithms} \label{sec:combining}
In the previous sections, we presented several techniques that converge for stochastic monotone operators.
These techniques can be combined in practice with existing algorithms. We propose to combine them to two standard algorithms used for training deep neural networks:
the Adam optimizer~\citep{kingma2014adam} and the SGD optimizer~\citep{robbins1951stochastic}. For the Adam optimizer, there are several possible choices on how to update the moments.
This choice can lead to different algorithms in practice: for example, even in the unconstrained case, our proposed Adam with extrapolation from the past (Alg.~\ref{alg:Extra-Adam}) is different from Optimistic Adam~\citep{daskalakis2017training} (the moments are updated differently). Note that in the case of a two-player game~\eqref{eq:two_player_games}, the previous convergence results can be generalized to gradient updates with a different step-size for each player by simply rescaling the objectives $\LL_G$ and $\LL_D$ by a different scaling factor. A detailed pseudo-code for Adam with extrapolation step (Extra-Adam) is given in Algorithm~\ref{alg:Extra-Adam}. Note that our interest regarding this algorithm is practical and that we do not provide any convergence proof.

\begin{figure}[thb]
\vspace*{-5mm}
\begin{center}
\scalebox{.94}{
\begin{minipage}{1.05\linewidth}
\begin{algorithm}[H]
    \caption{Extra-Adam: proposed Adam with extrapolation step. \label{alg:Extra-Adam}}
    \begin{algorithmic}
      \STATE \textbf{input:} step-size $\eta$, decay rates for moment estimates $\beta_1,\beta_2$, access to the stochastic gradients $\nabla \ell_t(\cdot)$ and to the projection $P_\Omega[\cdot]$ onto the constraint set $\Omega$, initial parameter $\vomega_0$, averaging scheme $(\rho_t)_{t\geq1}$
      \FOR{$t=0 \ldots T-1$}
      \STATE \textbf{Option 1: Standard extrapolation.}
      \bindent
      \STATE Sample new mini-batch and compute stochastic gradient: $g_t \leftarrow \nabla \ell_t(\vomega_t)$
      \eindent
      \STATE \textbf{Option 2: Extrapolation from the past}
      \bindent
      \STATE Load previously saved stochastic gradient: $g_t = \nabla \ell_{t-1/2 }(\vomega_{t-1/2 }) $
      \eindent
      \STATE Update estimate of first moment for extrapolation: $m_{t-1/2 } \leftarrow \beta_1m_{t-1} + (1- \beta_1)g_t$
      \STATE Update estimate of second moment for extrapolation: $v_{t-1/2 } \leftarrow \beta_2 v_{t-1} + (1- \beta_2)g_t^2$
      \STATE Correct the bias for the moments:
      $\hat m_{t-1/2} \leftarrow m_{t-1/2} /(1-\beta^{2t-1}_1)$, $\hat v_{t-1/2} \leftarrow v_{t-1/2} /(1-\beta_2^{2t-1})$ \\[-2mm]
      \STATE Perform \emph{extrapolation} step from iterate at time $t$: $\vomega_{t-1/2} \leftarrow P_\Omega[\vomega_{t} - \eta \frac{\hat m_{t-1/2 }}{\sqrt{\hat v_{t-1/2 }}+ \epsilon}]$
      \STATE Sample new mini-batch and compute stochastic gradient:
      $g_{t+1/2 } \leftarrow \nabla \ell_{t+1/2 }(\vomega_{t+1/2 })$
      \STATE Update estimate of first moment:
      $m_{t} \leftarrow \beta_1 m_{t-1/2 } + (1- \beta_1)g_{t+1/2 }$
      \STATE Update estimate of second moment:
      $v_{t} \leftarrow \beta_2 v_{t-1/2 } + (1- \beta_2)g_{t+1/2 }^2$
      \STATE Compute bias corrected for first and second moment:
      $\hat m_{t} \leftarrow m_{t} /(1-\beta^{2t}_1)$, $\hat v_{t} \leftarrow v_{t} /(1-\beta_2^{2t})$
      \STATE Perform \emph{update} step from the iterate at time $t$:
      $\vomega_{t+1} \leftarrow P_\Omega[\vomega_{t} - \eta \frac{\hat m_{t}}{\sqrt{\hat v_{t}}+\epsilon}]$
      \ENDFOR
      \STATE \textbf{Output:} $\vomega_{T-1/2 }$, $\vomega_{T}$ or $\bar \vomega_{T} = \sum_{t=0}^{T-1}\rho_{t+1} \vomega_{t+1/2}/ \sum_{t=0}^{T-1}\rho_{t+1}$ (see \eqref{eq:online_averaging} for online averaging)
    \end{algorithmic}
  \end{algorithm}
\end{minipage}
}
\end{center}
  \vspace*{-4mm}
\end{figure}

\section{Related Work} \label{sec:related_work}

The extragradient method is a standard algorithm to optimize variational inequalities. This algorithm has been originally introduced by~\citet{korpelevich1976extragradient} and extended by~\citet{nesterov2007dual} and~\citet{nemirovski_prox-method_2004}.
Stochastic versions of the extragradient have been recently analyzed~\citep{juditsky2011solving,kannan_optimal_2014,iusem_extragradient_2017} for stochastic variational inequalities with \emph{bounded constraints}. A linearly convergent variance reduced version of the stochastic gradient method has been proposed by~\citet{palaniappan2016stochastic} for strongly monotone variational inequalities. Extrapolation can also be related to \emph{optimistic methods}~\citep{chiang2012online,rakhlin2013online} proposed in the online learning literature (see more details in \S\ref{sub:extrapolation_from_the_past}).
Interesting non-convex results were proved, for a new notion of regret minimization, by~\citet{hazan2017efficient} and in the context of online learning for GANs by~\citet{grnarova2017online}.

 Several methods to stabilize GANs consist in transforming a zero-sum formulation into a more general game that can no longer be cast as a saddle point problem. This is the case of the \emph{non-saturating} formulation of GANs~\citep{goodfellow2014generative,fedus2017many}, the DCGANs~\citep{radford2016unsupervised}, the \emph{gradient penalty}\footnote{The gradient penalty is only added to the discriminator cost function. Since this gradient penalty depends also on the generator, WGAN-GP cannot be cast as a SP problem and is actually a non-zero sum game.} for WGANs~\citep{gulrajani2017improved}. \citet{yadav_stabilizing_2017} propose an optimization method for GANs based on AltSGD using an additional momentum-based step on the generator. \citet{daskalakis2017training} proposed a method inspired from game theory.
\citet{li2017dualing} suggest to dualize the GAN objective to reformulate it as a maximization problem and \citet{mescheder_numerics_2017} propose to add the norm of the gradient in the objective to get a better signal.
\citet{gidel2018momentum} analyzed a generalization of the bilinear example \eqref{eq:bilinear_wgan} with a focus put on the effect of momentum on this problem. They do not consider extrapolation (see \S\ref{sub:generalization_to_general_unconstrained_bilinear_objective} for more details).
\emph{Unrolling} steps~\citep{metz_unrolled_2017} can be confused with extrapolation but is fundamentally different: the perspective is to try to approximate the ``true generator objective function" unrolling for $K$ steps the updates of the discriminator and then updating the generator.

Regarding the averaging technique, some recent work appear to have already successfully used \emph{geometric averaging}~\eqref{eq:averaging_scheme} for GANs in practice, but only briefly
mention it~\citep{karras2017progressive,mescheder2018convergence}.
By contrast, the present work formally motivates and justifies the use of averaging for GANs by relating them to the VIP perspective, and sheds light on its underlying intuitions in \S\ref{sub:averaging}.
Subsequent to our first preprint, \citet{yazici2018unusual} explored averaging empirically in more depth, while \citet{mertikopoulos2018mirror} also investigated extrapolation, providing asymptotic convergence results (i.e. without any rate of convergence) in the context of \emph{coherent saddle point}. The coherence assumption is slightly weaker than monotonicity. \vspace{-2mm}

\section{Experiments} \vspace{-1mm} \label{sec:experiments}

Our goal in this experimental section is not to provide new state-of-the art results with architectural improvements or a new GAN formulation, but to show that using the \emph{techniques} (with theoretical guarantees in the monotone case) that we introduced earlier allows us to optimize standard GANs in a better way. These techniques, which are orthogonal to the design of new formulations of GAN optimization objectives, and to architectural choices, can potentially be used for the training of any type of GAN.
We will compare the following optimization algorithms: baselines are SGD and Adam using either simultaneous updates on the generator and on the discriminator (denoted \textbf{SimAdam} and \textbf{SimSGD})
or $k$ updates on the discriminator alternating with 1 update on the generator (denoted \textbf{AltSGD$\{k\}$} and \textbf{AltAdam$\{k\}$}).\footnote{In the original WGAN paper~\citep{arjovsky2017wasserstein}, the authors use $k=5$.}
Variants that use \emph{extrapolation} are denoted \textbf{ExtraSGD}  (Alg.~\ref{alg:AvgExtraSGD}) and \textbf{ExtraAdam}  (Alg.~\ref{alg:Extra-Adam}).  Variants using \emph{extrapolation from the past} are
\textbf{PastExtraSGD} (Alg.~\ref{alg:AvgPastExtraSGD})  and \textbf{PastExtraAdam} (Alg.~\ref{alg:Extra-Adam}).
We also present results using as output the \emph{averaged} iterates, adding \textbf{Avg} as a prefix of the algorithm name when we use (uniform) \emph{averaging}. \vspace{-1mm}

\subsection{Bilinear saddle point (stochastic)} \label{sub:bilinear_sp}

\begin{wrapfigure}{r}{5.8cm}
\vspace*{-10mm}
\includegraphics[width=1.00\linewidth, height = .80 \linewidth]{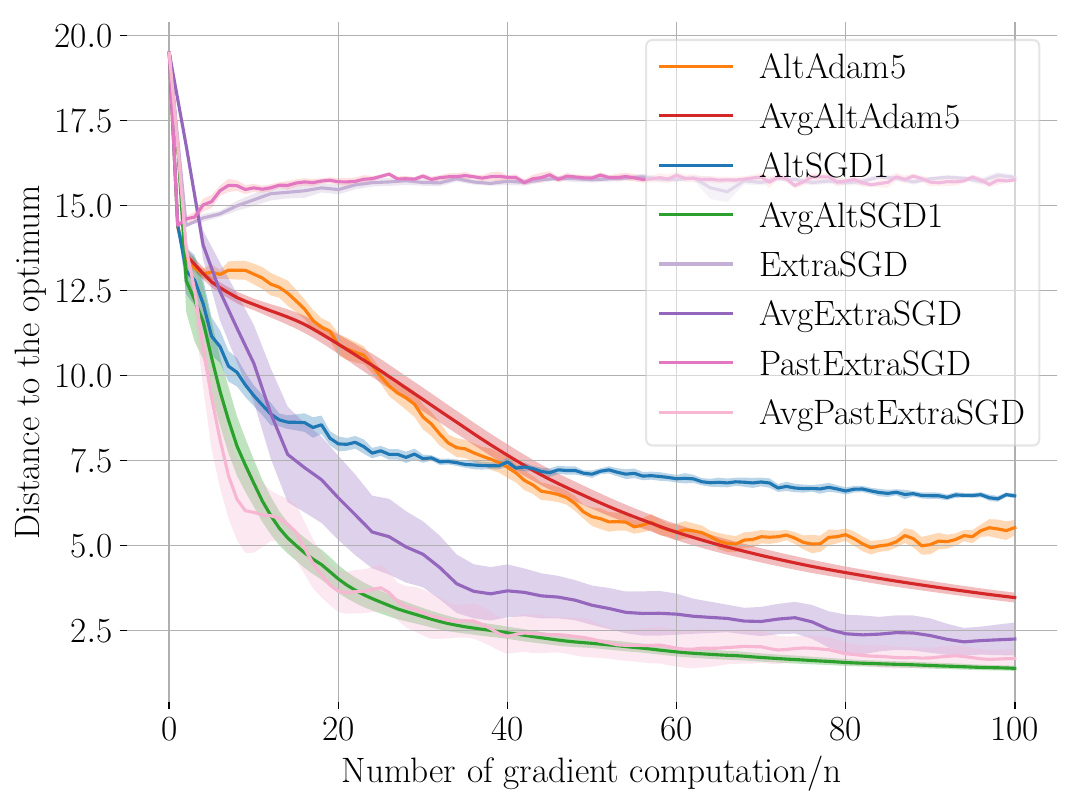}
\vspace*{-2mm}
\caption{
\small Performance of the considered stochastic optimization algorithms on the bilinear problem~\eqref{eq:sto_saddle_point}. Each method uses its respective optimal step-size found by grid-search.}
\label{fig:sto_bilinear_obj}
\vspace*{-2cm}
\end{wrapfigure}

We first test the various stochastic algorithms on a simple ($n = 10^3, d =10^3$) finite sum bilinear objective (a monotone operator) constrained to $[-1,1]^d$:
\begin{align}\label{eq:sto_saddle_point}
  &\frac{1}{n} \sum_{i=1}^n \left(  \vtheta^\top \bm{M}^{(i)} \vphi + \vtheta^\top \bm{a}^{(i)} + \vphi^\top \bm{b}^{(i)} \right) \\
  &\text{solved by $(\vtheta^*,\vphi^*)$ s.t.}  \notag
  \left\{
    \begin{array}
    {ll}
     \bar{\bm{M}} \vphi^* = -  \bar{\bm{a}} \\
     \bar{\bm{M}}^T \vtheta^* =  - \bar{\bm{b}}
    \end{array} \right.   ,
\end{align}
where
$\bar{\bm{a}} := \frac{1}{n}\sum_{i=1}^n \bm{a}^{(i)}$,
$\bar{\bm{b}} := \frac{1}{n}\sum_{i=1}^n \bm{b}^{(i)}$ and
$\bar{\bm{M}} := \frac{1}{n}\sum_{i=1}^n \bm{M}^{(i)}$.
The matrices $\bm{M}^{(i)}_{kj}, \,\bm{a}^{(i)}_k ,$ $\bm{b}^{(i)}_k\,;$
$1\leq i\leq n$, $1\leq j,k\leq d$
were randomly generated, but ensuring that $(\vtheta^*,\vphi^*)$ belongs to $[-1,1]^d$.
Results are shown in Fig.~\ref{fig:sto_bilinear_obj}. We can see that AvgAltSGD1 and AvgPastExtraSGD perform the best on this task.

\subsection{WGAN and WGAN-GP on CIFAR10} \label{sub:wgan_cifar10}
We evaluate the proposed techniques in the context of GAN training, which is a challenging stochastic optimization problem where the objectives of both players are non-convex.
We propose to evaluate the Adam variants of the different optimization algorithms (see Alg.~\ref{alg:Extra-Adam} for Adam with \emph{extrapolation}) by training two different architectures on the CIFAR10 dataset~\citep{krizhevsky2009learning}.
First, we consider a constrained zero-sum game by training the DCGAN architecture~\citep{radford2016unsupervised} with the WGAN objective and weight clipping as proposed by \citet{arjovsky2017wasserstein}. Then, we compare the different methods on a state-of-the-art architecture by training a ResNet with the WGAN-GP objective similar to \citet{gulrajani2017improved}. Models are evaluated using the inception score (IS)~\citep{salimans2016improved} computed on 50,000 samples. We also provide the FID~\citep{heusel2017gans} and the details on the ResNet architecture in \S\ref{sub:fid_wgangp}.

For each algorithm, we did an extensive search over the hyperparameters of Adam. We fixed $\beta_1=0.5$ and $\beta_2=0.9$ for all methods as they seemed to perform well. We note that as proposed by~\citet{heusel2017gans}, it is quite important to set different learning rates for the generator and discriminator.
Experiments were run with 5 random seeds for 500,000 updates of the generator.
\begin{table}
    \vspace*{-7mm}
  \centering
  \resizebox{\textwidth}{!}{
    \begin{tabular}{lcccccc}
      \toprule
      Model & \multicolumn{3}{c}{WGAN (DCGAN)}  & \multicolumn{3}{c}{WGAN-GP (ResNet)}                \\
    \cmidrule(r){2-4} \cmidrule(r){5-7}
      Method     & no avg   & uniform avg & EMA  & no avg   & uniform avg & EMA \\
      \midrule
      SimAdam  & $\mathit{6.05 \pm .12}$ & $5.85 \pm .16$ & $6.08 \pm .10$& $\mathit{7.51 \pm .17}$&$7.68 \pm .43$&$7.60 \pm .17$ \\
      AltAdam5 & $\mathit{5.45 \pm .08}$  & $5.72 \pm .06 $& $5.49 \pm .05$ & $\mathit{7.57 \pm .02}$&$8.01 \pm .05$&$7.66 \pm .03$\\
      ExtraAdam & $\mathbf{6.38\pm .09}$ & $\mathbf{6.38 \pm .20}$ & $\mathbf{6.37 \pm .08}$ & $7.90 \pm .11$ & $\mathbf{8.47 \pm .10}$&$8.13 \pm .07$  \\
      PastExtraAdam & $5.98 \pm .15$ & $6.07 \pm .19$ & $6.01 \pm .11$ & $7.84 \pm .06$ & $8.01\pm .09$ & $7.99\pm .03$ \\
      OptimAdam & $\mathit{5.74\pm .10}$ & $5.80\pm .08$ & $5.78\pm .05$ & $\mathit{7.98 \pm .08}$&$8.18 \pm .09$ & $8.10 \pm .06$\\

      \bottomrule
    \end{tabular}}
    \vspace*{-2mm}
    \caption{
    \small Best inception scores (averaged over 5 runs) achieved on CIFAR10 for every considered Adam variant. OptimAdam is the related \emph{Optimistic Adam}~\citep{daskalakis2017training} algorithm. EMA denotes \emph{exponential moving average} (with $\beta = 0.9999$, see Eq.~\ref{eq:online_averaging}). We see that the techniques of extrapolation and averaging consistently enable improvements over the baselines (in italic).
    \label{tab:inception_score} }
    \end{table}

Tab.~\ref{tab:inception_score} reports the best IS achieved on these problems by each considered method. We see that the techniques of \emph{extrapolation} and \emph{averaging} consistently enable improvements over the baselines (see \S\ref{sub:comparison_averaging} for more experiments on \emph{averaging}).
Fig.~\ref{fig:wgan_cifar10} shows training curves for each method (for their best performing learning rate), as well as samples from a ResNet generator trained with ExtraAdam on a WGAN-GP objective.
For both tasks, using an \emph{extrapolation step} and averaging with Adam (ExtraAdam) outperformed all other methods. Combining ExtraAdam with averaging yields results that improve significantly over the previous state-of-the-art IS (8.2) and FID (21.7) on CIFAR10 as reported by \citet{miyato2018spectral} (see Tab.~\ref{tab:fid_wgan_gp} for FID).
We also observed that methods based on \emph{extrapolation} are less sensitive to learning rate tuning and can be used with higher learning rates with less degradation; see \S\ref{sub:comparison_learning_rate} for more details. \vspace{-2mm}

\begingroup
\setlength{\abovecaptionskip}{5pt plus 3pt minus 2pt}
\begin{figure}
\vspace*{-4mm}
\hspace*{-2mm}
\centering
  \begin{subfigure}[b]{.39\linewidth}
  \includegraphics[width=\textwidth]{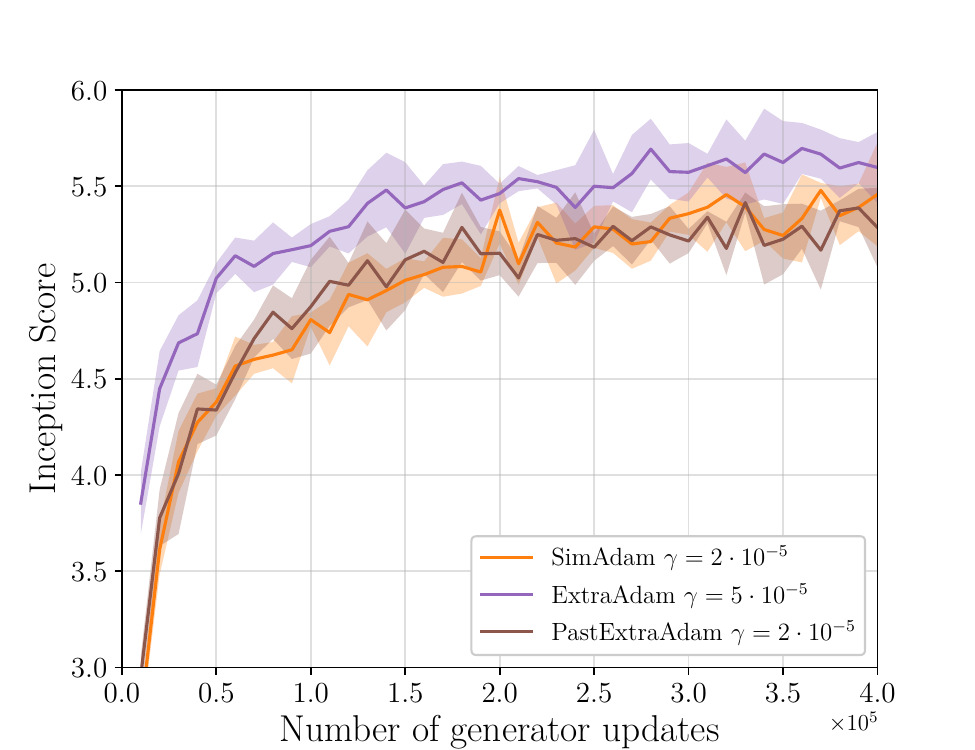}
\end{subfigure}
\hspace*{-2mm}
\begin{subfigure}[b]{.24\linewidth}
    \includegraphics[width=\textwidth]{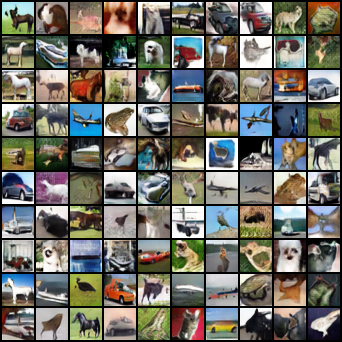}
  \label{fig:sample_wgan_cifar10}
\end{subfigure}
\hspace*{-3mm}
\begin{subfigure}[b]{.39 \linewidth}
  \centering
  \includegraphics[width=\textwidth]{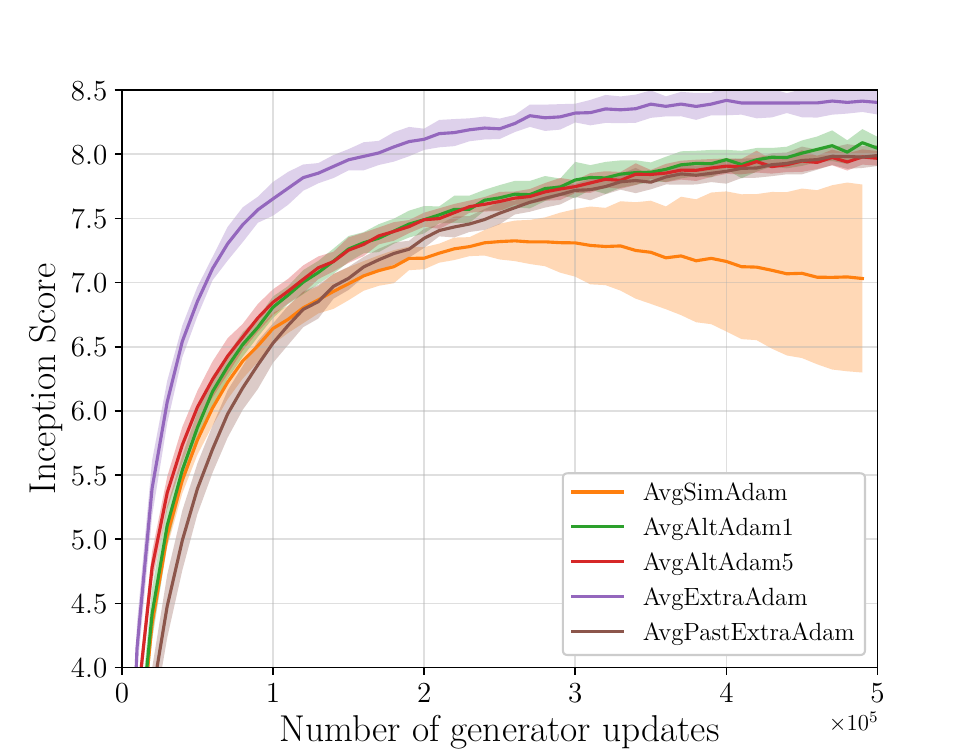}
    \hspace*{-4mm}
\end{subfigure}
    \caption{
  \small \textbf{Left: }Mean and standard deviation of the inception score computed over 5 runs for each method on WGAN trained on CIFAR10. To keep the graph readable we show only SimAdam but AltAdam performs similarly.
  \textbf{Middle:}
  Samples from a ResNet generator trained with the WGAN-GP objective using AvgExtraAdam.
  \textbf{Right: }
  WGAN-GP trained on CIFAR10: mean and standard deviation of the inception score computed over 5 runs for each method using the best performing learning rates; all experiments were run on a NVIDIA Quadro GP100 GPU. We see that ExtraAdam converges faster than the Adam baselines.
}
  \label{fig:wgan_cifar10}
  \vspace*{-4mm}
\end{figure}
\endgroup
 
\section{Conclusion} \label{sec:discussion}
We newly addressed GAN objectives in the framework of variational
inequality. We tapped into the optimization literature to provide more principled techniques to optimize such games.
We leveraged these techniques to develop practical optimization algorithms suitable for a wide range of GAN training objectives (including non-zero sum games and projections onto constraints).
We experimentally verified that this could yield better trained models, improving the previous state of the art.
The presented techniques address a fundamental problem in GAN training in a principled way, and are orthogonal to the design of new GAN architectures and objectives. They are thus likely to be widely applicable, and benefit future development of GANs.

\newpage
\subsubsection*{Acknowledgments.} \label{par:paragraph_name}
This research was partially supported by the Canada CIFAR AI Chair Program, the Canada Excellence Research Chair in “Data Science for Realtime Decision-making”, by the NSERC Discovery Grant RGPIN-2017-06936 and by a Google Focused Research award. Gauthier Gidel would like to acknowledge Benoît Joly and Florestan Martin-Baillon for bringing a fresh point of view on the proof of Proposition~\ref{prop:explicit}.

\bibliography{gan_extragrad}
\bibliographystyle{abbrvnat}

\appendix
\newpage
\section{Definitions} \label{app:definitions}
In this section, we recall usual definitions and lemmas from convex analysis.
We start with the definitions and lemmas regarding the projection mapping.
\subsection{Projection mapping} \label{sub:projection_mapping}
\begin{definition}\label{def:projection}
The projection $P_\Omega$ onto $\Omega$ is defined as,
\begin{equation}
P_\Omega(\vomega') \in \argmin_{\vomega' \in \Omega} \|\vomega - \vomega'\|_2^2 \,.
\end{equation}
\end{definition}
When $\Omega$ is a convex set, this projection is unique. This is a consequence of the following lemma that we will use in the following sections: the \emph{non-expansiveness} of the projection onto a convex set.
\begin{lemma} \label{lem:proj_non_contr}Let $\Omega$ a convex set, the projection mapping $P_\Omega: \RR^d \to \Omega$ is nonexpansive, i.e.,
\begin{equation}
  \|P_\Omega(\vomega) - P_\Omega(\vomega')\|_2\leq \|\vomega - \vomega'\|_2 \,,\quad \forall \vomega,\vomega' \in \Omega\,.
\end{equation}
\end{lemma}
This is standard convex analysis result which can be found for instance in~\citep{boyd2004convex}. The following lemma is also standard in convex analysis and its proof uses similar arguments as the proof of Lemma~\ref{lem:proj_non_contr}.
\begin{lemma}\label{lemma:ineg} Let $\vomega \in \Omega$ and $\vomega^+ \defas P_\Omega(\vomega + \uu)$,
then for all $\vomega' \in \Omega$ we have,
\begin{equation}
  \|\vomega^+-\vomega'\|_2^2 \leq \|\vomega-\vomega'\|_2^2 + 2 \uu^\top (\vomega^+-\vomega') - \|\vomega^+ -\vomega\|_2^2\,.
\end{equation}
\end{lemma}
\proof[\textbf{Proof of Lemma~\ref{lemma:ineg}}] We start by simply developing,
\begin{align*}
  \|\vomega^+-\vomega'\|_2^2 = \|(\vomega^+- \vomega) + (\vomega -\vomega')\|_2^2
  &= \|\vomega-\vomega'\|_2^2 + 2 (\vomega^+-\vomega)^\top (\vomega - \vomega') + \|\vomega^+ -\vomega\|_2^2 \\
  &= \|\vomega-\vomega'\|_2^2 + 2 (\vomega^+-\vomega)^\top (\vomega^+ - \vomega') - \|\vomega^+ -\vomega\|_2^2 \, .
\end{align*}
Then since $\vomega^+$ is the projection onto the convex set $\Omega$ of $\vomega + \uu$, we have that $(\vomega^+ - (\vomega + \uu))^\top(\vomega^+-\vomega') \leq 0 \,, \; \forall\, \vomega'\in \Omega$, leading to the result of the Lemma. \endproof

\subsection{Smoothness and Monotonicity of the operator} \label{sub:smoothness_and_monotonicity_of_the_operator}

Another important property used is the Lipschitzness of an operator.
\begin{definition}
A mapping $F: \RR^p \to \RR^d$ is said to be $L$-Lipschitz if,
\begin{equation}
  \|F(\vomega) - F(\vomega')\|_2\leq L\|\vomega-\vomega'\|_2 \,, \quad \forall \vomega,\vomega' \in \Omega\,.
\end{equation}
\end{definition}
In this paper, we also use the notion of \emph{strong monotonicity}, which is a generalization for operators of the notion of strong convexity. Let us first recall the definition of the latter,
\begin{definition}
A differentiable function $f : \Omega \to \RR$ is said to be \emph{$\mu$-strongly convex} if
\begin{equation}
  f(\vomega) \geq f(\vomega') + \nabla f(\vomega')^\top(\vomega-\vomega') + \frac{\mu}{2}\|\vomega-\vomega'\|_2^2\, \quad \forall \vomega,\vomega' \in \Omega \,.
\end{equation}
\end{definition}
\begin{definition}\label{def:convex-concave} A function $(\vtheta,\vphi) \mapsto \LL(\vtheta,\vphi)$ is said convex-concave if $\LL(\cdot,\vphi)$ is convex for all $\vphi \in \Phi$ and $\LL(\vtheta,\cdot)$ is concave for all $\vtheta \in \Theta$. An $\LL$ is said to be $\mu$-strongly convex-concave if $(\vtheta,\vphi) \mapsto \LL(\vtheta,\vphi) - \frac{\mu}{2}\|\vtheta\|_2^2 + \frac{\mu}{2}\|\vphi\|_2^2$ is convex-concave.
\end{definition}
If a function $f$ (resp. $\LL$) is strongly convex (resp. strongly convex-concave), its gradient $\nabla f$ (resp. $[\nabla_\vtheta \LL \; -\!\!\nabla_\vphi \LL]^\top$) is strongly monotone, i.e.,
\begin{definition}\label{def:strong_monotone} For $\mu>0$, an operator $F: \Omega \to \RR^d$ is said to be $\mu$-strongly monotone if
\begin{equation}
  (F(\vomega)-F(\vomega'))^\top (\vomega-\vomega') \geq \mu \|\vomega - \vomega'\|^2_2 \,.
 \end{equation}
\end{definition}

\section{Gradient methods on unconstrained bilinear games} \label{sec:explicit_implicit_and_extragradient_methods_on_unconstrained_bilinear_games}
In this section, we will prove the results provided in \S\ref{sec:optim}, namely Proposition~\ref{prop:explicit}, Proposition~\ref{prop:implicit_extra} and Theorem~\ref{thm:linear_con}. For Proposition~\ref{prop:explicit} and~\ref{prop:implicit_extra}, let us recall the context. We wanted to derive properties of some gradient methods on the following simple illustrative example
\begin{equation}\label{eq:app_unibilinear}
  \min_{\theta \in \R} \max_{\phi \in \R} \; \;\theta \cdot \phi
\end{equation}
\subsection{Proof of Proposition~\ref{prop:explicit}} \label{sub:explicit_method}
Let us first recall the proposition:
\begin{repproposition}{prop:explicit}
\propOne
\end{repproposition}
\proof Let us start with the \emph{simultaneous} update rule:
\begin{equation}\label{eq:app_update}
  \left\{\begin{aligned}
  \theta_{t+1} &=  \theta_t - \eta \phi_t \\
  \phi_{t+1} & = \phi_t +  \eta \theta_{t} \,.
  \end{aligned} \right.
\end{equation}
Then we have,
\begin{align}
  \theta_{t+1}^2 + \phi_{t+1}^2
  & =  (\theta_t - \eta \phi_t)^2 + (\phi_t +  \eta \theta_{t})^2 \\
  & = (1+\eta^2) (\theta_t^2+\phi_t^2) \,.
\end{align}
The update rule~\eqref{eq:app_update} also gives us,
\begin{equation} \label{eq:to_sum_update}
  \left\{\begin{aligned}
  \eta \phi_{t} &=  \theta_t - \theta_{t+1} \\
  \eta \theta_{t} & = \phi_{t+1} -\phi_t  \,.
  \end{aligned} \right.
\end{equation}
Summing~\eqref{eq:to_sum_update} for $0 \leq t \leq T-1$ to get telescoping sums, we get
\begin{align}
  (\eta^2 T^2) (\bar \phi_T^2 + \bar \theta_T^2)
  &= (\theta_0 - \theta_T)^2 + (\phi_0 - \phi_T)^2\\
  &= ((1+ \eta^2)^T + 1) (\theta_0^2+\phi_0^2) - 2 \theta_0 \theta_T - 2 \phi_0 \phi_T \\
  &= \Theta \left((1+ \eta^2)^T((\theta_0^2+\phi_0^2)\right).
  \end{align}

Let us continue with the \emph{alternating} update rule
\begin{equation}
	\left\{\begin{aligned}
	\theta_{t+1} &=  \theta_t - \eta \phi_t \\
	\phi_{t+1} & = \phi_t +  \eta \theta_{t+1} = \phi_t + \eta (\theta_t - \eta \phi_t)
	\end{aligned} \right.
\end{equation}
Then we have,
\begin{equation}
	\begin{bmatrix}
	\theta_{t+1} \\ \phi_{t+1}
	\end{bmatrix}
	= \begin{bmatrix}
	1 & - \eta \\ \eta & 1 - \eta^2
	\end{bmatrix}
	\begin{bmatrix}
	\theta_{t} \\ \phi_{t}
	\end{bmatrix} .
\end{equation}
By simple linear algebra, for $\eta < 2$, the matrix $M \defas \begin{bmatrix}
	1 & - \eta \\ \eta & 1 - \eta^2
	\end{bmatrix}$ has two complex conjugate eigenvalues which are
	\begin{equation}
		\lambda_{\pm} =1 - \eta\frac{ \eta \pm i \sqrt{4 - \eta^2}}{2}
	\end{equation}
	and their squared magnitude is equal to $\det(M) = 1 - \eta^2 + \eta^2 = 1$.
We can diagonalize $M$ meaning that there exists $P$ an invertible matrix such that $M = P^{-1} \diag(\lambda_+,\lambda_-) P$. Then, we have
\begin{equation}
	\begin{bmatrix}
	\theta_{t} \\ \phi_{t}
	\end{bmatrix}
	= M^t \begin{bmatrix}
	\theta_{0} \\ \phi_{0}
	\end{bmatrix}
	= P^{-1} \diag(\lambda_+^t,\lambda_-^t) P\begin{bmatrix}
	\theta_{0} \\ \phi_{0}
	\end{bmatrix}
\end{equation}
and consequently,
\begin{equation}
	\theta_t^2 + \phi_t^2
	=
	\left\|\begin{bmatrix}
	\theta_{t} \\ \phi_{t}
	\end{bmatrix}\right\|^2_{\mathbb{C}}
	= \left\| P^{-1}
	 \diag(\lambda_+^t,\lambda_-^t) P\begin{bmatrix}
	\theta_{0} \\ \phi_{0}
	\end{bmatrix} \right\|^2_{\mathbb{C}}
	\leq \|P^{-1}\|\|P\| (\theta_0^2 + \phi_0^2)
\end{equation}
where $\|\cdot\|_{\mathbb{C}}$ is the norm in $\mathbb{C}^2$ and $\|P\| := \max_{u \in \mathbb{C}^2} \frac{\|Pu\|_{\mathbb{C}}}{\|u\|_{\mathbb{C}}}$ is the induced matrix norm.
The same way we have,
\begin{equation}
	\theta_0^2 + \phi_0^2 = \left\|M^{-t}\begin{bmatrix}
	\theta_{t} \\ \phi_{t}
	\end{bmatrix}\right\|^2_{\mathbb{C}}
	= \left\| P^{-1}
	 \diag(\lambda_+^{-t},\lambda_-^{-t}) P\begin{bmatrix}
	\theta_{t} \\ \phi_{t}
	\end{bmatrix} \right\|^2_{\mathbb{C}}
	\leq \|P^{-1}\|\|P\| (\theta_t^2 + \phi_t^2)
\end{equation}
Hence, if $\theta_0^2 + \phi_0^2 >0$, the sequence $(\theta_t,\phi_t)$ is bounded but do not converge to 0. Moreover the update rule gives us,
\begin{equation}
	\left\{\begin{aligned}
	\eta \phi_{t}  &=  \theta_t - \theta_{t+1}\\
	\eta \theta_{t} & = \phi_{t} - \phi_{t-1}
	\end{aligned} \right.
	\Rightarrow
	\left\{\begin{aligned}
	\frac{\eta}{T}\sum_{t=0}^{T-1} \phi_{t}  &=  \frac{\theta_0 - \theta_{T}}{T}\\
	\frac{\eta}{T} \sum_{t=0}^{T-1}\theta_{t} & = \frac{\phi_{T-1} - \phi_{0}+ \eta \theta_0}{T}
	\end{aligned} \right.
	\Rightarrow
	\left\{\begin{aligned}
	\bar \phi_T  &=  \frac{\theta_0 - \theta_{T}}{\eta T}\\
	\bar \theta_T & = \frac{\phi_{T-1} - \phi_{0} + \eta \theta_0}{\eta T}
	\end{aligned} \right.
\end{equation}
Consequently, since $\theta_t^2 + \phi_t^2 =  \Theta(\theta_0^2 + \phi_0^2)$,
\begin{equation}
	\sqrt{\bar \theta_t^2 + \bar \phi_t^2} =  \Theta \left(\frac{\sqrt{\theta_0^2 + \phi_0^2}}{\eta t}\right)
\end{equation}
\endproof

\subsection{Implicit and extrapolation method} \label{app:extragradient_method}
In this section, we will prove a slightly more precise proposition than Proposition~\ref{prop:implicit_extra},
\begin{repproposition}{prop:implicit_extra}
The squared norm of the iterates $N_t^2 \defas \theta_t^2 + \phi_t^2$, where the update rule of $\theta_t$ and $\phi_t$ is defined in~\eqref{eq:update_implicit_extra}, decrease geometrically for any $0<\eta <1$ as,\footnote{Note that the relationship \eqref{eq:implicitConverge} holds actually for \emph{any} $\eta$ for the implicit method, and thus the decrease is geometric for any non-zero step size.} 
\begin{equation} \label{eq:implicitConverge}
\emph{Implicit:} \;\; N_{t+1}^2 = \frac{N_t^2}{1+ \eta^2}\,, \quad
\emph{Extrapolation:} \;\; N_{t+1}^2 = (1-\eta^2+\eta^4)N_t^2 \, ,\quad \forall t \geq 0
\end{equation}
\end{repproposition}
\proof
Let us recall the update rule for the implicit method
\begin{equation}
	\left\{\begin{aligned}
	\theta_{t+1} &=  \theta_t - \eta \phi_{t+1} \\
	\phi_{t+1} & = \phi_t +  \eta \theta_{t+1}
	\end{aligned} \right.
	\Rightarrow
	\left\{\begin{aligned}
	(1+ \eta^2)\theta_{t+1} &=  \theta_t - \eta \phi_{t} \\
	(1+ \eta^2)\phi_{t+1} & = \phi_t +  \eta \theta_{t}
	\end{aligned} \right.
\end{equation}
Then,
\begin{align}
	(1+\eta^2)^2(\theta_{t+1}^2 + \phi_{t+1}^2 )
	& = (\theta_t - \eta \phi_{t})^2 + (\phi_t +  \eta \theta_{t})^2 \\
	& = \theta_{t}^2 + \phi_{t}^2 +
	+\eta^2 (\theta_{t}^2+ \phi_{t}^2),
\end{align}
implying that
\begin{equation}
	\theta_{t+1}^2 + \phi_{t+1}^2
	=\frac{\theta_{t}^2 + \phi_{t}^2 }{1 + \eta^2},
\end{equation}
which is valid for \emph{any} $\eta$.

For the extrapolation method, we have the update rule
\begin{equation}
	\left\{\begin{aligned}
	\theta_{t+1} &=  \theta_t - \eta (\phi_{t} + \eta \theta_t) \\
	\phi_{t+1} & = \phi_t +  \eta(\theta_{t} - \eta \phi_t)
	\end{aligned} \right.
\end{equation}
Implying that,
\begin{align}
	\theta_{t+1}^2 + \phi_{t+1}^2
	& = (\theta_t - \eta (\phi_{t} + \eta \theta_t))^2 + (\phi_t +  \eta (\theta_{t} - \eta \phi_t))^2 \\
	& = \theta_{t}^2 + \phi_{t}^2 -
	2 \eta^2 (\theta^2 +\phi^2)
	+\eta^2 ((\theta_{t}-\eta \phi_t)^2+ (\phi_{t}+\eta \theta_t)^2)\\
	& = (1 - \eta^2 + \eta^4)(\theta_{t}^2 + \phi_{t}^2)
\end{align}
\endproof

\subsection{Generalization to general unconstrained bilinear objective} \label{sub:generalization_to_general_unconstrained_bilinear_objective}

In this section, we will show how to simply extend the study of the algorithm of interest provided in \S\ref{sec:optim} on the general unconstrained bilinear example,
\begin{equation}\label{eq:app_bilinear}
  \min_{\vtheta \in \R^d} \max_{\vphi \in \R^p} \; \;\vtheta^\top \mA \vphi - \vb^\top \vtheta -\vc^\top \vphi
\end{equation}
where, $\mA \in \R^{d\times p}, \, \vb \in \R^d$ and $\vc \in \R^p$. The only assumption we will make is that this problem is feasible which is equivalent to say that there exists a solution $(\vtheta^*,\vphi^*)$ to the system
\begin{equation}
\left\{
\begin{aligned}
  &\mA \vphi^* = \vb \\
  &\mA^\top \vtheta^* = \vc \;.
\end{aligned}
\right.
\end{equation}
In this case, we can re-write~\eqref{eq:app_bilinear} as
\begin{equation}\label{eq:app_bilinear2}
  \min_{\vtheta \in \R^d} \max_{\vphi \in \R^p} \; \;(\vtheta-\vtheta^*)^\top \mA (\vphi-\vphi^*) + c
\end{equation}
where $c:=-\vtheta^{*\top} \mA \vphi^*$ is a constant that does not depend on $\vtheta$ and $\vphi$.

First, let us show that we can reduce the study of simultaneous, alternating, extrapolation and implicit updates rules for~\eqref{eq:app_bilinear} to the study of the respective unidimensional updates~\eqref{eq:update_rules_bilinear} and \eqref{eq:update_implicit_extra}.

This reduction has already been proposed by~\citet{gidel2018momentum}. For completeness, we reproduce here similar arguments. The following lemma is a bit more general than the result provided by~\citet{gidel2018momentum}. It states that the study of a wide class of \emph{unconstrained} first order method on~\eqref{eq:app_bilinear} can be reduced to the study of the method on~\eqref{eq:app_unibilinear}, with potentially rescaled step-sizes.

Before explicitly stating the lemma, we need to introduce a bit of notation to encompass easily our several methods in a unified way. First, we let $\vomega_t := (\vtheta_t,\vphi_t)$, where the index $t$ here is a more general index which can vary more often than the one in \S\ref{sec:optim}. For example, for the extrapolation method, we could consider $\vomega_1 = \vomega'_{0+1/2}$ and $\vomega_2 = \vomega'_{1}$, where $\vomega'$ was the sequence defined for the extragradient. For the alternated updates, we can consider $\vomega_1 = (\vtheta'_1,\vphi'_0)$ and $\vomega_2 = (\vtheta'_1,\vphi'_1)$ (this also defines $\vtheta_2 = \vtheta'_1$), where $\vtheta'$ and $\vphi'$ were the sequences originally defined for alternated updates. We are thus ready to state the lemma.  

\begin{lemma}
Let us consider the following very general class of first order methods on~\eqref{eq:app_bilinear}, i.e.,

\begin{equation}
\begin{aligned}
\vtheta_t &\in  \vtheta_0 + \text{span}( F_\vtheta(\vomega_0),\ldots, F_\vtheta(\vomega_t)) \,,\quad \forall t \in \sN\,, \\
\vphi_t &\in  \vphi_0 + \text{span}( F_\vphi(\vomega_0),\ldots, F_\vphi(\vomega_t)) \,,\quad \forall t \in \sN\,, 
\end{aligned}
\end{equation}
where $\vomega_t := (\vtheta_t,\vphi_t)$ and $F_\vtheta(\vomega_t) := \mA \vphi_t - \vb$,  $F_\vphi(\vomega_t) = \mA^\top \vtheta_t - \vc$. Then, we have
\begin{equation}
  \vtheta_t = \mU^\top (\tilde \vtheta_t - \vtheta^*) \quad \text{and} \quad \vphi_t = \mV^\top (\tilde \vphi_t - \vphi^*) \,,
\end{equation}
where $\mA = \mU \mD \mV^\top$ (SVD decomposition) and the couples $([\tilde \vtheta_t]_i,[\tilde \vphi]_i)_{1 \leq i \leq r}$ follow the update rule of the same method on a unidimensional problem~\eqref{eq:app_unibilinear}. In particular, for our methods of interest, the couples $([\tilde \vtheta_t]_i,[\tilde \vphi]_i)_{1 \leq i \leq r}$ follow the same update rule with a respective step-size $\sigma_i \eta$, where $\sigma_i$ are the singular values on the diagonal of $\mD$.
\end{lemma}
\proof
Our general class of first order methods can be written with the following update rules:
\begin{align*}
  \vtheta_{t+1} = \vtheta_0 + \sum_{s=0}^{t+1} \lambda_{st} \mA (\vphi_s - \vphi^*) \\
  \vphi_{t+1} = \vphi_0 + \sum_{s=0}^{t+1} \mu_{st} \mA^\top( \vtheta_s - \vtheta^*) \,,
\end{align*}
where $\lambda_{it}, \, \mu_{it} \in \R\,, \; 0 \leq i \leq t+1$. We allow the dependence on $t$ for the algorithm coefficients $\lambda$ and $\mu$ (for example, the alternating rule would zero out some of the coefficients depending on whether we are updating $\vtheta$ or $\vphi$ at the current iteration). Notice also that if both $\lambda_{(t+1)t}$ and $\mu_{(t+1)t}$ are non-zero, we have an \emph{implicit} scheme.

Thus, using the SVD of $\mA=\mU\mD \mV^\top$, we get
\begin{align*}
  \mU^\top(\vtheta_{t+1} - \vtheta^*) =  \mU^\top(\vtheta_0-\vtheta^*) + \sum_{s=0}^{t+1} \lambda_{st} \mD \mV^\top (\vphi_s - \vphi^* ) \\
  \mV^\top(\vphi_{t+1} -\vphi^*) = \mV^\top( \vphi_0 -\vphi^* ) + \sum_{s=0}^{t+1} \mu_{st} \mD^\top \mU^\top (\vtheta_s - \vtheta^*) \,,
\end{align*}
which is equivalent to
\begin{equation}
  \left \{
\begin{aligned}
  \tilde \vtheta_{t+1} =  \tilde \vtheta_0 + \sum_{s=0}^{t+1} \lambda_{st} \mD \tilde \vphi_s  \\
  \tilde \vphi_{t+1}  = \tilde \vphi_0 + \sum_{s=0}^{t+1} \mu_{st} \mD^\top  \tilde \vtheta_s  \,,
\end{aligned}
\right.
\end{equation}
where $\mD$ is a rectangular matrix with zeros except on a diagonal block of size $r$.
Thus, each coordinate of $\tilde \vtheta_{t+1}$ and $\tilde \vphi_{t+1}$ are updated independently, reducing the initial problem to $r$ unidimensional problems,
\begin{equation}\label{eq:uni}
  \left \{
\begin{aligned}
  [\tilde \vtheta_{t+1}]_i =  [\tilde \vtheta_0]_i + \sum_{s=0}^{t+1} \lambda_{st} \sigma_i [\tilde \vphi_s]_i  \\
  [\tilde \vphi_{t+1}]_i  = [\tilde \vphi_0]_i + \sum_{s=0}^{t+1} \mu_{st} \sigma_i  [\tilde \vtheta_s]_i
\end{aligned}
\right. \qquad 1\leq i \leq r \,,
\end{equation}
where $\sigma_1 \geq \ldots \geq \sigma_r  > 0$ are the positive diagonal coefficients of $\mD$.

Finally, for the coordinate $i$ where the diagonal coefficient of $\mD$ is equal to 0, we can notice that the sequence $([\tilde \vtheta_{t}]_i, [\tilde \vphi_{t}]_i)$ is constant. Moreover, we have the freedom to chose any $[\vtheta^*]_i \in \R$ and $[\vphi^*]_i \in \R$ as a coordinate of the solution of~\eqref{eq:app_bilinear}. We thus set them respectively equal to $[\vtheta_0]_i$ and $[\vphi_0]_i$.
The update rule~\eqref{eq:uni} corresponds to the update rule of the general first order method considered on this proof on the unidimensional problem~\eqref{eq:app_unibilinear}.

Note that the only additional restriction is that the coefficients $(\lambda_{st})$ and $(\sigma_{st})$ (that are the same for $1\leq i \leq r$) are rescaled by the singular values of $\mA$. In practice, for our methods of interest with a step-size $\eta$, it corresponds to the study of $r$ unidimensional problem with a respective step-size $\sigma_i \eta \,,\, 1\leq i \leq r$.
\endproof

From this lemma, an extension of Proposition~\ref{prop:explicit} and~\ref{prop:implicit_extra} directly follows to the general unconstrained bilinear objective~\eqref{eq:app_bilinear}. We note
\begin{equation}
  N_t^2 := \dist(\vtheta_t,\Theta^*)^2 +  \dist(\vphi_t,\Phi^*)^2 \,,
\end{equation}
where $(\Theta^*,\Phi^*)$ is the set of solutions of~\eqref{eq:app_bilinear}. The following corollary is divided in two points, the first point is a result from~\citet{gidel2018momentum} (note that the result on the average is a straightforward extension of the one provided in Proposition~\ref{prop:explicit} and was not provided by~\citet{gidel2018momentum}), the second result is new. Very similar asymptotic upper bounds regarding extrapolation and implicit methods can be derived by~\citet{tseng1995linear} computing the exact values of the constant $\tau_1$ and $\tau_2$ (and noticing that $\tau_3 = \infty$) introduced in~\citep[Eq.~3 \& 4]{tseng1995linear} for the unconstrained bilinear case. However, since~\citet{tseng1995linear} works in a very general setting, the bound are not as tight as ours and his proof technique is a bit more technical. Our reduction above provides here a simple proof for our simple setting.
\begin{corollary}
\begin{itemize}
  \item{\citet{gidel2018momentum}}: The \emph{simultaneous} iterates diverge geometrically and the \emph{alternating} iterates are bounded but do not converge to 0 as,
\begin{equation}
      \emph{Simultaneous:} \;\; N_{t+1}^2 = (1+(\sigma_{\min}(\mA)\eta)^2) N_t^2 \,,
      \quad
       \emph{Alternating:} \;\;N_t^2 = \Theta (N_0^2) \,,
\end{equation}
where $u_t = \Theta(v_t) \Leftrightarrow \exists \alpha,\beta, t_0 >0 \text{ such that } \forall t \geq t_0, \, \alpha v_t \leq u_t \leq \beta v_t$.
The uniform average $(\bar \theta_t,\bar \phi_t) \defas \frac{1}{t} \sum_{s=0}^{t-1}( \theta_{s},\phi_s)$ of the \emph{simultaneous} updates (resp. the \emph{alternating updates}) diverges (resp. converges to 0) as,
\begin{equation}\notag
  \emph{Simultaneous:} \;\, \bar N_t^2 \leq \Theta\left(\frac{N_0^2}{t^2} (1+(\sigma_{\min}(\mA)\eta)^2)^t \right) \! ,
  \;
   \emph{Alternating:} \;\,
  \bar N_t^2 = \Theta \left(\frac{N_0^2}{t^2}\right) .
\end{equation}
  \item \emph{Extrapolation} and \emph{Implicit method}: The iterates respectively generated by the update rules~\eqref{eq:extrapolation} and~\eqref{eq:implicit_step} on a bilinear unconstrained problem~\eqref{eq:app_bilinear} do converge linearly for any $ 0<\eta < \frac{1}{\sigma_{\max}(\mA)}$ at a rate,\footnote{As before, the inequality~\eqref{eq:implicit72} for the implicit scheme is actually valid for any step-size.} 
  \begin{align}
&\emph{Implicit:} \;\; N_{t+1}^2 \leq \frac{N_t^2}{1+ (\sigma_{\min}(\mA)\eta)^2}\,, \quad \forall t \geq 0\\ \label{eq:implicit72}
&\emph{Extrapolation:} \;\; N_{t+1}^2 \leq (1-(\sigma_{\min}(\mA)\eta)^2+(\sigma_{\min}(\mA)\eta)^4)N_t^2 \, ,\quad \,\forall t \geq 0 \,.
\end{align}
Particularly, for $\eta = \frac{1}{2\sigma_{\max}(\mA)}$ we get for the extrapolation method,
\begin{equation}
  \emph{Extrapolation:} \;\; N_{t+1}^2 \leq (1- \tfrac{1}{8 \kappa})^tN_0^2 \, ,\quad \,\forall t \geq 0 \,. \end{equation}
where $\kappa:= \frac{\sigma_{\max}(\mA)^2}{\sigma_{\min}(\mA)^2}$ is the condition number of $\mA^\top \mA$.
\end{itemize}
\end{corollary}

\subsection{Extrapolation from the past for strongly convex objectives} \label{app:extrapolation_from_the_past}
Let us recall what we call \emph{projected extrapolation form the past}, where we used the notation $\vomega_t' = \vomega_{t+1/2}$ for compactness,
\begin{align}
  \text{Extrapolation from the past:} \; \;&\vomega_t' = P_\Omega[\vomega_t - \eta F(\vomega'_{t-1})] \quad \label{app:extrapolation_past}  \\
  \text{Perform update step:} \quad  &\vomega_{t+1} = P_\Omega[\vomega_t - \eta F(\vomega_t')]\;\; \text{and store:} \; \;F(\vomega_t') \label{app:update_past}
\end{align}
where $P_\Omega[\cdot]$ is the projection onto the constraint set $\Omega$.
An operator $F: \Omega \to \RR^d$ is said to be $\mu$-strongly monotone if
\begin{equation}
  (F(\vomega)-F(\vomega'))^\top (\vomega-\vomega') \geq \mu \|\vomega - \vomega'\|^2_2 \,.
 \end{equation}

If $F$ is strongly monotone, we can prove the following theorem:
\begin{reptheorem}{thm:linear_con}
  \theoremLinearConvPast
\end{reptheorem}
\proof In order to prove this theorem, we will prove a slightly more general result,
\begin{equation}
    \|\vomega_{t+1}-\vomega^*\|_2^2 + \|\vomega'_{t-1}-\vomega'_t\|_2^2
  \leq  \left( 1-  \frac{\mu}{4L} \right)(\|\vomega_t-\vomega^*\|_2^2 + \|\vomega'_{t-1}-\vomega'_{t-2}\|_2^2) \,.
\end{equation}
with the convention that $\vomega'_0 = \vomega'_{-1} =\vomega'_{-2}$.
It implies that
\begin{equation}
	\|\vomega_{t+1}-\vomega^*\|_2^2 \leq \|\vomega_{t+1}-\vomega^*\|_2^2 + \|\vomega'_{t-1}-\vomega'_t\|_2^2
  \leq  \left( 1-  \frac{\mu}{4L} \right)^t \|\vomega_0-\vomega^*\|_2^2 \,.
\end{equation}
Let us first proof three technical lemmas.
\begin{lemma}\label{lemma:strong_monotone} If $F$ is $\mu$-strongly monotone, we have
\begin{equation}
  \mu \left( \|\vomega_t-\vomega^*\|_2^2 - 2\|\vomega'_t-\vomega_t\|_2^2 \right) \leq 2F(\vomega'_t)^\top (\vomega'_t-\vomega^*)
  \,, \quad  \forall \,\vomega^* \in \Omega^* \,.
\end{equation}
\end{lemma}
\proof
By strong monotonicity and optimality of $\vomega^*$,
\begin{equation}
 2\mu  \|\vomega'_t-\vomega^*\|_2^2 \leq  2F(\vomega^*)^\top (\vomega'_t-\vomega^*)+ 2\mu  \|\vomega'_t-\vomega^*\|_2^2  \leq 2F(\vomega'_t)^\top (\vomega'_t-\vomega^*)
\end{equation}
and then we use the inequality $2\|\vomega'_t-\vomega^*\|_2^2 \geq  \|\vomega_t-\vomega^*\|_2^2 - 2\|\vomega'_t-\vomega_t\|_2^2 $ to get the result claimed.
\endproof
\begin{lemma}\label{lemma:4} If $F$ is $L$-Lipschitz, we have for any $\vomega \in \Omega$,
\begin{equation}
	2 \eta_t F(\vomega'_t)^\top(\vomega'_t-\vomega)
   \leq \|\vomega_t - \vomega\|_2^2 -
  \|\vomega_{t+1}-\vomega\|_2^2  - \|\vomega'_t -\vomega_t\|_2^2
  + \eta_t^2 L^2 \|\vomega'_{t-1}-\vomega'_t\|_2^2 \, .
\end{equation}
\end{lemma}
\proof
Applying Lemma~\ref{lemma:ineg} for $(\vomega,\uu,\vomega^+,\vomega') = (\vomega_t,-\eta_t F(\vomega'_t),\vomega_{t+1},\vomega)$ and $(\vomega,\uu,\vomega^+,\vomega') = (\vomega_{t},-\eta_t F(\vomega'_{t-1}),\vomega'_{t},\vomega_{t+1})$, we get,
\begin{equation}\label{eq:proof_alg_v1.2_1}
  \|\vomega_{t+1}-\vomega\|_2^2
  \leq \|\vomega_t - \vomega\|_2^2 - 2 \eta_t F(\vomega'_t)^\top(\vomega_{t+1}-\vomega) - \|\vomega_{t+1}-\vomega_t\|_2^2
\end{equation}
and
\begin{equation} \label{eq:proof_alg_v1.2_2}
  \|\vomega'_t -\vomega_{t+1}\|_2^2 \leq \|\vomega_t-\vomega_{t+1}\|_2^2 - 2 \eta_t F(\vomega'_{t-1})^\top(\vomega'_t-\vomega_{t+1}) - \|\vomega'_t -\vomega_t\|_2^2\,.
\end{equation}
Summing \eqref{eq:proof_alg_v1.2_1} and~\eqref{eq:proof_alg_v1.2_2} we get,
\begin{align}
  \|\vomega_{t+1}-\vomega\|_2^2
  &\leq \|\vomega_t - \vomega\|_2^2 - 2 \eta_t F(\vomega'_t)^\top(\vomega_{t+1}-\vomega)\\
  &\quad - 2 \eta_t F(\vomega'_{t-1})^\top(\vomega'_t-\vomega_{t+1})  - \|\vomega'_t -\vomega_t\|_2^2 - \|\vomega'_t -\vomega_{t+1}\|_2^2 \\
  & =  \|\vomega_t - \vomega\|_2^2 - 2 \eta_t F(\vomega'_t)^\top(\vomega'_t-\vomega)  - \|\vomega'_t -\vomega_t\|_2^2 - \|\vomega'_t -\vomega_{t+1}\|_2^2 \notag\\
  &\quad - 2 \eta_t (F(\vomega'_{t-1}) -F(\vomega'_t))^\top(\vomega'_t-\vomega_{t+1}) \, .
\end{align}
Then, we can use the Young's inequality $2a^\top b \leq \|a\|_2^2 + \|b\|_2^2$ to get,
\begin{align}
  \|\vomega_{t+1}-\vomega\|_2^2
  & \leq \|\vomega_t - \vomega\|_2^2 - 2 \eta_t F(\vomega'_t)^\top(\vomega'_t-\vomega) +  \eta_t^2 \|F(\vomega'_{t-1})-F(\vomega'_t)\|_2^2 \notag\\
  & \quad + \|\vomega'_t-\vomega_{t+1}\|_2^2
  - \|\vomega'_t -\vomega_t\|_2^2 - \|\vomega'_t -\vomega_{t+1}\|_2^2 \\
  & = \|\vomega_t - \vomega\|_2^2 - 2 \eta_t F(\vomega'_t)^\top(\vomega'_t-\vomega)+  \eta_t^2 \|F(\vomega'_{t-1})-F(\vomega'_t)\|_2^2 - \|\vomega'_t -\vomega_t\|_2^2   \notag \\
  & \leq \|\vomega_t - \vomega\|_2^2 - 2 \eta_t F(\vomega'_t)^\top(\vomega'_t-\vomega)+  \eta_t^2 L^2\|\vomega'_{t-1}-\vomega'_t\|_2^2 - \|\vomega'_t -\vomega_t\|_2^2 \, .
\end{align}
\endproof

\begin{lemma}\label{lemma:7} For all $t \geq 0$, if we set $\vomega'_{-2} = \vomega'_{-1} = \vomega'_0$ we have
\begin{equation}
  \|\vomega'_{t-1}-\vomega'_t\|_2^2 \leq 4  \|\vomega_t-\vomega'_t\|_2^2 + 4 \eta_{t-1}^2 L^2 \|\vomega'_{t-1}-\vomega'_{t-2}\|_2^2  - \|\vomega'_{t-1}-\vomega'_t\|_2^2 \,.
\end{equation}
\end{lemma}
\proof
We start with $\|a + b\|_2^2 \leq 2\|a\|^2 + 2 \|b\|^2$.
\begin{equation}
  \|\vomega'_{t-1}-\vomega'_t\|_2^2 \leq 2 \|\vomega_t-\vomega'_t\|_2^2 + 2 \|\vomega_t-\vomega'_{t-1}\|_2^2 \,.
  \label{eq:proof_lemma_2_3}
\end{equation}
Moreover, since the projection is contractive we have that
\begin{align}
  \|\vomega_t-\vomega'_{t-1}\|_2^2
  &\leq \|\vomega_{t-1} - \eta_{t-1} F(\vomega'_{t-1}) - \vomega_{t-1} - \eta_{t-1} F(\vomega'_{t-2})\|_2^2 \\
  &= \eta_{t-1}^2 \| F(\vomega'_{t-1})- F(\vomega'_{t-2}) \|_2^2 \\
  &\leq \eta_{t-1}^2  L^2 \|\vomega'_{t-1}-\vomega'_{t-2}\|_2^2 \,.
  \label{eq:proof_lemma_2_4}
\end{align}
 Combining~\eqref{eq:proof_lemma_2_3} and~\eqref{eq:proof_lemma_2_4} we get,
\begin{align}
  \|\vomega'_{t-1}-\vomega'_t\|_2^2
  & = 2 \|\vomega'_{t-1}-\vomega'_t\|_2^2 - \|\vomega'_{t-1}-\vomega'_t\|_2^2 \\
  & \leq 4 \|\vomega_t-\vomega'_t\|_2^2 + 4 \|\vomega_t-\vomega'_{t-1}\|_2^2 - \|\vomega'_{t-1}-\vomega'_t\|_2^2 \\
  & \leq 4  \|\vomega_t-\vomega'_t\|_2^2 + 4 \eta_{t-1}^2 L^2 \|\vomega'_{t-1}-\vomega'_{t-2}\|_2^2- \|\vomega'_{t-1}-\vomega'_t\|_2^2 \,.
\end{align}
\endproof

\proof[\textbf{Proof of Theorem~\ref{thm:linear_con}.}]
Let $\vomega^* \in \Omega^*$ be an optimal point of~\eqref{eq:VI_problem_weak}.
Combining Lemma~\ref{lemma:strong_monotone} and Lemma~\ref{lemma:4} we get,
\begin{equation*}
    \eta_t\mu \left( \|\vomega_t-\vomega^*\|_2^2 - 2\|\vomega'_t-\vomega_t\|_2^2 \right) \leq \|\vomega_t-\vomega^*\|_2^2 - \|\vomega_{t+1}-\vomega^*\|_2^2 + \eta_t^2L^2 \|\vomega'_{t-1} -\vomega'_t\|_2^2 - \|\vomega'_t-\vomega_t\|_2^2
\end{equation*}
leading to,
\begin{equation}
  \|\vomega_{t+1}-\vomega^*\|_2^2
  \leq  \left( 1- \eta_t \mu \right) \|\vomega_t-\vomega^*\|_2^2
        + \eta_t^2L^2 \|\vomega'_{t-1} -\vomega'_t\|_2^2 - (1 - 2 \eta_t \mu )\|\vomega'_t-\vomega_t\|_2^2
\end{equation}
Now using Lemma~\ref{lemma:7} we get,
\begin{align}
  \|\vomega_{t+1}-\vomega^*\|_2^2
  &\leq  \left( 1- \eta_t \mu \right) \|\vomega_t-\vomega^*\|_2^2 \notag
         +\eta_t^2L^2 ( 4 \eta_{t-1}^2 L^2 \|\vomega'_{t-1}-\vomega'_{t-2}\|_2^2 - \|\vomega'_{t-1}-\vomega'_t\|_2^2) \\
  & \quad- (1 - 2 \eta_t \mu - 4  \eta_t^2 L^2 )\|\vomega'_t-\vomega_t\|_2^2
\end{align}

Now with $\eta_t = \frac{1}{4L} \leq \frac{1}{4 \mu}$ we get,
\begin{equation*}
  \|\vomega_{t+1}-\vomega^*\|_2^2
  \leq  \left( 1-  \frac{\mu}{4L} \right) \|\vomega_t-\vomega^*\|_2^2
          + \frac{1}{16}  \left(\frac{1}{4}\|\vomega'_{t-1}-\vomega'_{t-2}\|_2^2 - \|\vomega'_{t-1}-\vomega'_t\|_2^2 \right)
\end{equation*}
Hence, using the fact that $\frac{\mu}{4L} \leq \frac{1}{4}$ we get,
\begin{equation}
    \|\vomega_{t+1}-\vomega^*\|_2^2 + \tfrac{1}{16} \|\vomega'_{t-1}-\vomega'_t\|_2^2
  \leq  \left( 1-  \frac{\mu}{4L} \right)\left(\|\vomega_t-\vomega^*\|_2^2 + \tfrac{1}{16} \|\vomega'_{t-1}-\vomega'_{t-2}\|_2^2\right) \,.
\end{equation}
\endproof

\section{More on merit functions} \label{sec:more_merit_functions}
In this section, we will present how to handle an \emph{unbounded} constraint set $\Omega$ with a more refined \emph{merit function} than~\eqref{eq:merit_main} used in the main paper.
Let $F$ be the continuous operator and $\Omega$ be the constraint set associated with the VIP,
\begin{equation*} \tag{VIP}
  \text{find} \; \vomega^* \in \Omega \quad \text{such that} \quad F(\vomega^*)^\top (\vomega - \vomega^*) \geq 0 \, , \; \;\forall \vomega \in \Omega\,.
\end{equation*}
When the operator $F$ is monotone, we have that $F(\vomega^*)^\top (\vomega - \vomega^*)\leq F(\vomega)^\top (\vomega - \vomega^*) \, , \; \forall \vomega, \vomega^*$. Hence, in this case
\eqref{eq:VI_problem_weak} implies a stronger formulation sometimes called \emph{Minty variational inequality}~\citep{crespi2005MVI}:
\begin{equation*}\label{eq:VI_problem_strong} \tag{MVI}
  \text{find} \; \vomega^* \in \Omega \quad \text{such that} \quad F(\vomega)^\top (\vomega - \vomega^*) \geq 0 \, , \; \;\forall \vomega \in \Omega\,.
\end{equation*} This formulation is stronger in the sense that if \eqref{eq:VI_problem_strong} holds for some $\vomega^* \in \Omega$, then~\eqref{eq:VI_problem_weak} holds too.
A \emph{merit function} useful for our analysis can be derived from this formulation.
Roughly, a merit function is a convergence measure.
More formally, a function $g: \Omega \to \RR\,$ is called a \emph{merit function} if $g$ is non-negative such that $g(\vomega) = 0 \Leftrightarrow \vomega \in \Omega^*$~\citep{larsson1994class}. A way to derive a merit function from~\eqref{eq:VI_problem_strong} would be to use $g(\vomega^*) = \sup_{\vomega \in \Omega} F(\vomega)^\top (\vomega^*-\vomega)$ which is zero if and only if \eqref{eq:VI_problem_strong} holds for $\vomega^*$. To deal with unbounded constraint sets (leading to a potentially infinite valued function outside of the optimal set), we use the \emph{restricted merit function}~\citep{nesterov2007dual}:
\begin{equation}
\label{eq:merit_VI}
  \Err_R(\vomega_t) \defas \max_{\vomega\in \Omega, \|\vomega-\vomega_0\|\leq R} F(\vomega)^\top(\vomega_t-\vomega) \,.
\end{equation}
This function acts as merit function for~\eqref{eq:VI_problem_weak} on the interior of the open ball of radius $R$ around $\vomega_0$, as shown in Lemma~1 of~\citet{nesterov2007dual}. That is, let $\Omega_R \defas \Omega \cap \{\vomega : {\|\vomega - \vomega_0\| < R}\}$. Then for any point $\hat \vomega \in \Omega_R$, we have:
\begin{equation}
\Err_R(\hat \vomega) = 0 \Leftrightarrow \hat \vomega \in \Omega^*\cap\Omega_R .
 \end{equation}
The reference point $\vomega_0$ is arbitrary, but in practice it is usually the initialization point of the algorithm. $R$ has to be big enough to ensure that $\Omega_R$ contains a solution. $\Err_R$ measures how much~\eqref{eq:VI_problem_strong} is violated on the restriction $\Omega_R$.
Such merit function is standard in the variational inequality literature. A similar one is used in \citep{nemirovski_prox-method_2004, juditsky2011solving}.
When $F$ is derived from the gradients~\eqref{eq:SP_stationnary_conditions} of a zero-sum game, we can define a more interpretable merit function. One has to be careful though when extending properties from the minimization setting to the saddle point setting (e.g. the merit function used by \citet{yadav_stabilizing_2017} is vacuous for a bilinear game as explained in App~\ref{sub:previous_criteria}).

In the appendix, we adopt a set of assumptions a little more general than the one in the main paper:

\begin{assumption}\label{assum:monotone_bounded_2}

\begin{itemize}
  \item $F$ is \emph{monotone} and $\Omega$ is convex and closed.
  \item $R$ is set big enough such that $R > \|\vomega_0-\vomega^*\|$ and $F$ is a \emph{monotone} operator.
\end{itemize}
\end{assumption}
Contrary to Assumption~\ref{assum:monotone_bounded}, in Assumption~\ref{assum:monotone_bounded_2} the constraint set in no longer assumed to be bounded. Assumption~\ref{assum:monotone_bounded_2} is implied by Assumption~\ref{assum:monotone_bounded} by setting $R$ to the diameter of $\Omega$, and is thus more general.

\subsection{More general merit functions} \label{sub:another_merit_function_for_the_saddle_point_problem}

In this appendix, we will note $\Err_R^{(\textup{VI})}$ the \emph{restricted merit function} defined in~\eqref{eq:merit_VI}. Let us recall its definition,
\begin{equation}\label{eq:app_merit_VI}
  \Err_R^{(\textup{VI})}(\vomega_t) \defas \max_{\vomega\in \Omega, \|\vomega-\vomega_0\|\leq R} F(\vomega)^\top(\vomega_t-\vomega) \,.
\end{equation}

When the objective is a saddle point problem i.e.,
\begin{equation}
   F(\vtheta,\vphi) = [\nabla_\vtheta \LL(\vtheta,\vphi) \,\, -\! \!\nabla_\vphi \LL(\vtheta,\vphi)]^\top
 \end{equation}
and $\LL$ is \emph{convex-concave} (see Definition~\ref{def:convex-concave} in \S\ref{app:definitions}), we can use another merit function than~\eqref{eq:app_merit_VI} on~$\Omega_R$ that is more interpretable and more directly related to the cost function of the minimax formulation:
\begin{equation}\label{eq:merit_SP}
  \Err_R^{(\text{SP})}(\vtheta_t,\vphi_t) \defas \!\!\!\!\!\!\!\! \max_{\substack{\vphi \in \Phi, \vtheta \in \Theta \\ \|(\vtheta,\vphi) - (\vtheta_0,\vphi_0)\|\leq R}} \!\!\!\!\!\!\! \LL(\vtheta_t, \vphi) -  \LL(\vtheta,\vphi_t) \,.
\end{equation}
In particular, if the equilibrium $(\vtheta^*,\vphi^*) \in \Omega^*\cap\Omega_R$ and we have that $\LL(\cdot,\vphi^*)$ and $-\LL(\vtheta^*,\cdot)$ are $\mu$-strongly convex (see \S\ref{app:definitions}), then the merit function for saddle points upper bounds the distance for $(\vtheta,\vphi) \in \Omega_R$ to the equilibrium as:
\begin{equation}
  \Err_R^{(\text{SP})}(\vtheta,\vphi) \geq \frac{\mu}{2} (\|\vtheta-\vtheta^*\|_2^2 + \|\vphi - \vphi^*\|_2^2) \,.
\end{equation}
In the appendix, we provide our convergence results with the merit functions~\eqref{eq:app_merit_VI} and~\eqref{eq:merit_SP}, depending on the setup:
\begin{equation}\label{eq:general_merit}
  \Err_R (\vomega) \!\defas\!
  \left\{
  \begin{aligned}
  \Err_R^{(\text{SP})} (\vomega) \;\; & \text{if $F$ is a SP operator~\eqref{eq:SP_stationnary_conditions}} \\
  \Err_R^{(\textup{VI})} (\vomega) \;\; & \text{otherwise.}
  \end{aligned}
  \right.
\end{equation}

\subsection{On the importance of the merit function} \label{sub:previous_criteria}
In this section, we illustrate the fact that one has to be careful when extending results and properties from the minimization setting to the minimax setting (and consequently to the variational inequality setting).
Another candidate as a merit function for saddle point optimization would be to naturally extend the suboptimality $f(\vomega) - f(\vomega^*)$ used in standard minimization (i.e. find $\vomega^*$ the minimizer of~$f$) to the gap $P(\vtheta,\vphi) = \LL(\vtheta,\vphi^*) - \LL(\vtheta^*,\vphi)$.
In a previous analysis of a modification of the stochastic gradient descent (SGD) method for GANs, \citet{yadav_stabilizing_2017} gave their convergence rate on $P$ that they called the ``primal-dual`` gap. Unfortunately, if we do not assume that the function $\LL$ is strongly convex-concave (a stronger assumption defined in \S\ref{app:definitions} and which fails for bilinear objective e.g.), $P$ may not be a \emph{merit function}. It can be 0 for a non optimal point, see for instance the discussion on the differences between~\eqref{eq:merit_SP} and $P$ in~\citep[Section~3]{gidel2017frank}.
In particular, for the simple 2D bilinear example $\LL(\vtheta,\vphi) = \vtheta \cdot \vphi$, we have that $\vtheta^* = \vphi^* =0$ and thus $P(\vtheta,\vphi) = 0\,\; \; \forall \vtheta,\vphi\,$.
\subsection{Variational inequalities for non-convex cost functions} \label{sub:variational_inequalities_for_non_convex_cost_functions}
When the cost functions defined in~\eqref{eq:two_player_games} are non-convex, the operator $F$ is no longer monotone. Nevertheless,~\eqref{eq:VI_problem_weak} and~\eqref{eq:VI_problem_strong} can still be defined, though a solution to~\eqref{eq:VI_problem_strong} is less likely to exist.
We note that~\eqref{eq:VI_problem_weak} is a local condition for $F$ (as only evaluating $F$ at the points $\vomega^*$). On the other hand, an appealing property of~\eqref{eq:VI_problem_strong} is that it is a global condition. In the context of minimization of a function $f$ for example (where $F = \nabla f$), if $\vomega^*$ solves~\eqref{eq:VI_problem_strong} then $\vomega^*$ is a \emph{global} minimum of~$f$ (and not just a stationary point for the solution of~\eqref{eq:VI_problem_strong}; see Proposition 2.2 from~\citet{crespi2005MVI}).

A less restrictive way to consider variational inequalities in the non-monotone setting is to use a local version of~\eqref{eq:VI_problem_strong}. If the cost functions are locally convex around the optimal couple $(\vtheta^*,\vphi^*)$ and if our iterates eventually fall and stay into that neighborhood, then we can consider our restricted merit function $\Err_R(\cdot)$ with a well suited constant $R$ and  apply our convergence results for monotone operators.

\section{Another way of implementing extrapolation to SGD} \label{sec:another_extension_of_sem}
We now introduce another way to combine extrapolation and SGD. This extension is very similar to AvgExtraSGD Alg.~\ref{alg:AvgExtraSGD}, the only difference is that it re-uses the mini-batch sample of the extrapolation step for the update of the current point. The intuition is that it correlates the estimator of the gradient of the extrapolation step and the one of the update step leading to a better correction of the oscillations which are also due to the stochasticity. One emerging issue (for the analysis) of this method is that since $\vomega'_t$ depend on $\xi_t$, the quantity $F(\vomega'_t,\xi_{t})$ is a biased estimator of $F(\vomega'_t)$.
\begin{algorithm}[H]
\caption{Re-used mini-batches for stochastic extrapolation (ReExtraSGD)}\label{alg:ReAvgExtraSGD}
\begin{algorithmic}[1]
  \STATE Let $\vomega_0 \in \Omega$
  \FOR{$t=0 \ldots T-1$}
  \STATE Sample $\xi_t \sim P$
  \STATE $\vomega'_t \defas P_\Omega[\vomega_t - \eta_t F(\vomega_t,\xi_t)]$  \hfill $\triangleright$ Extrapolation step
  \STATE $\vomega_{t+1} \defas P_{\Omega}[\vomega_t - \eta_t F\big( \vomega'_t,\xi_t\big)]$ \hfill $\triangleright$ Update step with the \textbf{same} sample
  \ENDFOR
  \STATE Return $\bar \vomega_{T} = \sum_{t=0}^{T-1}\eta_t \vomega'_t/ \sum_{t=0}^{T-1}\eta_t$
\end{algorithmic}
\end{algorithm}

\begin{theorem} \label{thm:NSEGM}
Assume that $\|\vomega'_t - \vomega_0\|\leq R, \, \forall t\geq 0$ where $(\vomega'_t)_{t \geq 0}$ are the iterates of Alg.~\ref{alg:ReAvgExtraSGD}. Under Assumption~\ref{assum:var_bounded} and~\ref{assum:monotone_bounded_2},
for any $T\geq 1$, Alg.~\ref{alg:ReAvgExtraSGD} with constant step-size $\eta \leq \frac{1}{\sqrt{2}L}$ has the following convergence properties:
\begin{equation*}
    \EE[\Err_R(\bar \vomega_T)] \leq \frac{R^2}{\eta T} + \eta\frac{\sigma^2 + 4L^2(4R^2+\sigma^2)}{2}
    \quad \text{where} \quad
    \bar \vomega_T \defas \frac{1}{T}\sum_{t=0}^{T-1} \vomega'_t \,.
\end{equation*}
Particularly,  $\eta_t = \frac{\eta}{\sqrt{T}}$ gives $\EE[\Err_R(\bar \vomega_T)] \leq \frac{O(1)}{\sqrt{T}}$.
\end{theorem}
The assumption that the sequence of the iterates provided by the algorithm is bounded is strong, but has also been made for instance in~\citep{yadav_stabilizing_2017}. The proof of this result is provided in \S\ref{app:proof_of_thm_thm:}.

\section{Variance comparison between AvgSGD and SGD with prediction method}
\label{app:SGDPreComparison}
To compare the variance term of AvgSGD in~\eqref{eq:thm_1_variance_term} with the one of the \emph{SGD with prediction method}~\citep{yadav_stabilizing_2017}, we need to have the same convergence certificate. Fortunately, their proof can be adapted to our convergence criterion (using Lemma~\ref{lemma:fictive_iterate} in \S\ref{app:proof_of_thm_thm:}), revealing an extra $\sigma^2/2$ in the variance term from their paper. The resulting variance can be summarized with our notation as
$(M^2(1 + L) + \sigma^2)/2$ where the $L$ is the Lipschitz constant of the operator $F$. Since $M \gg \sigma$, their variance term is then $1+L$ time larger than the one provided by the AvgSGD method.

\section{Proof of Theorems} \label{app:proof_of_thm_thm:}
This section is dedicated on the proof of the theorems provided in this paper in a slightly more general form working with the merit function defined in~\eqref{eq:general_merit}. First we prove an additional lemma necessary to the proof of our theorems.
 \begin{lemma}\label{lemma:fictive_iterate}Let $F$ be a monotone operator and let $(\vomega_t),(\vomega'_t),(\zz_t),(\Delta_t),(\xi_t)$ and $(\zeta_t)$ be six random sequences such that, for all $t\geq 0$
     \begin{equation*}\label{eq:proof_lemma}
      2\eta_tF(\vomega'_t)^\top (\vomega'_t-\vomega) \leq N_t - N_{t+1} + \eta_t^2(M_1(\vomega_t,\xi_t)+M_2(\vomega'_t,\zeta_t)) + 2 \eta_t \Delta_t^\top (\zz_t -\vomega)\,,
    \end{equation*}
    where $N_t = N(\vomega_t,\vomega'_{t-1},\vomega'_{t-2}) \geq 0$ and we extend $(\vomega'_t)$ with $ \vomega'_{-2} = \vomega'_{-1} = \vomega'_0$. Let also assume that
    with $N_0 \leq R \,,\;\EE[\|\Delta_t\|_2^2] \leq \sigma^2, \;\EE[\Delta_t |\zz_t,\Delta_0,\ldots,\Delta_{t-1} ] = 0\,, \;\EE[M_1(\vomega_t,\xi_t)] \leq M_1 $ and $ \,\EE[M_2(\vomega'_t,\zeta_t)] \leq M_2\,,$ then,
    \begin{equation}
      \EE[\Err_R(\bar \vomega_T)]
   \leq  \frac{R^2}{S_T} + \frac{M_1 + M_2 +\sigma^2}{2S_T} \sum_{t=0}^{T-1} \eta_t^2
    \end{equation}
    where $\bar \vomega_T \defas \sum_{t=0}^{T-1} \eta_t\vomega'_t / S_T$ and $S_T \defas \sum_{t=0}^{T-1} \eta_t$.
 \end{lemma}
 \proof[\textbf{Proof of Lemma~\ref{lemma:fictive_iterate}}] We sum~\eqref{eq:proof_lemma} for $0 \leq t \leq T-1$ to get,
\begin{align}
 \label{eq:sum_proof_lemma}
   &2 \sum_{t=0}^{T-1} \eta_t F(\vomega'_t)^\top (\vomega'_t-\vomega)
   \leq \notag\\
   &\qquad \qquad \sum_{t=0}^{T-1}\left[ (N_t-N_{t+1}) + \eta_t^2((M_1(\vomega_t,\xi_t)+M_2(\vomega'_t,\zeta_t))  +2 \eta_t \Delta_t^\top (\zz_t -\vomega) \right] \,.
\end{align}
We will then upper bound each sum in the right-hand side,
\begin{align*}
   \Delta_t^\top (\zz_t -\vomega)
  =  \Delta_t^\top (\zz_t -\uu_t) +
   \Delta_t^\top (\uu_t -\vomega)
\end{align*}
where $\uu_{t+1} \defas P_\Omega( \uu_t - \eta_t\Delta_t)$ and $\uu_0 \defas \vomega_0$. Then,
\begin{align*}
  \|\uu_{t+1}-\vomega\|_2^2 \leq \|\uu_t-\vomega\|_2^2- 2 \eta_t \Delta_t^\top (\uu_t -\vomega) +\eta_t^2 \|\Delta_t\|_2^2
\end{align*}
leading to
\begin{equation}
  2 \eta_t\Delta_t^\top (\zz_t -\vomega)
  \leq 2 \eta_t \Delta_t^\top (\zz_t -\uu_t) +
    \|\uu_t-\vomega\|_2^2 - \|\uu_{t+1}-\vomega\|_2^2 + \eta_t^2 \|\Delta_t\|_2^2
\end{equation}
Then noticing that $\zz_0 \defas \vomega_0$, back to \eqref{eq:sum_proof_lemma} we get a telescoping sum,
\begin{equation}
   2 \sum_{t=0}^{T-1} \eta_t F(\vomega'_t)^\top (\vomega'_t-\vomega)
   \leq
    2N_0 + \sum_{t=0}^{T-1}\big[ \eta_t^2 ((M_1(\vomega_t,\xi_t)+M_2(\vomega'_t,\zeta_t)) + \|\Delta_t\|_2^2) + 2 \eta_t \Delta_t^\top (\zz_t -\uu_t)\big] \,.
\end{equation}
If $F$ is the operator of a convex-concave saddle point~\eqref{eq:SP_stationnary_conditions}, we get, with $\vomega'_t = (\vtheta_t,\vphi_t)$
\begin{align*}
  F(\vomega'_t)^\top (\vomega'_t-\vomega)
  &\geq \nabla_\vtheta \LL(\vtheta_t,\vphi_t)^\top (\vtheta_t - \vtheta) - \nabla_\vphi \LL(\vtheta_t,\vphi_t)^\top (\vphi_t - \vphi) \\
  &\geq \LL(\vtheta_t,\vphi) - \LL(\vtheta_t,\vphi_t) + \LL(\vtheta_t,\vphi_t) - \LL(\vtheta,\vphi_t) \\
  & \qquad (\text{by convexity and concavity}) \\
  & = \LL(\vtheta_t,\vphi) - \LL(\vtheta,\vphi_t)
\end{align*}
then by convexity of $\LL(\cdot,\vphi)$ and concavity of $\LL(\vtheta,\cdot)$, we have that,
\begin{equation}
  2 S_T \sum_{t=0}^{T-1} \frac{\eta_t}{S_T} F(\vomega'_t)^\top (\vomega'_t-\vomega)
   \geq
   2 S_T \sum_{t=0}^{T-1} \frac{\eta_t}{S_T} (\LL(\vtheta_t,\vphi) - \LL(\vtheta,\vphi_t))
   \geq
   \bar2 S_T(\LL(\bar \vtheta_t,\vphi) - \LL(\bar\vtheta,\vphi_t))
\end{equation}
Otherwise if the operator $F$ is just monotone since $F(\vomega'_t)^\top (\vomega'_t-\vomega)\geq F(\vomega')^\top (\vomega'_t-\vomega)  $ we have that
\begin{equation}
    2 S_T \sum_{t=0}^{T-1} \eta_t F(\vomega'_t)^\top (\vomega'_t-\vomega)
    \geq  2 S_T \sum_{t=0}^{T-1} \eta_t F(\vomega')^\top (\vomega'_t-\vomega)
     =  2 S_T F(\vomega')^\top (\bar \vomega_t-\vomega)
\end{equation}
In both cases, we can now maximize the left hand side respect to $\vomega$ (since the RHS does not depend on $\vomega$) to get,
\begin{equation}
   2 S_T \Err_R(\bar \vomega_t)
   \leq
    2R^2 + \sum_{t=0}^{T-1}\big[ \eta_t^2 ((M_1(\vomega_t,\xi_t)+M_2(\vomega'_t,\zeta_t)) + \|\Delta_t\|_2^2) + 2 \eta_t \Delta_t^\top (\zz_t -\uu_t)\big] \,.
\end{equation}
Then taking the expectation, since $\EE[\Delta_t|\zz_t,\uu_t] =\EE[\Delta_t|\zz_t,\Delta_0,\ldots,\Delta_{t-1}]  = 0$,
$\EE_{\zeta_t}[\|\Delta_t\|_2^2] \leq \sigma^2, \;\EE_{\xi_t}[M_1(\vomega_t,\xi_t)] \leq M_1 $ and $ \,\EE_{\zeta_t}[M_2(\vomega'_t,\zeta_t)] \leq M_2\,,$ we get that,
\begin{equation}
      \EE[\Err_R(\bar \vomega_T)]
   \leq  \frac{R^2}{S_T} + \frac{M_1 + M_2 +\sigma^2}{2S_T} \sum_{t=0}^{T-1} \eta_t^2
\end{equation}
 \endproof

\subsection{Proof of Thm.~\ref{thm:SGM} } \label{sub:_thm:sgm_}
First let us state Theorem~\ref{thm:SGM} in its general form,
\begin{reptheorem}{thm:SGM}
 Under Assumption~\ref{assum:var_bounded},~\ref{assum:sub_bounded} and~\ref{assum:monotone_bounded_2}, Alg.~\ref{alg:AvgSGD} with constant step-size $\eta$ has the following convergence rate for all $T \geq 1$,
\begin{equation}
  \EE[\Err_R(\bar \vomega_T)] \leq \frac{R^2}{2\eta T} + \eta \frac{M^2 +\sigma^2}{2}  \quad \text{where} \quad \bar \vomega_T \defas \frac{1}{T}\sum_{t=0}^{T-1} \vomega_t \,.
\end{equation}
Particularly, $\eta = \frac{R}{\sqrt{T(M^2+\sigma^2)}}$ gives $\EE[\Err_R(\bar \vomega_T)] \leq \frac{R \sqrt{M^2 + \sigma^2}}{\sqrt{T}} \,.$
\end{reptheorem}
\proof[\textbf{Proof of Theorem~\ref{thm:SGM}}] Let any $\vomega \in \Omega$ such that $\|\vomega_0 - \vomega\|_2 \leq R$,
\begin{align*}
  \|\vomega_{t+1}- \vomega\|^2_2
  &= \| P_\Omega(\vomega_t - \eta_t F(\vomega_t,\xi_t)) - \vomega\|_2^2 \\
  & \leq \|\vomega_t - \eta_t F(\vomega_t,\xi_t)) - \vomega\|_2^2  \\
  & \quad (\text{projections are non-contractive, Lemma~\ref{lem:proj_non_contr}})  \\
  &= \|\vomega_{t}- \vomega\|^2_2 -2\eta_tF(\vomega_t,\xi_t)^\top(\vomega_t-\vomega) + \|\eta_t F(\vomega_t,\xi_t)\|^2_2
\end{align*}
Then we can make appear the quantity $F(\vomega_t)^\top (\vomega_t-\vomega)$ on the left-hand side,
\begin{equation}\label{eq:proof_2}
  2\eta_tF(\vomega_t)^\top (\vomega_t-\vomega) \leq \|\vomega_{t}- \vomega\|^2_2 - \|\vomega_{t+1}-\vomega\|^2_2 + \eta_t^2\|F(\vomega_t,\xi_t)\|^2_2 + 2 \eta_t (F(\vomega_t)-F(\vomega_t,\xi_t))^\top (\vomega_t -\vomega)
\end{equation}
 we can sum~\eqref{eq:proof_2} for $0\leq t \leq T-1$ to get,
\begin{multline} \label{eq:proof_sGM}
  2\sum_{t=0}^{T-1} \eta_t F(\vomega_t)^\top (\vomega_t-\vomega)
  \leq \\
  \sum_{t=0}^{T-1}\left[ (\|\vomega_t-\vomega\|^2-\|\vomega_{t+1}-\vomega\|^2) + \eta^2_t \|F(\vomega_t,\xi_t)\|^2_2 +2 \eta_t \Delta_t^\top (\vomega_t -\vomega) \right]
\end{multline}
where we noted $\Delta_t \defas F(\vomega_t)-F(\vomega_t,\xi_t)$.\\
By monotonicity, $F(\vomega_t)^\top(\vomega_t-\vomega) \geq F(\vomega)^\top(\vomega_t-\vomega) $ we get,
\begin{equation}\label{eq:proof_1}
   2 S_T F(\vomega)^\top (\bar \vomega_T-\vomega)
   \leq
   \sum_{t=0}^{T-1}\left[ (\|\vomega_t-\vomega\|^2-\|\vomega_{t+1}-\vomega\|^2) + \eta_t^2 \|F(\vomega_t,\xi_t)\|^2_2 +2 \eta_t \Delta_t^\top (\vomega_t -\vomega) \right] \,,
\end{equation}
where $S_T \defas \sum_{t=0}^{T-1}\eta_t$ and $\bar{\vomega}_T \defas \frac{1}{S_T} \sum_{t=0}^{T-1} \eta_t \vomega_t$. \\ We will then upper bound each sum in the right hand side,
\begin{align*}
   \Delta_t^\top (\vomega_t -\vomega)
  =  \Delta_t^\top (\vomega_t -\uu_t) +
   \Delta_t^\top (\uu_t -\vomega)
\end{align*}
where $\uu_{t+1} \defas P_\Omega( \uu_t - \eta_t\Delta_t)$ and $\uu_0 = \vomega_0$. Then,
\begin{align*}
  \|\uu_{t+1}-\vomega\|_2^2 \leq \|\uu_t-\vomega\|_2^2 - 2 \eta_t \Delta_t^\top (\uu_t -\vomega) +\eta_t^2 \|\Delta_t\|_2^2
\end{align*}
leading to
\begin{equation}
  2 \eta_t\Delta_t^\top (\vomega_t -\vomega)
  \leq 2 \eta_t \Delta_t^\top (\vomega_t -\uu_t) +
    \|\uu_t-\vomega\|_2^2 - \|\uu_{t+1}-\vomega\|_2^2 + \eta_t^2 \|\Delta_t\|_2^2
\end{equation}
Then noticing that $\uu_0 \defas \vomega_0$, back to \eqref{eq:proof_1} we get a telescoping sum,
\begin{align*}
   2S_T F(\vomega)^\top (\bar \vomega_T-\vomega)
   &\leq
    2\|\vomega_0-\vomega\|^2 + \sum_{t=0}^{T-1} \eta_t^2 (\|F(\vomega_t,\xi_t)\|^2_2 + \|\Delta_t\|_2^2) + 2 \sum_{t=0}^{T-1} \eta_t \Delta_t^\top (\vomega_t -\uu_t) \\
   &\leq  2R + \sum_{t=0}^{T-1} \eta_t^2 (\|F(\vomega_t,\xi_t)\|^2_2 + \|\Delta_t\|_2^2) + 2 \sum_{t=0}^{T-1} \eta_t \Delta_t^\top (\vomega_t -\uu_t)
\end{align*}
Then the right hand side does not depends on $\vomega$, we can maximize over $\vomega$ to get,
\begin{equation}
  2S_T \Err_R(\bar \vomega_T)
   \leq  2R + \sum_{t=0}^{T-1} \eta_t^2 (\|F(\vomega_t,\xi_t)\|^2_2 + \|\Delta_t\|_2^2) + 2 \sum_{t=0}^{T-1} \eta_t \Delta_t^\top (\vomega_t -\uu_t)
\end{equation}
Noticing that $\EE[\Delta_t|\vomega_t,\uu_t] = 0$ (the estimates of $F$ are unbiased), by Assumption~\ref{assum:sub_bounded} $\EE[(\|F(\vomega_t,\xi_t)\|^2_2 ] \leq M^2$ and by Assumption~\ref{assum:var_bounded} $\EE[\|\Delta_t\|_2^2] \leq \sigma^2$ we get,
\begin{equation}
  \EE[\Err_R(\bar \vomega_T)]
   \leq  \frac{R}{S_T} + \frac{M^2 +\sigma^2}{2S_T} \sum_{t=0}^{T-1} \eta_t^2
\end{equation}
particularly for $\eta_t = \eta$ and $\eta_t = \frac{\eta}{\sqrt{t+1}}$ we respectively get,
\begin{equation}
  \EE[\Err_R(\bar \vomega_T)]
   \leq  \frac{2R}{\eta T} + \frac{\eta}{2}(M^2 +\sigma^2)
\end{equation}
and
\begin{equation}
   \EE[\Err_R(\bar \vomega_T)]
   \leq  \frac{4R}{\eta\sqrt{T+1}-1} + 2 \eta \ln(T+1)\frac{M^2 +\sigma^2}{\sqrt{T+1}-1}
\end{equation}
 \endproof

\subsection{Proof of Thm.~\ref{thm:SEGM}} \label{sub:proof_of_thm_thm:segm}
\begin{reptheorem}{thm:SEGM}
Under Assumption~\ref{assum:var_bounded} and~\ref{assum:monotone_bounded_2}, if $\EE_\xi[F]$ is $L$-Lipschitz, then Alg.~\ref{alg:AvgExtraSGD} with a constant step-size $\eta \leq \frac{1}{\sqrt{3}L}$ has the following convergence rate for any $T\geq 1$,
\begin{equation}\label{eq:thm_2_variance_term:app}
  \EE[\Err_R(\bar \vomega_T)] \leq \frac{R^2}{\eta T} + \frac{7}{2} \eta \sigma^2
  \quad \text{where} \quad
  \bar \vomega_T \defas \frac{1}{T}\sum_{t=0}^{T-1} \vomega'_t \,.
\end{equation}
Particularly, $\eta = \frac{\sqrt{2}R}{ \sigma \sqrt{7T}} $ gives $\EE[\Err_R(\bar \vomega_T)]\leq \frac{\sqrt{14}R\sigma}{\sqrt{T}}$.

\end{reptheorem}
\proof[\textbf{Proof of Thm.~\ref{thm:SEGM}}] Let any $\vomega \in \Omega$ such that $\|\vomega_0 - \vomega\|_2 \leq R$. Then, the update rules become $\vomega_{t+1} = P_\Omega(\vomega_t - \eta_t F(\vomega'_t,\zeta_t))$ and $\vomega'_t = P_\Omega(\vomega_t - \eta F(\vomega_t,\xi_t))$.
We start by applying Lemma~\ref{lemma:ineg} for $(\vomega,\uu,\vomega',\vomega^+) = (\vomega_t,-\eta F(\vomega'_t,\zeta_t),\vomega,\vomega_{t+1})$ and $(\vomega,\uu,\vomega',\vomega^+) = (\vomega_t,-\eta_tF(\vomega_t,\xi_t),\vomega_{t+1},\vomega'_t)$,
\begin{align*}
  \|\vomega_{t+1}- \vomega\|^2_2
  & \leq \|\vomega_t - \vomega\|_2^2 - 2\eta_tF(\vomega'_t,\zeta_t)^\top(\vomega_{t+1}-\vomega) -\|\vomega_{t+1} - \vomega_t\|_2^2 \\
  \|\vomega'_{t}-\vomega_{t+1}\|_2^2
  & \leq  \|\vomega_t - \vomega_{t+1}\|_2^2 - 2\eta_tF(\vomega_t,\xi_t)^\top(\vomega'_{t} -\vomega_{t+1}) -\|\vomega'_t- \vomega_{t}\|_2^2
\end{align*}
Then, summing them we get
\begin{multline}
  \|\vomega_{t+1}- \vomega\|^2_2
  \leq \|\vomega_t - \vomega\|_2^2 - 2\eta_tF(\vomega'_t,\zeta_t)^\top(\vomega_{t+1}-\vomega) \\ - 2\eta_tF(\vomega_t,\xi_t)^\top(\vomega'_{t} -\vomega_{t+1}) -\|\vomega_{t} - \vomega'_t\|_2^2 - \|\vomega_{t+1} - \vomega'_{t}\|_2^2
\end{multline}
leading to
\begin{multline*}
  \|\vomega_{t+1}- \vomega\|^2_2
  \leq \|\vomega_t - \vomega\|_2^2 - 2\eta_tF(\vomega'_t,\zeta_t)^\top(\vomega'_{t}-\vomega) \\ + 2\eta_t(F(\vomega'_t,\zeta_t) - F(\vomega_t,\xi_t))^\top(\vomega'_{t} -\vomega_{t+1}) -\|\vomega_{t} - \vomega'_t\|_2^2 - \|\vomega_{t+1} - \vomega'_{t}\|_2^2
\end{multline*}
Then with $2\bm{a}^\top \bm{b} \leq \|\bm{a}\|_2^2 + \|\bm{b}\|^2_2 $ we get
\begin{align*}
  \|\vomega_{t+1}- \vomega\|^2_2
  &\leq  \|\vomega_{t}- \vomega\|^2_2 -2\eta_tF(\vomega'_t,\zeta_t)^\top(\vomega'_t-\vomega) \\ & \quad - \|\vomega_t - \vomega'_t\|^2_2 + \eta_t^2\| F(\vomega'_t,\zeta_t) -F(\vomega_{t},\xi_{t})\|_2^2
\end{align*}
Using the inequality $\|\bm{a} + \bm{b} + \bm{c}\|_2^2 \leq 3(\|\bm{a}\|^2_2 + \|\bm{b}\|_2^2 + \|\bm{c}\|_2^2)$ we get,
\begin{align*}
  \|\vomega_{t+1}- \vomega\|^2_2
    &\leq  \|\vomega_{t}- \vomega\|^2_2 -2\eta_t F(\vomega'_t,\zeta_t)^\top(\vomega'_t-\vomega) - \|\vomega_t - \vomega'_t\|^2_2 \\
  & \quad + 3 \eta_t^2 (\|F(\vomega_t)-F(\vomega_t,\xi_t)\|_2^2 + \|F(\vomega'_t)-F(\vomega'_t,\zeta_t)\|_2^2 + \|F(\vomega'_t) -F(\vomega_t)\|_2^2)
\end{align*}
Then we can use the $L$-Lipschitzness of $F$ to get,
\begin{align*}
  \|\vomega_{t+1}- \vomega\|^2_2
    &\leq  \|\vomega_{t}- \vomega\|^2_2 -2\eta_t F(\vomega'_t,\zeta_t)^\top(\vomega'_t-\vomega) - \|\vomega_t - \vomega'_t\|^2_2 \\
  & \quad + 3 \eta_t^2 (\|F(\vomega_t)-F(\vomega_t,\xi_t)\|_2^2 + \|F(\vomega'_t)-F(\vomega'_t,\zeta_t)\|_2^2 + L^2\|\vomega_t -\vomega'_t\|_2^2)
\end{align*}
As we restricted the step-size to $\eta_t \leq \frac{1}{\sqrt{3}L}$ we get,
\begin{align*}
2\eta_t F(\vomega'_t)^\top(\vomega'_t-\vomega)
  & \leq \|\vomega_{t}- \vomega\|^2_2 - \|\vomega_{t+1}- \vomega\|^2_2 + 2\eta_t (F(\vomega'_t)- F(\vomega'_t,\zeta_t))^\top(\vomega'_t-\vomega) \\
  & \quad+ 3\eta_t^2 \|F(\vomega_t)-F(\vomega_t,\xi_t)\|_2^2 + 3 \eta_t^2\|F(\vomega'_t)-F(\vomega'_t,\zeta_t)\|_2^2
\end{align*}
We get a particular case of \eqref{eq:proof_lemma} so we can use Lemma~\ref{lemma:fictive_iterate} where $N_t = \|\vomega_t -\vomega\|_2^2,$ $M_1(\vomega_t,\xi_t) = 3\|F(\vomega_t)-F(\vomega_t,\xi_t)\|_2^2 \, , \, M_2(\vomega'_t,\zeta_t) = 3\|F(\vomega'_t)-F(\vomega'_t,\zeta_t)\|_2^2, \, \Delta_t = F(\vomega'_t)- F(\vomega'_t,\zeta_t)$ and $\zz_t = \vomega'_t$.  By Assumption~\ref{assum:var_bounded}, $M_1 = M_2 = 3\sigma^2$ and by the fact that $\EE[F(\vomega'_t)- F(\vomega'_t,\zeta_t) \, |\vomega'_t,\Delta_0,\ldots,\Delta_{t-1}] = \EE[\EE[F(\vomega'_t)- F(\vomega'_t,\zeta_t) \, |\vomega'_t]|\Delta_0,\ldots,\Delta_{t-1}] =0$ the hypothesis of Lemma~\ref{lemma:fictive_iterate} hold and we get,
\begin{equation}
   \EE[\Err_R(\bar \vomega_T)]
   \leq  \frac{R^2}{S_T} + \frac{7\sigma^2}{2S_T} \sum_{t=0}^{T-1} \eta_t^2
\end{equation}
\endproof

\subsection{Proof of Thm.~\ref{thm:new_AvgExtraSGD}} \label{sub:proof_of_thm_thm:new_sem}

\begin{reptheorem}{thm:new_AvgExtraSGD}
Under Assumption~\ref{assum:var_bounded}, if $\EE_\xi[F]$ is $L$-Lipschitz, then AvgPastExtraSGD (Alg.~\ref{alg:AvgPastExtraSGD}) with a constant step-size $\eta \leq \frac{1}{2\sqrt{3}L}$ has the following convergence rate for any $T\geq 1$,
\begin{equation}\label{eq:thm_3_variance_term:app}
  \EE[\Err_R(\bar \vomega_T)] \leq \frac{R^2}{\eta T} + \frac{13}{2} \eta \sigma^2
  \quad \text{where} \quad
  \bar \vomega_T \defas \frac{1}{T}\sum_{t=0}^{T-1} \vomega'_t \,.
\end{equation}
Particularly, $\eta = \frac{\sqrt{2}R}{ \sigma \sqrt{13T}} $ gives $\EE[\Err_R(\bar \vomega_T)]\leq \frac{\sqrt{26}R\sigma}{\sqrt{T}}$.
\end{reptheorem}

First let us recall the update rule

\begin{equation}
  \left\{
  \begin{aligned}
  &\vomega_{t+1} = P_\Omega[\vomega_t - \eta_t F(\vomega'_t,\xi_t)] \\
  &\vomega'_{t+1} = P_\Omega[\vomega_{t+1} - \eta_{t+1} F(\vomega'_t,\xi_t)] \,.
  \end{aligned}
  \right.
\end{equation}

\begin{lemma}\label{lemma:3} We have for any $\vomega \in \Omega$,
\begin{align}
  2 \eta F(\vomega'_t,\xi_t)^\top(\vomega'_t-\vomega)
  & \leq \|\vomega_t - \vomega\|_2^2 -\|\vomega_{t+1}-\vomega\|_2^2   - \|\vomega'_t -\vomega_t\|_2^2
  + 3 \eta_t^2 L^2 \|\vomega'_{t-1}-\vomega'_t\|_2^2 \notag \\
  & \quad  +  3\eta_t^2 \big[\|F(\vomega'_{t-1},\xi_{t-1})-F(\vomega'_{t-1})\|_2^2 + \|F(\vomega'_{t})-F(\vomega'_t,\xi_t)\|_2^2  \big]\, .
\end{align}
\end{lemma}
\proof
Applying Lemma~\ref{lemma:ineg} for $(\vomega,\uu,\vomega^+,\vomega') = (\vomega_t,-\eta_t F(\vomega'_t,\xi_{t}),\vomega_{t+1},\vomega)$ and $(\vomega,\uu,\vomega^+,\vomega') = (\vomega_{t},-\eta_t F(\vomega'_{t-1},\xi_{t-1}),\vomega'_{t},\vomega_{t+1})$, we get,
\begin{equation}
  \|\vomega_{t+1}-\vomega\|_2^2
  \leq \|\vomega_t - \vomega\|_2^2 - 2 \eta_t F(\vomega'_t,\xi_t)^\top(\vomega_{t+1}-\vomega) - \|\vomega_{t+1}-\vomega_t\|_2^2 \label{eq:proof_alg_v2.2_1}
\end{equation}
and
\begin{equation} \label{eq:proof_alg_v2.2_2}
  \|\vomega'_t -\vomega_{t+1}\|_2^2 \leq \|\vomega_t-\vomega_{t+1}\|_2^2 - 2 \eta_t F(\vomega'_{t-1},\xi_{t-1})^\top(\vomega'_t-\vomega_{t+1}) - \|\vomega'_t -\vomega_t\|_2^2\,.
\end{equation}
Summing~\eqref{eq:proof_alg_v2.2_1} and~\eqref{eq:proof_alg_v2.2_2} we get,
\begin{align}
  \|\vomega_{t+1}-\vomega\|_2^2
  &\leq \|\vomega_t - \vomega\|_2^2 - 2 \eta_t F(\vomega'_t,\xi_t)^\top(\vomega_{t+1}-\vomega)\\
  &\quad - 2 \eta_t F(\vomega'_{t-1},\xi_{t-1})^\top(\vomega'_t-\vomega_{t+1})  - \|\vomega'_t -\vomega_t\|_2^2 - \|\vomega'_t -\vomega_{t+1}\|_2^2 \\
  & =  \|\vomega_t - \vomega\|_2^2 - 2 \eta_t F(\vomega'_t,\xi_t)^\top(\vomega'_t-\vomega)  - \|\vomega'_t -\vomega_t\|_2^2 - \|\vomega'_t -\vomega_{t+1}\|_2^2\\
  &\quad - 2 \eta_t (F(\vomega'_{t-1},\xi_{t-1}) -F(\vomega'_t,\xi_t))^\top(\vomega'_t-\vomega_{t+1}) \, .
\end{align}
Then, we can use the inequality of arithmetic and geometric means $2a^\top b \leq \|a\|_2^2 + \|b\|_2^2$ to get,
\begin{align}
  \|\vomega_{t+1}-\vomega\|_2^2
  & \leq \|\vomega_t - \vomega\|_2^2 - 2 \eta_t F(\vomega'_t,\xi_t)^\top(\vomega'_t-\vomega) +  \eta_t^2 \|F(\vomega'_{t-1},\xi_{t-1})-F(\vomega'_t,\xi_t)\|_2^2 \notag\\
  & \quad + \|\vomega'_t-\vomega_{t+1}\|_2^2
  - \|\vomega'_t -\vomega_t\|_2^2 - \|\vomega'_t -\vomega_{t+1}\|_2^2 \\
  & = \|\vomega_t - \vomega\|_2^2 - 2 \eta_t F(\vomega'_t,\xi_t)^\top(\vomega'_t-\vomega)\\
  & \quad  +  \eta_t^2 \|F(\vomega'_{t-1},\xi_{t-1})-F(\vomega'_t,\xi_t)\|_2^2 - \|\vomega'_t -\vomega_t\|_2^2 \label{eq:proff_4_int_result} \, .
\end{align}
Using the inequality $\|\bm{a} + \bm{b} + \bm{c}\|_2^2 \leq 3(\|\bm{a}\|^2_2 + \|\bm{b}\|_2^2 + \|\bm{c}\|_2^2)$ we get,
\begin{align}
  \|F(\vomega'_{t-1},\xi_{t-1})-F(\vomega'_t,\xi_t)\|_2^2
  &\leq 3(\|F(\vomega'_{t-1},\xi_{t-1})-F(\vomega'_{t-1})\|_2^2 + \|F(\vomega'_{t-1})-F(\vomega'_t)\|_2^2 \notag \\
  & \quad + \|F(\vomega'_{t})-F(\vomega'_t,\xi_t)\|_2^2) \\
  & \leq 3(\|F(\vomega'_{t-1},\xi_{t-1})-F(\vomega'_{t-1})\|_2^2 + L^2 \|\vomega'_{t-1}-\vomega'_t\|_2^2 \notag \\
  & \quad + \|F(\vomega'_{t})-F(\vomega'_t,\xi_t)\|_2^2) \label{eq:variance_lip_decomp}\,,
\end{align}
where we used the $L$-Lipschitzness of $F$ for the last inequality.\\

Combining~\eqref{eq:proff_4_int_result} with~\eqref{eq:variance_lip_decomp} we get,
\begin{align}
  \|\vomega_{t+1}-\vomega\|_2^2
  & \leq \|\vomega_t - \vomega\|_2^2 - 2 \eta_t F(\vomega'_t,\xi_t)^\top(\vomega'_t-\vomega) - \|\vomega'_t -\vomega_t\|_2^2
  + 3 \eta_t^2 L^2 \|\vomega'_{t-1}-\vomega'_t\|_2^2 \notag \\
  & \quad  +  3\eta_t^2 \big[\|F(\vomega'_{t-1},\xi_{t-1})-F(\vomega'_{t-1})\|_2^2 + \|F(\vomega'_{t})-F(\vomega'_t,\xi_t)\|_2^2  \big]\, .
\end{align}
\endproof

\begin{lemma}\label{lemma:2} For all $t \geq 0$, if we set $\vomega'_{-2} = \vomega'_{-1} = \vomega'_0$ we have
\begin{align}
  \|\vomega'_{t-1}-\vomega'_t\|_2^2 &\leq 4  \|\vomega_t-\vomega'_t\|_2^2 + 12 \eta_{t-1}^2  \big( \|F(\vomega'_{t-1},\xi_{t-1})-F(\vomega'_{t-1})\|_2^2 + L^2 \|\vomega'_{t-1}-\vomega'_{t-2}\|_2^2 \notag  \\
  & \quad + \|F(\vomega'_{t-2})-F(\vomega'_{t-2},\xi_{t-2})\|_2^2)\big) - \|\vomega'_{t-1}-\vomega'_t\|_2^2 \,. \label{eq:proof_lemma_2_1}
\end{align}
\end{lemma}
\proof
We start with $\|a + b\|_2^2 \leq 2\|a\|^2 + 2 \|b\|^2$.
\begin{equation}
  \|\vomega'_{t-1}-\vomega'_t\|_2^2 \leq 2 \|\vomega_t-\vomega'_t\|_2^2 + 2 \|\vomega_t-\vomega'_{t-1}\|_2^2 \,.
\end{equation}
Moreover, since the projection is contractive we have that
\begin{align}
  \|\vomega_t-\vomega'_{t-1}\|_2^2
  &\leq \|\vomega_{t-1} - \eta_{t-1} F(\vomega'_{t-1},\xi_{t-1}) - \vomega_{t-1} - \eta_{t-1} F(\vomega'_{t-2},\xi_{t-2})\|_2^2 \\
  &= \eta_{t-1}^2 \| F(\vomega'_{t-1},\xi_{t-1})- F(\vomega'_{t-2},\xi_{t-2}) \|_2^2 \\
  &\leq 3\eta_{t-1}^2 \big( \|F(\vomega'_{t-1},\xi_{t-1})-F(\vomega'_{t-1})\|_2^2 + L^2 \|\vomega'_{t-1}-\vomega'_{t-2}\|_2^2 \notag \\
  & \quad + \|F(\vomega'_{t-2})-F(\vomega'_{t-2},\xi_{t-2})\|_2^2)\big) \,. \label{eq:x_y}
\end{align}
where in the last line we used the same inequality as in~\eqref{eq:variance_lip_decomp}.
 Combining~\eqref{eq:proof_lemma_2_1} and~\eqref{eq:x_y} we get,
\begin{align}
  \|\vomega'_{t-1}-\vomega'_t\|_2^2
  & = 2 \|\vomega'_{t-1}-\vomega'_t\|_2^2 - \|\vomega'_{t-1}-\vomega'_t\|_2^2 \\
  & \leq 4 \|\vomega_t-\vomega'_t\|_2^2 + 4 \|\vomega_t-\vomega'_{t-1}\|_2^2 - \|\vomega'_{t-1}-\vomega'_t\|_2^2 \\
  & \leq 4  \|\vomega_t-\vomega'_t\|_2^2 + 12 \eta_{t-1}^2 \big( \|F(\vomega'_{t-1},\xi_{t-1})-F(\vomega'_{t-1})\|_2^2 + L^2 \|\vomega'_{t-1}-\vomega'_{t-2}\|_2^2 \notag \\
  & \quad + \|F(\vomega'_{t-2})-F(\vomega'_{t-2},\xi_{t-2})\|_2^2)\big) - \|\vomega'_{t-1}-\vomega'_t\|_2^2 \,.
\end{align}
\endproof
\proof[\textbf{Proof of Theorem~\ref{thm:new_AvgExtraSGD}}]
Combining Lemma~\ref{lemma:2} and Lemma~\ref{lemma:3} we get,
\begin{align}
  2\eta_t F(\vomega'_t,\xi_t)^\top(\vomega'_t-\vomega)
  & \leq \|\vomega_t-\vomega\|_2^2 - \|\vomega_{t+1}-\vomega\|_2^2 \notag\\
  & \quad +  36\eta_t^2 \eta_{t-1}^2 L^2 \big( \|F(\vomega'_{t-1},\xi_{t-1})-F(\vomega'_{t-1})\|_2^2 + L^2 \|\vomega'_{t-1}-\vomega'_{t-2}\|_2^2 \notag \\
  & \qquad + \|F(\vomega'_{t-2})-F(\vomega'_{t-2},\xi_{t-2})\|_2^2\big) \notag \\
  & \quad - 3 \eta_t^2L^2 \|\vomega'_{t-1}-\vomega'_t\|_2^2  + (12 \eta_t^2L^2   - 1) \|\vomega'_t-\vomega_t\|_2^2 \notag \\
  & \quad +  3\eta_t^2 \big[\|F(\vomega'_{t-1},\xi_{t-1})-F(\vomega'_{t-1})\|_2^2 + \|F(\vomega'_{t})-F(\vomega'_t,\xi_t)\|_2^2 ] \,.
\end{align}
Then for $\eta_t \leq \frac{1}{2\sqrt{3}L}$ we have $36\eta_t^2 \eta_{t-1}^2 L^4\leq 3 \eta_{t-1}^2 L^2$,
\begin{align}
  2\eta_t F(\vomega'_t)^\top(\vomega'_t-\vomega)
  & \leq \|\vomega_t-\vomega\|_2^2 - \|\vomega_{t+1}-\vomega\|_2^2 \notag\\
  & \quad +  3L^2(\eta_{t-1}^2\|\vomega'_{t-1}-\vomega'_{t-2}\|_2^2 - \eta_t^2\|\vomega'_{t-1}-\vomega'_t\|_2^2) \notag \\
  & \quad + 2\eta_t (F(\vomega'_t)- F(\vomega'_t,\xi_t))^\top(\vomega'_t-\vomega)  \notag \\
  & \quad + 3 \eta_t^2 \big[|F(\vomega'_{t-2},\xi_{t-2})-F(\vomega'_{t-2})\|_2^2+ 2\|F(\vomega'_{t-1},\xi_{t-1})-F(\vomega'_{t-1})\|_2^2 \notag \\
  & \qquad + \|F(\vomega'_{t})-F(\vomega'_t,\xi_t)\|_2^2 ] \,.
\end{align}
We can then use Lemma~\ref{lemma:fictive_iterate} where
\begin{align*}
N_t
&= \|\vomega_t -\vomega\|_2^2 + 3L^3 \eta_{t-1} \|\vomega'_{t-1} - \vomega'_{t-2}\|_2^2,\\
M_1(\vomega_{t},\xi_{t})
&=  0  \\
M_2(\vomega'_t,\xi_t)
&= 3\|F(\vomega'_t)-F(\vomega'_t,\xi_t)\|_2^2 + 6\|F(\vomega'_{t-1})-F(\vomega'_{t-1},\xi_{t-1})\|_2^2 \\
& \quad+ 3\|F(\vomega'_{t-2})-F(\vomega'_{t-2},\xi_{t-2})\|_2^2 \\
 \Delta_t
&= F(\vomega'_t)- F(\vomega'_t,\xi_t) \\
\zz_t
&= \vomega'_t \,.
\end{align*}
By Assumption~\ref{assum:var_bounded}, $M_2 = 12\sigma^2$ and by the fact that $\EE[F(\vomega'_t)- F(\vomega'_t,\xi_t) \, |\vomega'_t,\Delta_0,\ldots,\Delta_{t-1}] = \EE[\EE[F(\vomega'_t)- F(\vomega'_t,\xi_t) \, |\vomega'_t]|\Delta_0,\ldots,\Delta_{t-1}] =0$ the hypothesis of Lemma~\ref{lemma:fictive_iterate} hold and we get,

\begin{equation}
   \EE[\Err_R(\bar \vomega_T)]
   \leq  \frac{R^2}{S_T} + \frac{13\sigma^2}{2S_T} \sum_{t=0}^{T-1} \eta_t^2
\end{equation}
\endproof

\subsection{Proof of Theorem~\ref{thm:NSEGM}} \label{sub:proof_of_theorem_thm:nsegm}

Theorem~\ref{thm:NSEGM} has been introduced in \S\ref{sec:another_extension_of_sem}. This theorem is about Algorithm~\ref{alg:ReAvgExtraSGD} which consists in another way to implement extrapolation to SGD. Let us first restate this theorem,
\begin{reptheorem}{thm:NSEGM}
Assume that $\|\vomega'_t - \vomega_0\|\leq R, \, \forall t\geq 0$ where $(\vomega'_t)_{t \geq 0}$ are the iterates of Alg.~\ref{alg:ReAvgExtraSGD}. Under Assumption~\ref{assum:var_bounded} and~\ref{assum:monotone_bounded_2},
for any $T\geq 1$, Alg.~\ref{alg:ReAvgExtraSGD} with constant step-size $\eta \leq \frac{1}{\sqrt{2}L}$ has the following convergence properties:
\begin{equation*}
    \EE[\Err_R(\bar \vomega_T)] \leq \frac{R^2}{\eta T} + \eta\frac{\sigma^2 + 4L^2(4R^2+\sigma^2)}{2}
    \quad \text{where} \quad
    \bar \vomega_T \defas \frac{1}{T}\sum_{t=0}^{T-1} \vomega'_t \,.
\end{equation*}
Particularly,  $\eta_t = \frac{\eta}{\sqrt{T}}$ gives $\EE[\Err_R(\bar \vomega_T)] \leq \frac{O(1)}{\sqrt{T}}$.
\end{reptheorem}

\proof[\textbf{Proof of Thm.~\ref{thm:NSEGM}}] Let any $\vomega \in \Omega$ such that $\|\vomega_0 - \vomega\|_2 \leq R$. Then, the update rules become $\vomega_{t+1} = P_\Omega(\vomega_t - \eta_t F(\vomega'_t,\xi_t))$ and $\vomega'_t = P_\Omega(\vomega_t - \eta F(\vomega_t,\xi_t))$.
We start the same way as the proof of Thm.~\ref{thm:SEGM} by applying Lemma~\ref{lemma:ineg} for $(\vomega,\uu,\vomega',\vomega^*) = (\vomega_t,-\eta F(\vomega'_t,\xi_t),\vomega,\vomega_{t+1})$ and $(\vomega,\uu,\vomega',\vomega^+) = (\vomega_t,-\eta_tF(\vomega_t,\xi_t),\vomega_{t+1},\vomega'_t)$,
\begin{align*}
  \|\vomega_{t+1}- \vomega\|^2_2
  & \leq \|\vomega_t - \vomega\|_2^2 - 2\eta_tF(\vomega'_t,\xi_t)^\top(\vomega_{t+1}-\vomega) -\|\vomega_{t+1} - \vomega_t\|_2^2 \\
  \|\vomega'_{t} - \vomega_{t+1}\|_2^2
  & \leq  \|\vomega_t - \vomega_{t+1}\|_2^2 - 2\eta_tF(\vomega_t,\xi_t)^\top(\vomega'_{t} -\vomega_{t+1}) -\|\vomega'_{t} - \vomega_t\|_2^2
\end{align*}
Then, summing them we get
\begin{multline}
  \|\vomega_{t+1}- \vomega\|^2_2
  \leq \|\vomega_t - \vomega\|_2^2 - 2\eta_tF(\vomega'_t,\xi_t)^\top(\vomega_{t+1}-\vomega) \\ - 2\eta_tF(\vomega_t,\xi_t)^\top(\vomega'_{t} -\vomega_{t+1}) -\|\vomega'_{t} - \vomega_t\|_2^2 - \|\vomega_{t+1} - \vomega'_{t}\|_2^2
\end{multline}
leading to
\begin{multline*}
  \|\vomega_{t+1}- \vomega\|^2_2
  \leq \|\vomega_t - \vomega\|_2^2 - 2\eta_tF(\vomega'_t,\xi_t)^\top(\vomega'_{t}-\vomega) \\+2\eta_t(F(\vomega'_t,\xi_t) - F(\vomega_t,\xi_t))^\top(\vomega'_{t} -\vomega_{t+1}) -\|\vomega'_{t} - \vomega_t\|_2^2 - \|\vomega_{t+1} - \vomega'_{t}\|_2^2
\end{multline*}
Then with $2\bm{a}^\top \bm{b} \leq \|\bm{a}\|_2^2 + \|\bm{b}\|^2_2 $ we get
\begin{align*}
  \|\vomega_{t+1}- \vomega\|^2_2
  &\leq  \|\vomega_{t}- \vomega\|^2_2 -2\eta_tF(\vomega'_t,\xi_t)^\top(\vomega'_t-\vomega) \\ + \eta_t^2\| F(\vomega'_t,\xi_t) -F(\vomega_{t},\xi_{t})\|_2^2 - \|\vomega'_t - \vomega_t\|^2_2
\end{align*}
Using the Lipschitz assumption we get
\begin{align*}
  \|\vomega_{t+1}- \vomega\|^2_2
    &\leq  \|\vomega_{t}- \vomega\|^2_2 -2\eta_t F(\vomega'_t,\xi_t)^\top(\vomega'_t-\vomega) + (\eta_t^2L^2-1) \|\vomega_t - \vomega'_t\|^2_2
\end{align*}
Then we add $2 \eta_tF(\vomega'_t)^\top (\vomega'_t-\vomega)$ in both sides to get,
\begin{multline}\label{eq:lemma:1}
  2 \eta_tF(\vomega'_t)^\top (\vomega'_t-\vomega)
    \leq  \|\vomega_{t}- \vomega\|^2_2 - \|\vomega_{t+1}- \vomega\|^2_2 \\ -2\eta_t (F(\vomega'_t,\xi_t)-F(\vomega'_t))^\top(\vomega'_t-\vomega) + (\eta_t^2L^2-1) \|\vomega_t - \vomega'_t\|^2_2
\end{multline}
Here, unfortunately we cannot use Lemma~\ref{lemma:fictive_iterate} because
$F(\vomega'_t,\xi_t)$ is biased.
We will then deal with the quantity $A=(F(\vomega'_t,\xi_t)-F(\vomega'))^\top(\vomega'_t-\vomega)$
. We have that,
\begin{align*}
  A
  & = (F(\vomega'_t,\xi_t)-F(\vomega_t,\xi_t))^\top(\vomega-\vomega'_t) + (F(\vomega_t)-F(\vomega'_t))^\top(\vomega-\vomega'_t) \\
  & \quad + (F(\vomega_t,\xi_t)-F(\vomega_t))^\top(\vomega_t -\vomega'_t) + (F(\vomega_t,\xi_t)-F(\vomega_t))^\top(\vomega-\vomega_t) \\
  & \leq 2L\|\vomega'_t-\vomega_t\|_2 \|\vomega'_t-\vomega\|_2 + \|F(\vomega_t,\xi_t)-F(\vomega_t)\|\|\vomega'_t-\vomega_t\|_2 \\
  & \quad +(F(\vomega_t,\xi_t)-F(\vomega_t))^\top(\vomega-\vomega_t) \\
  & \qquad (\text{Using Cauchy-Schwarz and the $L$-Lip of $F$})
  \end{align*}
Then using $2\|a\|\|b\| \leq \delta \|a\|_2^2 + \frac{1}{\delta} \|b\|_2^2$, for $\delta = 4$,
\begin{align*}
  -2\eta_t (F(\vomega'_t,\xi_t)-F(\vomega'_t))^\top(\vomega'_t-\vomega)
  &\leq \frac{1}{2}\|\vomega'_t-\vomega_t\|^2 + 8\eta_t^2L^2 \|\vomega'_t-\vomega\|_2^2 \\
  & \quad + 4 \eta_t^2\|F(\vomega_t,\xi_t)-F(\vomega_t)\|^2_2 \\& \quad+  \frac{1}{4}\|\vomega'_t-\vomega_t\|_2^2 + 2 \eta_t(F(\vomega_t,\xi_t)-F(\vomega_t))^\top(\vomega-\vomega_t)
\end{align*}
leading to,
\begin{align*}
  2\eta_tF(\vomega'_t)^\top (\vomega'_t-\vomega)
    &\leq  \|\vomega_{t}- \vomega\|^2_2 - \|\vomega_{t+1}- \vomega\|^2_2 + 2 \eta_t(F(\vomega_t,\xi_t)-F(\vomega_t))^\top(\vomega-\vomega_t) \\
    & \quad + (\eta_t^2L^2- \tfrac{1}{4}) \|\vomega_t - \vomega'_t\|^2_2 \\
    & \quad + 4\eta_t^2 (2L^2\|\vomega'_t-\vomega\|_2^2 +\|F(\vomega_t,\xi_t)-F(\vomega_t)\|^2_2)
\end{align*}
If one assumes finally that $\|\vomega'_t-\vomega_0\|_2 \leq R$ (assumption of the theorem) and that $\eta_t \leq \frac{1}{2L}$ we get,
\begin{align*}
  2\eta_tF(\vomega'_t)^\top (\vomega'_t-\vomega)
    & \leq  \|\vomega_{t}- \vomega\|^2_2 - \|\vomega_{t+1}- \vomega\|^2_2 + 2 \eta_t(F(\vomega_t,\xi_t)-F(\vomega_t))^\top(\vomega-\vomega_t) \\
    & \quad + 4\eta_t^2 (4L^2R^2 +\|F(\vomega_t,\xi_t)-F(\vomega_t)\|^2_2)
\end{align*}
where we used that $\|\vomega'_t-\vomega\|_2 \leq \|\vomega'_t-\vomega_0\|_2  + \|\vomega_0-\vomega\|_2  \leq 2R $.
Once again this equation is a particular case of Lemma~\ref{lemma:fictive_iterate} where $N_t = \|\vomega_t -\vomega\|_2^2$, $M_1(\vomega_t,\xi_t) = 4 (4L^2R^2 +\|F(\vomega_t,\xi_t)-F(\vomega_t)\|^2_2), \; M_2(\vomega'_t,\zeta_t) = 0, \; \zz_t = \vomega_t$ and $\Delta_t = F(\vomega_t,\xi_t)-F(\vomega_t)$. By Assumption~\ref{assum:var_bounded} $\EE[M_1(\vomega_t,\xi_t)] \leq 16L^2R^2+4\sigma^2$ and $\EE[\Delta_t|\vomega_t,\Delta_0,\ldots,\Delta_{t-1}] = \EE[\EE[\Delta_t|\vomega_t]|\Delta_0,\ldots,\Delta_{t-1}] = 0 $ so we can use Lemma~\ref{lemma:fictive_iterate} and get,
\begin{equation}
  \EE[\Err_R(\bar \vomega_T)] \leq \frac{R^2}{S_T} + \frac{\sigma^2 + 16L^2R^2+4\sigma^2}{2S_T} \sum_{t=0}^{T-1} \eta_t^2\,.
\end{equation}
\endproof

\newpage

\section{Additional experimental results} \label{app:additional_results}
\subsection{Toy non-convex GAN (2D and deterministic)} \label{sub:non_convex_gan}
\begin{figure}[h]
\vspace*{4mm}
\centering
\begin{subfigure}[b]{.48\linewidth}
\includegraphics[width=\linewidth,height = .8 \textwidth]{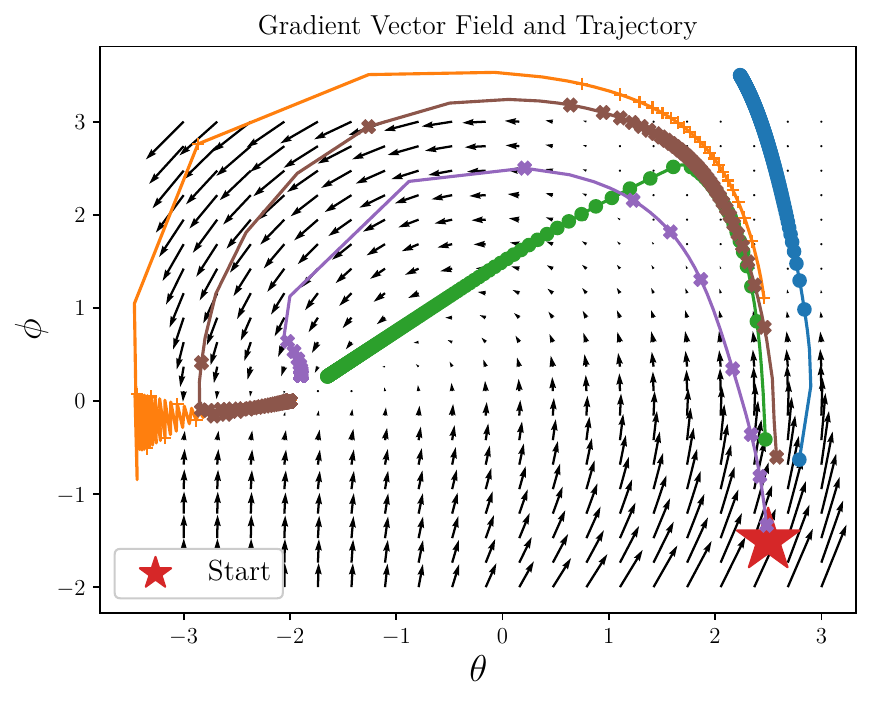}
\end{subfigure}
\hfill
\begin{subfigure}[b]{.48\linewidth}
\includegraphics[width=\linewidth,height = .8 \textwidth]{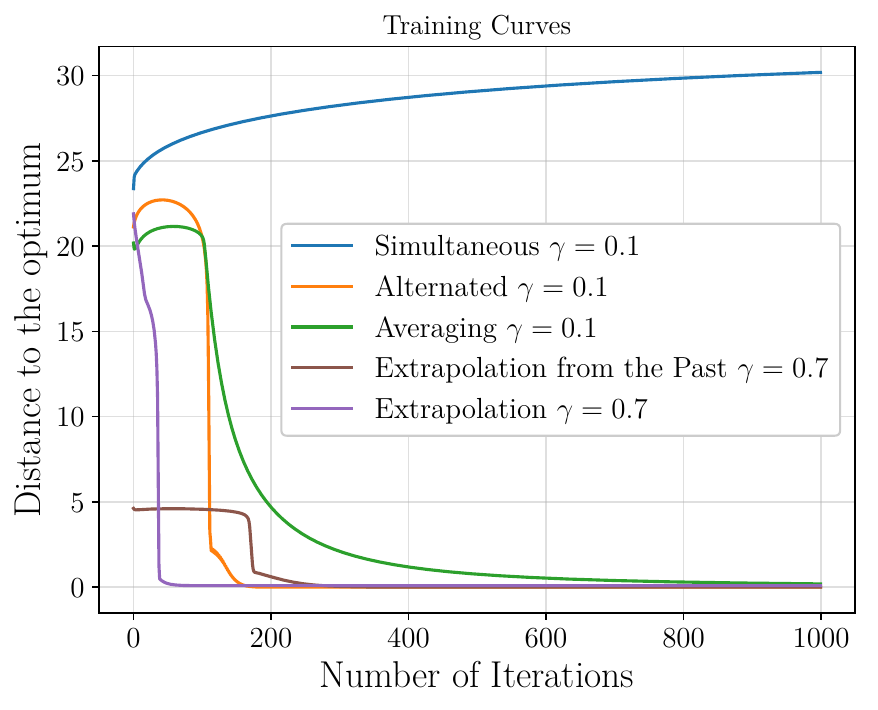}
\end{subfigure}

\caption{Comparison of five algorithms (described in Section~\ref{sec:optim}) on the non-convex GAN objective (\ref{eq:non_convex_gan}), using the optimal step-size for each method. \textbf{Left}: The gradient vector field and the dynamics of the different methods. \textbf{Right}:The distance to the optimum as a function of the number of iterations.
}
\label{fig:gan_toy}
\end{figure}

We now consider a task similar to \citep{mescheder2018convergence} where the discriminator is linear $D_\vphi(\vomega) = \vphi^T \vomega$, the generator is a Dirac distribution at $\vtheta$, $q_\vtheta = \delta_\vtheta$ and the distribution we try to match is also a Dirac at $\vomega^*$, $p = \delta_{\vomega^*}$. The minimax formulation from \citet{goodfellow2014generative} gives:
\begin{equation} \label{eq:non_convex_gan}
  \min_\vtheta \max_\vphi  - \log\big(1+e^{-\vphi^T\vomega^*}\big) - \log\big(1+ e^{\vphi^T\vtheta}\big)
\end{equation}
Note that as observed by \citet{nagarajan_gradient_2017}, this objective is concave-concave, making it hard to optimize. We compare the methods on this objective where we take $\vomega^*=-2$, thus the position of the equilibrium is shifted towards the position $(\vtheta, \vphi)=(-2, 0)$. The convergence and the gradient vector field are shown in Figure~\ref{fig:gan_toy}. We observe that depending on the initialization, some methods can fail to converge but
\emph{extrapolation}~\eqref{eq:update_implicit_extra} seems to perform better than the other methods.

\subsection{DCGAN with WGAN-GP objective}
\label{sub:dcgan_wgangp}

\begin{table}[h]\centering
\begin{tabular}{@{}c@{}}\toprule
\textbf{Generator}\\\toprule
\textit{Input:} $z \in \mathds{R}^{128} \sim \mathcal{N}(0, I) $ \\
Linear $128 \rightarrow 512 \times 4 \times 4$ \\
Batch Normalization \\
ReLU  \\
transposed conv. (kernel: $4{\times}4$, $512 \rightarrow 256$, stride: $2$, pad: 1)\\
Batch Normalization \\
ReLU  \\
transposed conv. (kernel: $4{\times}4$, $256 \rightarrow 128$, stride: $2$, pad: 1) \\
Batch Normalization \\
ReLU  \\
transposed conv. (kernel: $4{\times}4$, $128 \rightarrow 3$, stride: $2$, pad: 1) \\
$Tanh(\cdot)$\\
\bottomrule \\\textbf{Discriminator}\\\toprule
\textit{Input:} $x \in \mathds{R}^{3{\times}32{\times}32} $ \\
conv. (kernel: $4{\times}4$, $1 \rightarrow 64$; stride: $2$; pad:1) \\
LeakyReLU (negative slope: $0.2$) \\
conv. (kernel: $4{\times}4$, $64 \rightarrow 128$; stride: $2$; pad:1) \\
Batch Normalization \\
LeakyReLU (negative slope: $0.2$) \\
conv. (kernel: $4{\times}4$, $128 \rightarrow 256$; stride: $2$; pad:1) \\
Batch Normalization \\
LeakyReLU (negative slope: $0.2$) \\
Linear $128 \times 4 \times 4 \times 4 \rightarrow 1$\\
\bottomrule
\end{tabular}
\caption{DCGAN architecture used for our CIFAR-10 experiments. When using the gradient penalty (WGAN-GP), we remove the Batch Normalization layers in the discriminator.}\label{tab:mnist_arch}
\end{table}

\begin{figure}[h]
\centering
\begin{subfigure}[b]{.4 \linewidth}
\includegraphics[width=\textwidth]{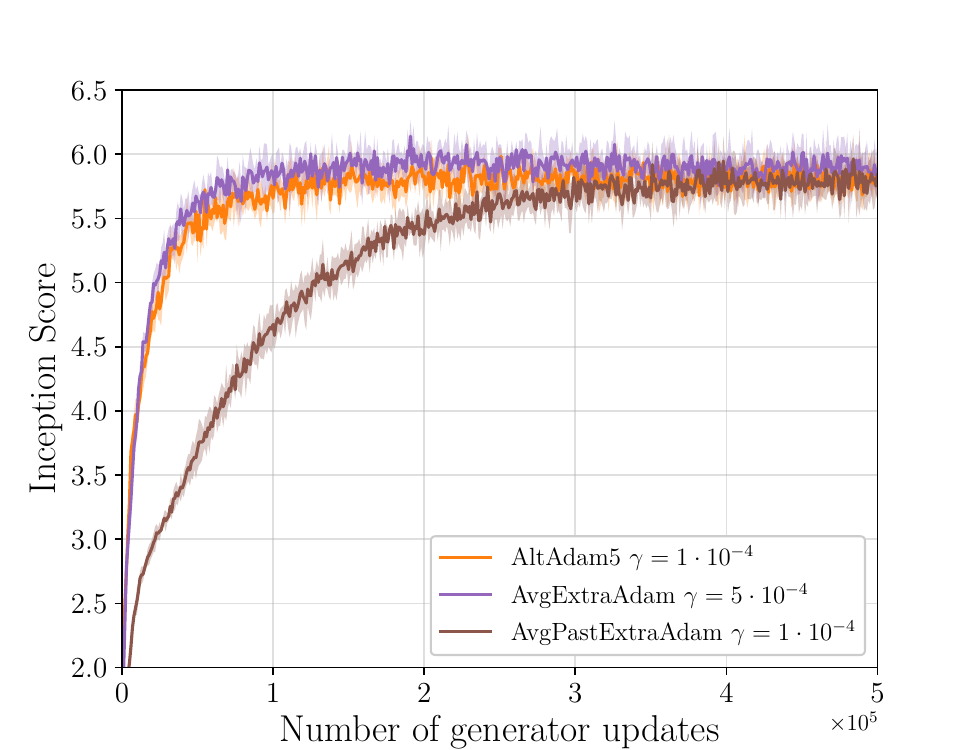}
\end{subfigure}
\begin{subfigure}[b]{.4 \linewidth}
\includegraphics[width=\textwidth]{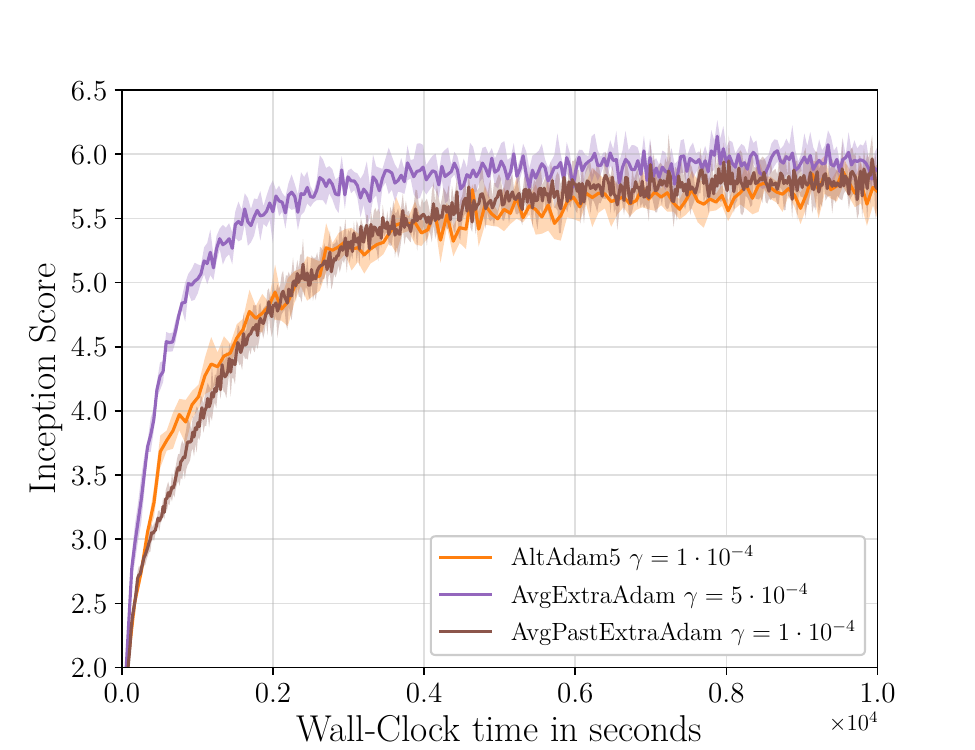}
\end{subfigure}
\caption{DCGAN architecture with WGAN-GP trained on CIFAR10: mean and standard deviation of the inception score computed over 5 runs for each method using the best performing learning rate plotted over number of generator updates (\textbf{Left}) and wall-clock time (\textbf{Right}); all experiments were run on a NVIDIA Quadro GP100 GPU. We see that ExtraAdam converges faster than the Adam baselines.
\label{fig:wgangp_time_cifar10}}
\end{figure}

In addition to the results presented in section \S\ref{sub:wgan_cifar10}, we also trained the DCGAN architecture with the WGAN-GP objective. The results are shown in Table~\ref{tab:inception_score_wgan_gp}.
The best results are achieved with \emph{uniform averaging} of AltAdam5. However, its iterations require to update the discriminator 5 times for every generator update. With a small drop in best final score, ExtraAdam can train WGAN-GP significantly faster (see Fig.~\ref{fig:wgangp_time_cifar10} right) as the discriminator and generator are updated only twice.

\begin{table}[h]
\vspace*{4mm}
      \centering
    \begin{tabular}{lcc}
      \toprule
      Model & \multicolumn{2}{c}{WGAN-GP (DCGAN)} \\
    \cmidrule(r){2-3}
      Method     & no averaging   & uniform avg\\
      \midrule
      SimAdam  & $\mathit{6.00 \pm .07}$&$6.01 \pm .08$   \\
      AltAdam5 & $\mathit{6.25 \pm .05}$&$\mathbf{6.51 \pm .05}$ \\
      ExtraAdam & $6.22 \pm .04$ & $6.35 \pm .05$  \\
      PastExtraAdam & $6.27\pm 0.06$ & $6.23\pm 0.13$ \\
      \bottomrule
    \end{tabular}
        \caption{
    Best inception scores (averaged over 5 runs) achieved on CIFAR10 for every considered Adam variant. We see that the techniques of extrapolation and averaging consistently enable improvements over the baselines (in italic).
    \label{tab:inception_score_wgan_gp} }
    \end{table}
 
\subsection{FID scores for ResNet architecture with WGAN-GP objective}
\label{sub:fid_wgangp}

\begin{table}[h]\centering
\begin{tabular}{@{}c@{}}\toprule
\textbf{Generator}\\\toprule
\textit{Input:} $z \in \mathds{R}^{128} \sim \mathcal{N}(0, I) $ \\
Linear $128 \rightarrow 128 \times 4 \times 4$ \\
ResBlock $128 \rightarrow 128$\\
ResBlock $128 \rightarrow 128$\\
ResBlock $128 \rightarrow 128$\\
Batch Normalization \\
ReLU  \\
transposed conv. (kernel: $3{\times}3$, $128 \rightarrow 3$, stride: $1$, pad: 1) \\
$Tanh(\cdot)$\\
\bottomrule \\\textbf{Discriminator}\\\toprule
\textit{Input:} $x \in \mathds{R}^{3{\times}32{\times}32} $ \\
ResBlock $3 \rightarrow 128$\\
ResBlock $128 \rightarrow 128$\\
ResBlock $128 \rightarrow 128$\\
ResBlock $128 \rightarrow 128$\\
Linear $128 \rightarrow 1$\\
\bottomrule
\end{tabular}
\caption{ResNet architecture used for our CIFAR-10 experiments. When using the gradient penalty (WGAN-GP), we remove the Batch Normalization layers in the discriminator.}\label{tab:mnist_arch}
\end{table}

In addition to the inception scores, we also computed the FID scores~\citep{heusel2017gans} using 50,000 samples for the ResNet architecture with the WGAN-GP objective; the results are presented in Table~\ref{tab:fid_wgan_gp}. We see that the results and conclusions are similar to the one obtained from the inception scores, adding an extrapolation step as well as using Exponential Moving Average (EMA) consistently improves the FID scores. However, contrary to the results from the inception score, we observe that uniform averaging does not necessarily improve the performance of the methods. This could be due to the fact that the samples produced using uniform averaging are more blurry and FID is more sensitive to blurriness; see \S\ref{sub:fid_wgangp} for more details about the effects of uniform averaging. 
\begin{table}[h]
\vspace*{4mm}
      \centering
    \begin{tabular}{lccc}
      \toprule
      Model & \multicolumn{3}{c}{WGAN-GP (ResNet)} \\
    \cmidrule(r){2-4}
      Method     & no averaging   & uniform avg & EMA\\
      \midrule
      SimAdam  & $\mathit{23.74 \pm 2.79}$&$26.29 \pm 5.56$ &$21.89 \pm 2.51$  \\
      AltAdam5 & $\mathit{21.65 \pm 0.66}$&$19.91 \pm 0.43$ &$20.69 \pm 0.37$\\
      ExtraAdam & $19.42 \pm 0.15$ & $18.13 \pm 0.51$ & $\mathbf{16.78 \pm 0.21}$ \\
      PastExtraAdam & $19.95\pm 0.38$ & $22.45\pm 0.93$ & $17.85 \pm 0.40$ \\
      OptimAdam & $\mathit{18.88\pm 0.55}$ & $21.23\pm 1.19$ & $16.91 \pm 0.32$ \\
      \bottomrule
    \end{tabular}
        \caption{
    Best FID scores (averaged over 5 runs) achieved on CIFAR10 for every considered Adam variant. OptimAdam is the related \emph{Optimistic Adam}~\citep{daskalakis2017training} algorithm. We see that the techniques of extrapolation and EMA consistently enable improvements over the baselines (in italic).
    \label{tab:fid_wgan_gp} }
    \end{table}

\clearpage

\subsection{Comparison of the methods with the same learning rate} \label{sub:comparison_learning_rate}

\begin{figure}
\centering
\begin{subfigure}[b]{.4 \linewidth}
  \includegraphics[width= \textwidth]{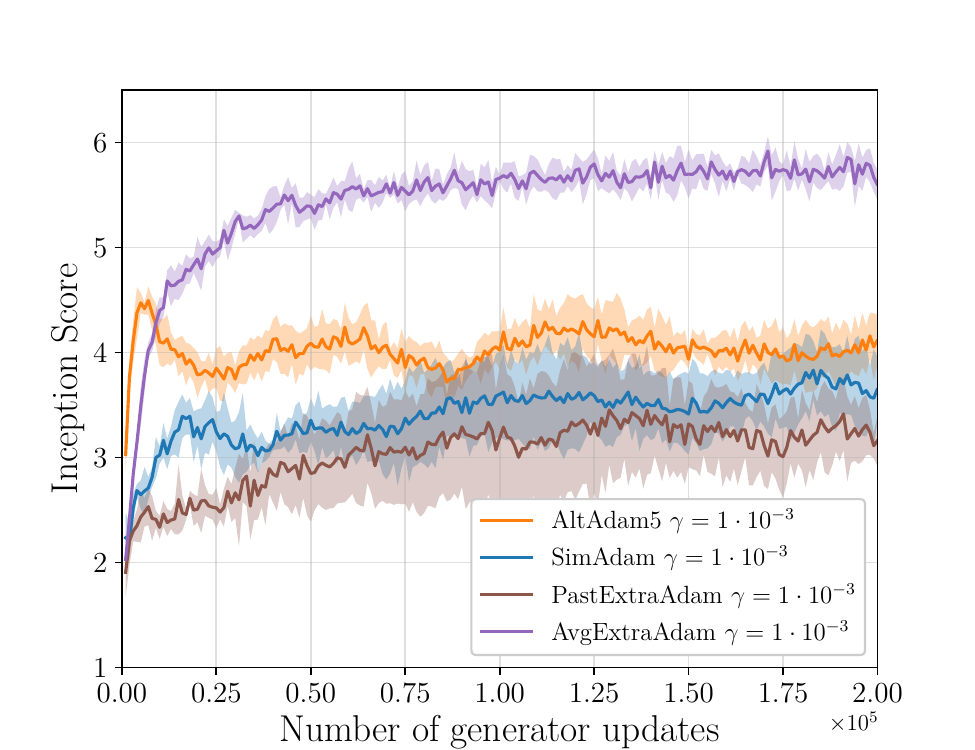}
  \caption{learning rate of $10^{-3}$}
\end{subfigure}\begin{subfigure}[b]{.4\linewidth}
  \includegraphics[width=\textwidth]{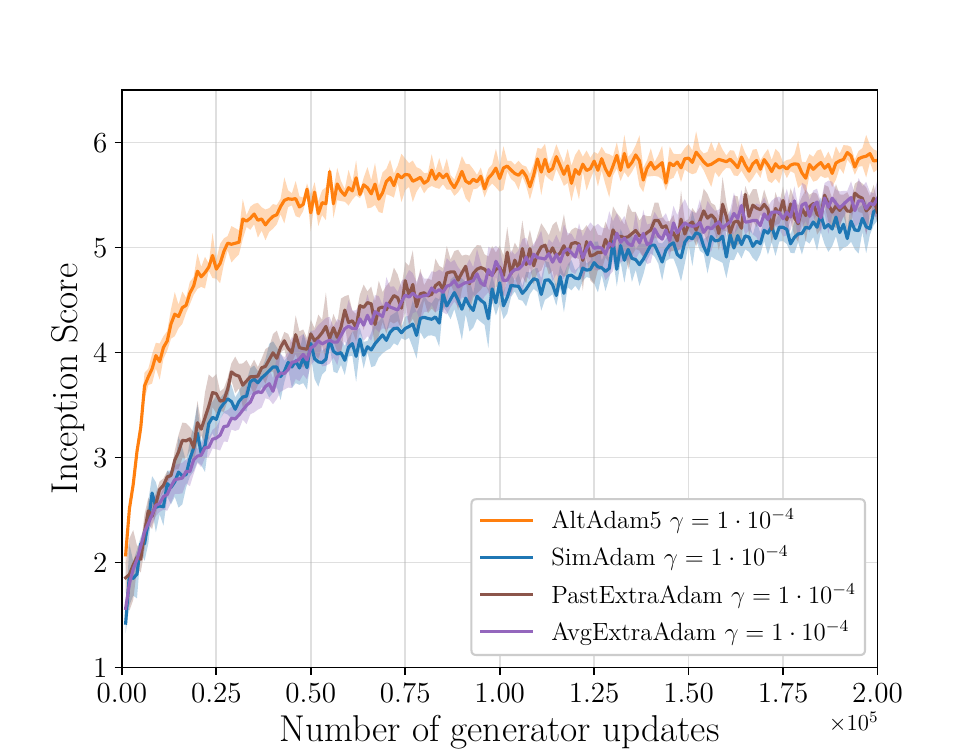}
  \caption{learning rate of $10^{-4}$}
\end{subfigure}
\caption{Inception score on CIFAR10 for WGAN-GP (DCGAN) over number of generator updates for different learning rates. We can see that AvgExtraAdam is less sensitive to the choice of learning rate.}
\label{fig:wgangp_cifar10_app}
\end{figure}

In this section, we compare how the methods presented in \S\ref{sec:experiments} perform with the same step-size. We follow the same protocol as in the experimental section \S\ref{sec:experiments}, we consider the DCGAN architecture with WGAN-GP experiment described in App\ \S\ref{sub:dcgan_wgangp}. In Figure~\ref{fig:wgangp_cifar10_app} we plot the inception score provided by each training method as a function of the number of generator updates. Note that these plots advantage \textbf{AltAdam5} a bit because each iteration of this algorithm is a bit more costly (since it perform 5 discriminator updates for each generator update).
Nevertheless, the goal of this experiment is not to show that \textbf{AltAdam5} is faster but to show that \textbf{ExtraAdam} is less sensitive to the choice of learning rate and can be used with higher learning rates with less degradation.

\begin{figure}
\centering
\begin{subfigure}[b]{0.4\linewidth}
  \includegraphics[width=\textwidth]{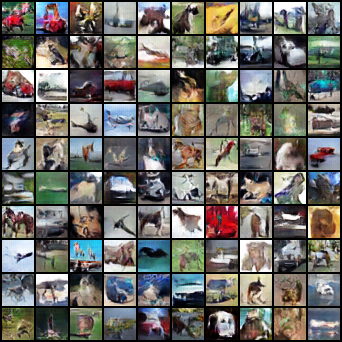}
  \caption{AvgExtraAdam with $\eta=10^{-3}$}
\end{subfigure}\quad
\begin{subfigure}[b]{0.4\linewidth}
  \includegraphics[width=\textwidth]{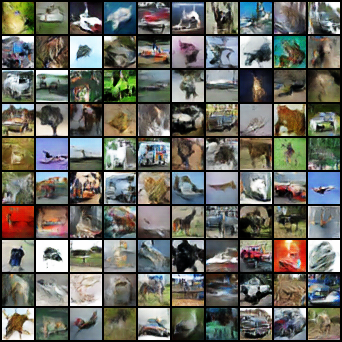}
  \caption{AvgExtraAdam with $\eta=10^{-4}$}
\end{subfigure}
\begin{subfigure}[b]{0.4\linewidth}
  \includegraphics[width=\textwidth]{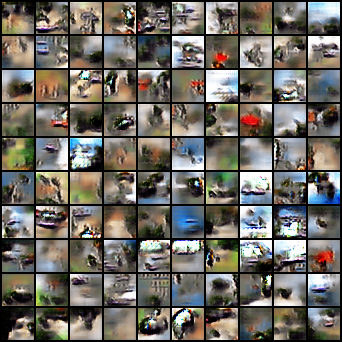}
  \caption{AltAdam with $\eta=10^{-3}$}
\end{subfigure}\quad
\begin{subfigure}[b]{0.4\linewidth}
  \includegraphics[width=\textwidth]{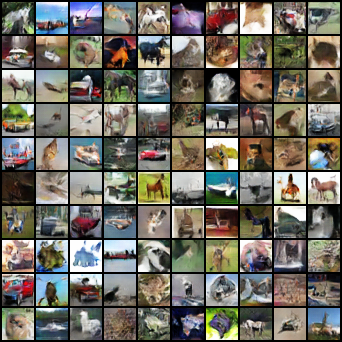}
  \caption{AltAdam with $\eta=10^{-4}$}
\end{subfigure}
\caption{Comparison of the samples quality on the WGAN-GP (DCGAN) experiment for different methods and learning rate $\eta$.}
\label{fig:wgangp_cifar10_sample}
\end{figure}

In Figure~\ref{fig:wgangp_cifar10_sample}, we compare the sample quality on the DCGAN architecture with the WGAN-GP objective of \textbf{AltAdam5} and \textbf{AvgExtraAdam} for different step-sizes.
We notice that for \textbf{AvgExtraAdam}, the sample quality does not significantly change whereas the sample quality of \textbf{AltAdam5} seems to be really sensitive to step-size tunning.

We think that robustness to step-size tuning is a key property for an optimization algorithm in order to save as much time as possible to tune other hyperparameters of the learning procedure such as regularization.

\newpage

\subsection{Comparison of the methods with and without uniform averaging} \label{sub:comparison_averaging}

\begin{figure}[h]
  \centering
  \begin{subfigure}[b]{.43 \linewidth}
    \includegraphics[width= \textwidth]{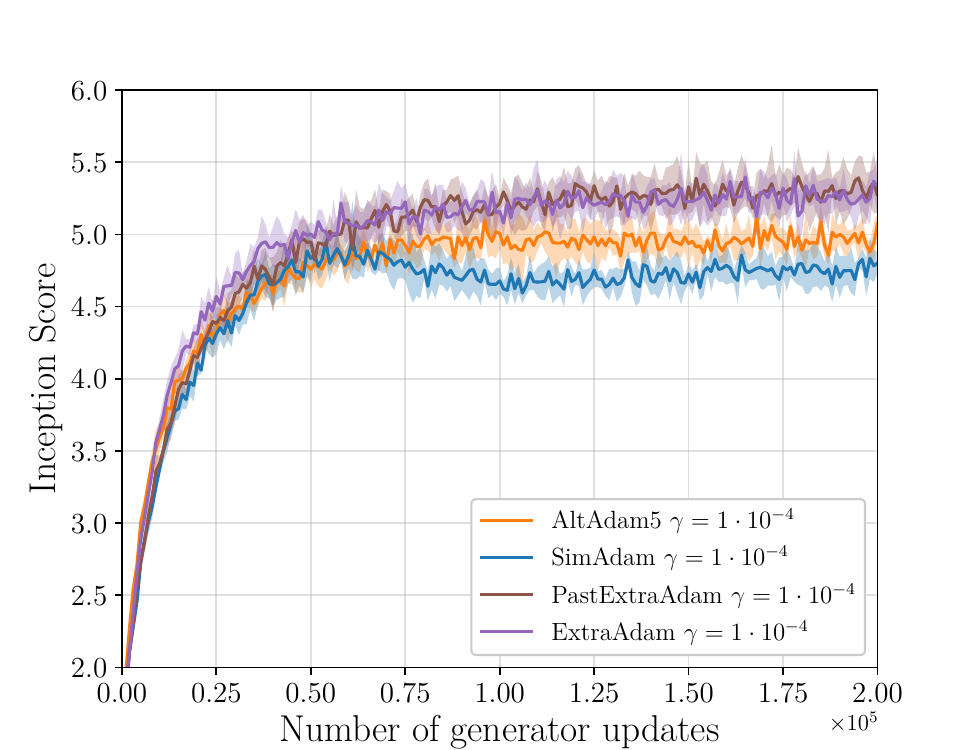}
    \caption{with averaging}
  \end{subfigure}  \begin{subfigure}[b]{.43 \linewidth}
    \includegraphics[width= \textwidth]{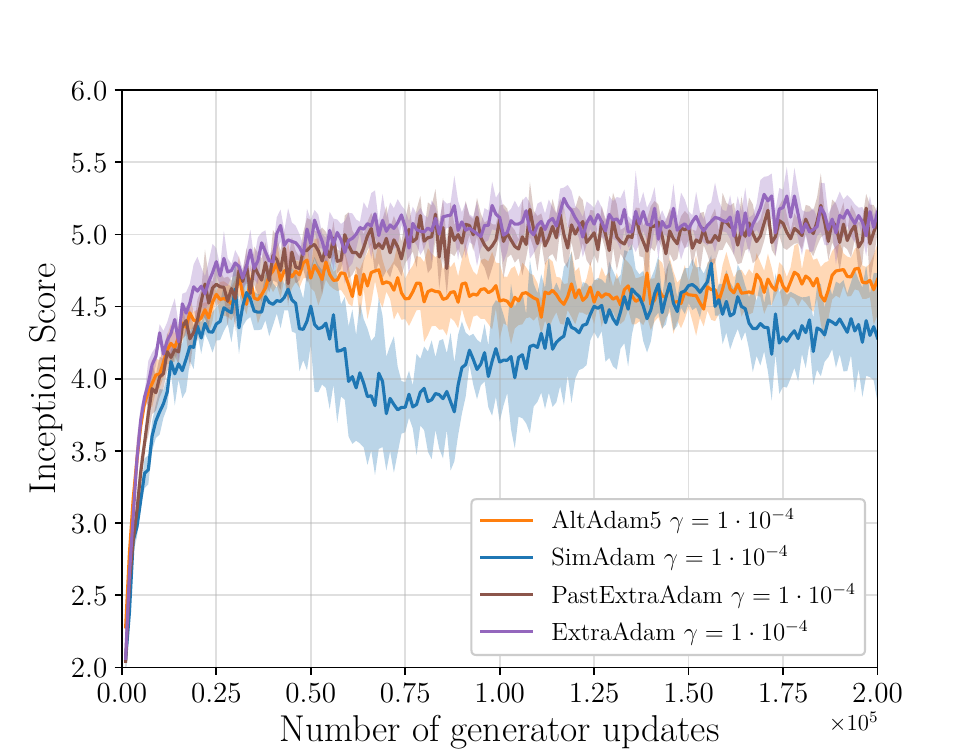}
    \caption{without averaging}
  \end{subfigure}  \caption{Inception Score on CIFAR10 for WGAN over number of generator updates with and without averaging. We can see that averaging improve the inception score.}
  \label{fig:with_without_wgan}
\end{figure}

\begin{figure}
  \centering
  \begin{subfigure}[b]{.43 \linewidth}
    \includegraphics[width= \textwidth]{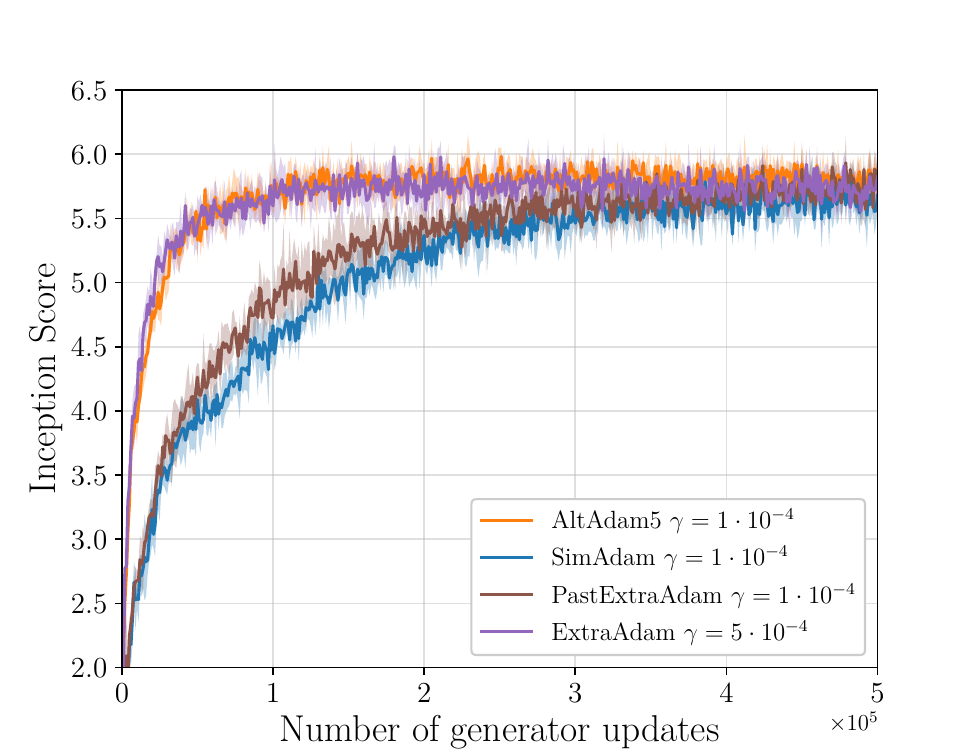}
    \caption{with averaging}
  \end{subfigure}  \begin{subfigure}[b]{.43 \linewidth}
    \includegraphics[width= \textwidth]{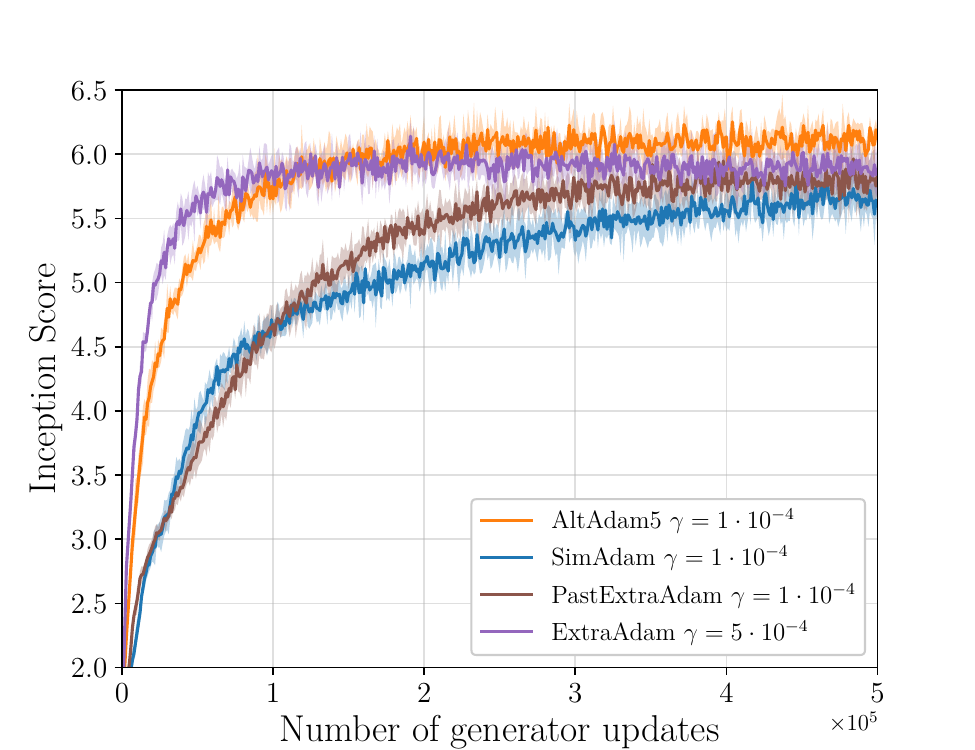}
    \caption{without averaging}
  \end{subfigure}  \caption{Inception score on CIFAR10 for WGAN-GP (DCGAN) over number of generator updates}
    \label{fig:with_without_wgangp}
\end{figure}

In this section, we compare how uniform averaging affect the performance of the methods presented in \S\ref{sec:experiments}. We follow the same protocol as in the experimental section \S\ref{sec:experiments}, we consider the DCGAN architecture with the WGAN and weight clipping objective as well as the WGAN-GP objective. In Figure~\ref{fig:with_without_wgan} and~\ref{fig:with_without_wgangp}, we plot the inception score provided by each training method as a function of the number of generator updates with and without uniform averaging.

\begin{figure}
\centering
\begin{subfigure}[b]{0.43\textwidth}
  \includegraphics[width= \textwidth]{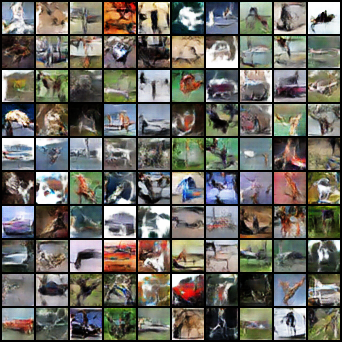}
  \caption{PastExtraAdam without averaging}
\end{subfigure}\quad
\begin{subfigure}[b]{0.43\textwidth}
  \includegraphics[width= \textwidth]{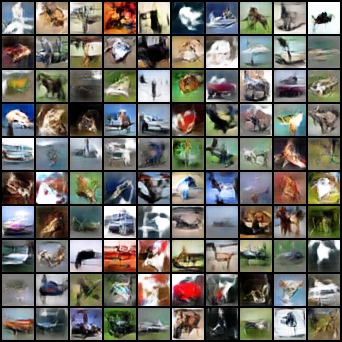}
  \caption{PastExtraAdam with averaging}
\end{subfigure}
\begin{subfigure}[b]{0.43\textwidth}
  \includegraphics[width= \textwidth]{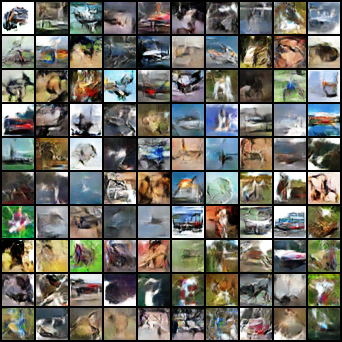}
  \caption{AltAdam5 without averaging}
\end{subfigure}\quad
\begin{subfigure}[b]{0.43\textwidth}
  \includegraphics[width= \textwidth]{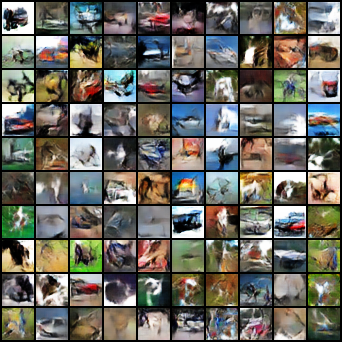}
  \caption{AltAdam5 with averaging}
\end{subfigure}
\caption{Comparison of the samples of a WGAN trained with the different methods with and without averaging. Although averaging improves the inception score, the samples seem more blurry}
\label{fig:wgangp_cifar10_lr}
\end{figure}

\begin{figure}
  \centering
  \includegraphics[width=0.5\textwidth]{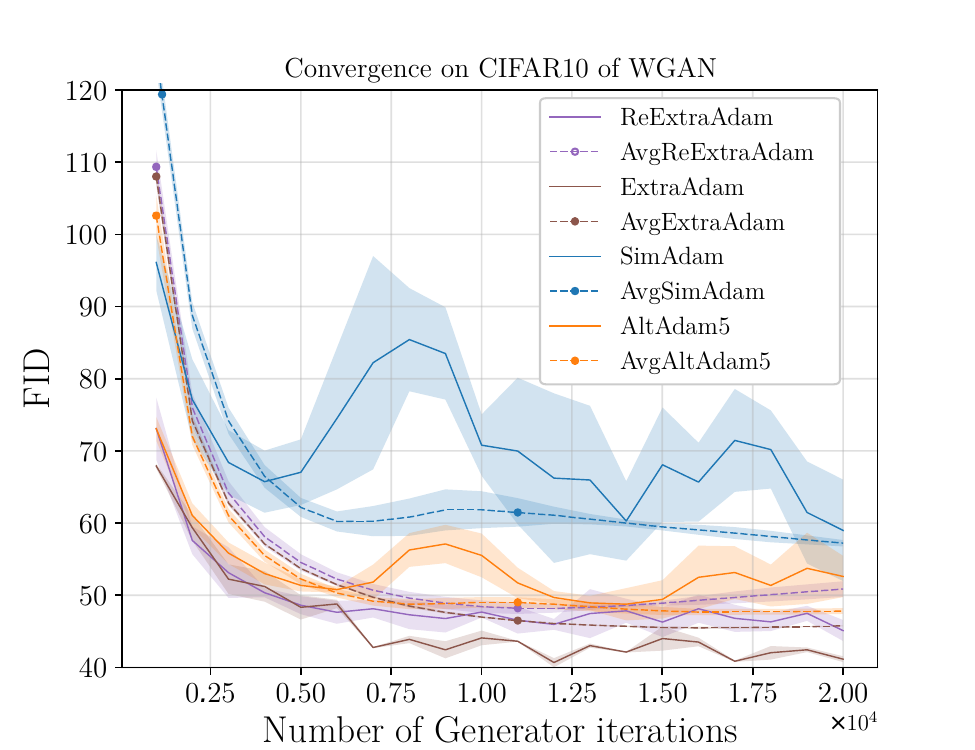}
  \caption{The Fr\'echet Inception Distance (FID) from \citet{heusel2017gans} computed using 50,000 samples, on the WGAN experiments. ReExtraAdam refers to Alg.~\ref{alg:ReAvgExtraSGD} introduced in \S\ref{sec:another_extension_of_sem}. We can see that averaging performs worse than when comparing with the Inception Score. We observed that the samples generated by using averaging are a little more blurry and that the FID is more sensitive to blurriness, thus providing an explanation for this observation.}
  \label{fig:fid_cifar10_wgan}
\end{figure}

We notice that uniform averaging seems to improve the inception score, nevertheless it looks like the sample are a bit more blurry (see Figure~\ref{fig:wgangp_cifar10_lr}). This is confirmed by our result (Figure~\ref{fig:fid_cifar10_wgan}) on the Fréchet Inception Distance (FID) which is more sensitive to blurriness. A similar observation about FID was made in \S\ref{sub:fid_wgangp}.

\clearpage

\section{Hyperparameters}
\begin{table}[h]
\centering
\begin{tabular}{ll}\toprule
\textbf{(DCGAN) WGAN Hyperparameters}\\\toprule
Batch size &= $64$ \\
Number of generator update &= $500,000$ \\
Adam $\beta_1$ &= $0.5$ \\
Adam $\beta_2$ &= $0.9$ \\
Weight clipping for the discriminator &= $0.01$ \\
Learning rate for generator &= $2\times10^{-5}$ (for Adam1, Adam5, PastExtraAdam, OptimisticAdam) \\
 &= $5\times10^{-5}$ (for ExtraAdam)\\
Learning rate for discriminator &= $2\times10^{-4}$ (for Adam1, Adam5, PastExtraAdam, OptimisticAdam) \\
 &= $5\times10^{-4}$ (for ExtraAdam) \\
$\beta$ for EMA &= $0.999$ \\
\bottomrule
\end{tabular}
\end{table}

\begin{table}[h]
\centering
\begin{tabular}{ll}\toprule
\textbf{(DCGAN) WGAN-GP Hyperparameters}\\\toprule
Batch size &= $64$ \\
Number of generator update &= $500,000$ \\
Adam $\beta_1$ &= $0.5$ \\
Adam $\beta_2$ &= $0.9$ \\
Gradient penalty &= $10$ \\
Learning rate for generator &= $1\times10^{-4}$ (for Adam1, Adam5, PastExtraAdam, OptimisticAdam) \\
 &= $5\times10^{-4}$ (for ExtraAdam)\\
Learning rate for discriminator &= $1\times10^{-4}$ (for Adam1, Adam5, PastExtraAdam, OptimisticAdam) \\
 &= $5\times10^{-4}$ (for ExtraAdam) \\
$\beta$ for EMA &= $0.999$ \\
\bottomrule
\end{tabular}
\end{table}

\begin{table}[h]
\centering
\begin{tabular}{ll}\toprule
\textbf{(ResNet) WGAN-GP Hyperparameters}\\\toprule
Batch size &= $64$ \\
Number of generator update &= $500,000$ \\
Adam $\beta_1$ &= $0.5$ \\
Adam $\beta_2$ &= $0.9$ \\
Gradient penalty &= $10$ \\
Learning rate for generator &= $2\times10^{-5}$ (for Adam1, Adam5, PastExtraAdam, OptimisticAdam) \\
 &= $5\times10^{-5}$ (for ExtraAdam)\\
Learning rate for discriminator &= $2\times10^{-4}$ (for Adam1, Adam5, PastExtraAdam, OptimisticAdam) \\
 &= $5\times10^{-4}$ (for ExtraAdam) \\
$\beta$ for EMA &= $0.9999$ \\
\bottomrule
\end{tabular}
\end{table}

\end{document}

%% file: math_commands.tex
\usepackage{amsmath,amsfonts,bm}
\usepackage{algorithm,algorithmic}

\def\1{\bm{1}}

\def\rvx{{\mathbf{x}}}

\def\rvz{{\mathbf{z}}}

\def\vtheta{{\bm{\theta}}}
\def\vphi{{\bm{\varphi}}}

\def\vomega{{\bm{\omega}}}

\def\vb{{\bm{b}}}
\def\vc{{\bm{c}}}

\def\mA{{\bm{A}}}

\def\mD{{\bm{D}}}

\def\mU{{\bm{U}}}
\def\mV{{\bm{V}}}

\DeclareMathAlphabet{\mathsfit}{\encodingdefault}{\sfdefault}{m}{sl}
\SetMathAlphabet{\mathsfit}{bold}{\encodingdefault}{\sfdefault}{bx}{n}

\def\sN{{\mathbb{N}}}

\newcommand{\R}{\mathbb{R}}

\DeclareMathOperator*{\argmin}{arg\,min}

\DeclareMathOperator{\dist}{dist}

\DeclareMathOperator{\diag}{\bm{diag}}

\newlength\myindent
\newcommand\bindent{%
  \begingroup
  \setlength{\itemindent}{\myindent}
  \addtolength{\algorithmicindent}{\myindent}
}
\newcommand\eindent{\endgroup}

%% file: defs.tex
\definecolor{beaublue}{rgb}{0.74, 0.83, 0.9}
\definecolor{babyblueeyes}{rgb}{0.63, 0.79, 0.95} 
\definecolor{myblue}{HTML}{D2E4FC}

\DeclareMathOperator{\Err}{Err}

\usepackage{pifont}

\usepackage{todonotes}

\def\RR{{\mathbb R}}
\def\EE{{\mathbb E}}

\def\uu{{\boldsymbol u}}

\def\zz{{\boldsymbol z}}

\def\LL{{\mathcal L}}

\def\defas{\stackrel{\text{def}}{=}}

\newtheorem{assumption}{Assumption}
\newtheorem{theorem}{Theorem}
\newtheorem{lemma}{Lemma}
\newtheorem{proposition}{Proposition}

\newtheorem{corollary}{Corollary}
\newtheorem{definition}{Definition}

\newtheorem*{rep@theorem}{\rep@title}
\newcommand{\newreptheorem}[2]{%
\newenvironment{rep#1}[1]{%
 \def\rep@title{#2 \ref{##1}}%
 \begin{rep@theorem}}%
 {\end{rep@theorem}}}
\newreptheorem{theorem}{Theorem'}
\newtheorem*{rep@proposition}{\rep@title}
\newcommand{\newrepproposition}[2]{%
\newenvironment{rep#1}[1]{%
 \def\rep@title{#2 \ref{##1}}%
 \begin{rep@proposition}}%
 {\end{rep@proposition}}}
\newreptheorem{proposition}{Proposition'}

%% file: main.bbl
\begin{thebibliography}{53}
\providecommand{\natexlab}[1]{#1}
\providecommand{\url}[1]{\texttt{#1}}
\expandafter\ifx\csname urlstyle\endcsname\relax
  \providecommand{\doi}[1]{doi: #1}\else
  \providecommand{\doi}{doi: \begingroup \urlstyle{rm}\Url}\fi

\bibitem[Arjovsky et~al.(2017)Arjovsky, Chintala, and
  Bottou]{arjovsky2017wasserstein}
M.~Arjovsky, S.~Chintala, and L.~Bottou.
\newblock Wasserstein generative adversarial networks.
\newblock In \emph{ICML}, 2017.

\bibitem[Atkinson(2003)]{atkinson2003introduction}
K.~E. Atkinson.
\newblock \emph{An introduction to numerical analysis}.
\newblock John Wiley \& Sons, 2003.

\bibitem[Boyd and Vandenberghe(2004)]{boyd2004convex}
S.~Boyd and L.~Vandenberghe.
\newblock \emph{Convex optimization}.
\newblock Cambridge university press, 2004.

\bibitem[Bruck(1977)]{bruck_weak_1977}
R.~E. Bruck.
\newblock On the weak convergence of an ergodic iteration for the solution of
  variational inequalities for monotone operators in {H}ilbert space.
\newblock \emph{Journal of Mathematical Analysis and Applications}, 1977.

\bibitem[Chen and Rockafellar(1997)]{chen1997convergence}
G.~H. Chen and R.~T. Rockafellar.
\newblock Convergence rates in forward--backward splitting.
\newblock \emph{SIAM Journal on Optimization}, 1997.

\bibitem[Chiang et~al.(2012)Chiang, Yang, Lee, Mahdavi, Lu, Jin, and
  Zhu]{chiang2012online}
C.-K. Chiang, T.~Yang, C.-J. Lee, M.~Mahdavi, C.-J. Lu, R.~Jin, and S.~Zhu.
\newblock Online optimization with gradual variations.
\newblock In \emph{COLT}, 2012.

\bibitem[Crespi et~al.(2005)Crespi, Guerraggio, and Rocca]{crespi2005MVI}
G.~P. Crespi, A.~Guerraggio, and M.~Rocca.
\newblock Minty variational inequality and optimization: scalar and vector
  case.
\newblock In \emph{Generalized Convexity, Generalized Monotonicity and
  Applications}, 2005.

\bibitem[Daskalakis et~al.(2018)Daskalakis, Ilyas, Syrgkanis, and
  Zeng]{daskalakis2017training}
C.~Daskalakis, A.~Ilyas, V.~Syrgkanis, and H.~Zeng.
\newblock Training {GAN}s with optimism.
\newblock In \emph{ICLR}, 2018.

\bibitem[Fedus et~al.(2018)Fedus, Rosca, Lakshminarayanan, Dai, Mohamed, and
  Goodfellow]{fedus2017many}
W.~Fedus, M.~Rosca, B.~Lakshminarayanan, A.~M. Dai, S.~Mohamed, and
  I.~Goodfellow.
\newblock Many paths to equilibrium: {GAN}s do not need to decrease a
  divergence at every step.
\newblock In \emph{ICLR}, 2018.

\bibitem[Gidel et~al.(2017)Gidel, Jebara, and Lacoste-Julien]{gidel2017frank}
G.~Gidel, T.~Jebara, and S.~Lacoste-Julien.
\newblock Frank-{Wolfe} algorithms for saddle point problems.
\newblock In \emph{AISTATS}, 2017.

\bibitem[Gidel et~al.(2019)Gidel, Askari~Hemmat, Mohammad, Gabriel, R\'emi,
  Simon, and Ioannis]{gidel2018momentum}
G.~Gidel, R.~Askari~Hemmat, P.~Mohammad, H.~Gabriel, L.~R\'emi, L.-J. Simon,
  and M.~Ioannis.
\newblock Negative momentum for improved game dynamics.
\newblock In \emph{AISTATS}, 2019.

\bibitem[Goodfellow(2016)]{goodfellow2016nips}
I.~Goodfellow.
\newblock {NIPS} 2016 tutorial: Generative adversarial networks.
\newblock \emph{arXiv:1701.00160}, 2016.

\bibitem[Goodfellow et~al.(2014)Goodfellow, Pouget-Abadie, Mirza, Xu,
  Warde-Farley, Ozair, Courville, and Bengio]{goodfellow2014generative}
I.~Goodfellow, J.~Pouget-Abadie, M.~Mirza, B.~Xu, D.~Warde-Farley, S.~Ozair,
  A.~Courville, and Y.~Bengio.
\newblock Generative adversarial nets.
\newblock In \emph{NIPS}, 2014.

\bibitem[Grnarova et~al.(2018)Grnarova, Levy, Lucchi, Hofmann, and
  Krause]{grnarova2017online}
P.~Grnarova, K.~Y. Levy, A.~Lucchi, T.~Hofmann, and A.~Krause.
\newblock An online learning approach to generative adversarial networks.
\newblock In \emph{ICLR}, 2018.

\bibitem[Gulrajani et~al.(2017)Gulrajani, Ahmed, Arjovsky, Dumoulin, and
  Courville]{gulrajani2017improved}
I.~Gulrajani, F.~Ahmed, M.~Arjovsky, V.~Dumoulin, and A.~C. Courville.
\newblock Improved training of wasserstein {GAN}s.
\newblock In \emph{NIPS}, 2017.

\bibitem[Harker and Pang(1990)]{harker1990finite}
P.~T. Harker and J.-S. Pang.
\newblock Finite-dimensional variational inequality and nonlinear
  complementarity problems: a survey of theory, algorithms and applications.
\newblock \emph{Mathematical programming}, 1990.

\bibitem[Hazan et~al.(2017)Hazan, Singh, and Zhang]{hazan2017efficient}
E.~Hazan, K.~Singh, and C.~Zhang.
\newblock Efficient regret minimization in non-convex games.
\newblock In \emph{ICML}, 2017.

\bibitem[Heusel et~al.(2017)Heusel, Ramsauer, Unterthiner, Nessler, and
  Hochreiter]{heusel2017gans}
M.~Heusel, H.~Ramsauer, T.~Unterthiner, B.~Nessler, and S.~Hochreiter.
\newblock {GAN}s trained by a two time-scale update rule converge to a local
  {N}ash equilibrium.
\newblock In \emph{NIPS}, 2017.

\bibitem[Iusem et~al.(2017)Iusem, Jofr{\'e}, Oliveira, and
  Thompson]{iusem_extragradient_2017}
A.~Iusem, A.~Jofr{\'e}, R.~I. Oliveira, and P.~Thompson.
\newblock Extragradient method with variance reduction for stochastic
  variational inequalities.
\newblock \emph{SIAM Journal on Optimization}, 2017.

\bibitem[Juditsky et~al.(2011)Juditsky, Nemirovski, and
  Tauvel]{juditsky2011solving}
A.~Juditsky, A.~Nemirovski, and C.~Tauvel.
\newblock Solving variational inequalities with stochastic mirror-prox
  algorithm.
\newblock \emph{Stochastic Systems}, 2011.

\bibitem[Karras et~al.(2018)Karras, Aila, Laine, and
  Lehtinen]{karras2017progressive}
T.~Karras, T.~Aila, S.~Laine, and J.~Lehtinen.
\newblock Progressive growing of {GAN}s for improved quality, stability, and
  variation.
\newblock In \emph{ICLR}, 2018.

\bibitem[Kingma and Ba(2015)]{kingma2014adam}
D.~P. Kingma and J.~Ba.
\newblock Adam: A method for stochastic optimization.
\newblock In \emph{ICLR}, 2015.

\bibitem[Korpelevich(1976)]{korpelevich1976extragradient}
G.~Korpelevich.
\newblock The extragradient method for finding saddle points and other
  problems.
\newblock \emph{Matecon}, 12, 1976.

\bibitem[Krizhevsky and Hinton(2009)]{krizhevsky2009learning}
A.~Krizhevsky and G.~Hinton.
\newblock Learning multiple layers of features from tiny images.
\newblock Master's thesis, University of Toronto, Canada, 2009.

\bibitem[Larsson and Patriksson(1994)]{larsson1994class}
T.~Larsson and M.~Patriksson.
\newblock A class of gap functions for variational inequalities.
\newblock \emph{Math. Program.}, 1994.

\bibitem[Ledig et~al.(2017)Ledig, Theis, Husz{\'a}r, Caballero, Cunningham,
  Acosta, Aitken, Tejani, Totz, Wang, et~al.]{ledig2017photo}
C.~Ledig, L.~Theis, F.~Husz{\'a}r, J.~Caballero, A.~Cunningham, A.~Acosta,
  A.~P. Aitken, A.~Tejani, J.~Totz, Z.~Wang, et~al.
\newblock Photo-realistic single image super-resolution using a generative
  adversarial network.
\newblock In \emph{CVPR}, 2017.

\bibitem[Li et~al.(2017)Li, Schwing, Wang, and Zemel]{li2017dualing}
Y.~Li, A.~Schwing, K.-C. Wang, and R.~Zemel.
\newblock Dualing {GAN}s.
\newblock In \emph{NIPS}, 2017.

\bibitem[Mertikopoulos et~al.(2019)Mertikopoulos, Zenati, Lecouat, Foo,
  Chandrasekhar, and Piliouras]{mertikopoulos2018mirror}
P.~Mertikopoulos, H.~Zenati, B.~Lecouat, C.-S. Foo, V.~Chandrasekhar, and
  G.~Piliouras.
\newblock Mirror descent in saddle-point problems: going the extra (gradient)
  mile.
\newblock In \emph{ICLR}, 2019.
\newblock To appear.

\bibitem[Mescheder et~al.(2017)Mescheder, Nowozin, and
  Geiger]{mescheder_numerics_2017}
L.~Mescheder, S.~Nowozin, and A.~Geiger.
\newblock The numerics of {GANs}.
\newblock In \emph{NIPS}, 2017.

\bibitem[Mescheder et~al.(2018)Mescheder, Geiger, and
  Nowozin]{mescheder2018convergence}
L.~Mescheder, A.~Geiger, and S.~Nowozin.
\newblock Which training methods for {GANs} do actually converge?
\newblock In \emph{ICML}, 2018.

\bibitem[Metz et~al.(2017)Metz, Poole, Pfau, and
  Sohl-Dickstein]{metz_unrolled_2017}
L.~Metz, B.~Poole, D.~Pfau, and J.~Sohl-Dickstein.
\newblock Unrolled generative adversarial networks.
\newblock In \emph{ICLR}, 2017.

\bibitem[Miyato et~al.(2018)Miyato, Kataoka, Koyama, and
  Yoshida]{miyato2018spectral}
T.~Miyato, T.~Kataoka, M.~Koyama, and Y.~Yoshida.
\newblock Spectral normalization for generative adversarial networks.
\newblock In \emph{ICLR}, 2018.

\bibitem[Nagarajan and Kolter(2017)]{nagarajan_gradient_2017}
V.~Nagarajan and J.~Z. Kolter.
\newblock Gradient descent {GAN} optimization is locally stable.
\newblock In \emph{NIPS}, 2017.

\bibitem[Nedić and Ozdaglar(2009)]{nedic_subgradient_2009}
A.~Nedić and A.~Ozdaglar.
\newblock Subgradient methods for saddle-point problems.
\newblock \emph{J Optim Theory Appl}, 2009.

\bibitem[Nemirovski(2004)]{nemirovski_prox-method_2004}
A.~Nemirovski.
\newblock Prox-method with rate of convergence {$O(1/t)$} for variational
  inequalities with lipschitz continuous monotone operators and smooth
  convex-concave saddle point problems.
\newblock \emph{{SIAM} J. Optim.}, 2004.

\bibitem[Nemirovski et~al.(2009)Nemirovski, Juditsky, Lan, and
  Shapiro]{nemirovski2009robust}
A.~Nemirovski, A.~Juditsky, G.~Lan, and A.~Shapiro.
\newblock Robust stochastic approximation approach to stochastic programming.
\newblock \emph{SIAM Journal on optimization}, 2009.

\bibitem[Nesterov(1983)]{yurii1983introductory}
Y.~Nesterov.
\newblock \emph{Introductory Lectures On Convex Optimization.}
\newblock Springer, 1983.

\bibitem[Nesterov(2007)]{nesterov2007dual}
Y.~Nesterov.
\newblock Dual extrapolation and its applications to solving variational
  inequalities and related problems.
\newblock \emph{Math. Program.}, 2007.

\bibitem[Nowozin et~al.(2016)Nowozin, Cseke, and Tomioka]{nowozin_f-gan:_2016}
S.~Nowozin, B.~Cseke, and R.~Tomioka.
\newblock f-{GAN}: Training generative neural samplers using variational
  divergence minimization.
\newblock In \emph{NIPS}, 2016.

\bibitem[Palaniappan and Bach(2016)]{palaniappan2016stochastic}
B.~Palaniappan and F.~Bach.
\newblock Stochastic variance reduction methods for saddle-point problems.
\newblock In \emph{NIPS}, 2016.

\bibitem[Polyak(1963)]{polyak1963gradient}
B.~T. Polyak.
\newblock Gradient methods for minimizing functionals.
\newblock \emph{Zhurnal Vychislitel'noi Matematiki i Matematicheskoi Fiziki},
  1963.

\bibitem[Popov(1980)]{popov1980modification}
L.~D. Popov.
\newblock A modification of the arrow-hurwicz method for search of saddle
  points.
\newblock \emph{Mathematical notes of the Academy of Sciences of the USSR},
  1980.

\bibitem[Radford et~al.(2016)Radford, Metz, and
  Chintala]{radford2016unsupervised}
A.~Radford, L.~Metz, and S.~Chintala.
\newblock Unsupervised representation learning with deep convolutional
  generative adversarial networks.
\newblock In \emph{ICLR}, 2016.

\bibitem[Rakhlin and Sridharan(2013)]{rakhlin2013online}
A.~Rakhlin and K.~Sridharan.
\newblock Online learning with predictable sequences.
\newblock In \emph{COLT}, 2013.

\bibitem[Robbins and Monro(1951)]{robbins1951stochastic}
H.~Robbins and S.~Monro.
\newblock A stochastic approximation method.
\newblock \emph{The Annals of Mathematical Statistics}, 1951.

\bibitem[Salimans et~al.(2016)Salimans, Goodfellow, Zaremba, Cheung, Radford,
  and Chen]{salimans2016improved}
T.~Salimans, I.~Goodfellow, W.~Zaremba, V.~Cheung, A.~Radford, and X.~Chen.
\newblock Improved techniques for training {GAN}s.
\newblock In \emph{NIPS}, 2016.

\bibitem[Sutskever(2013)]{sutskever2013training}
I.~Sutskever.
\newblock \emph{Training recurrent neural networks}.
\newblock PhD thesis, 2013.

\bibitem[Tseng(1995)]{tseng1995linear}
P.~Tseng.
\newblock On linear convergence of iterative methods for the variational
  inequality problem.
\newblock \emph{Journal of Computational and Applied Mathematics}, 1995.

\bibitem[Von~Neumann and Morgenstern(1944)]{von1944theory}
J.~Von~Neumann and O.~Morgenstern.
\newblock \emph{Theory of games and economic behavior.}
\newblock Princeton University Press, 1944.

\bibitem[Yadav et~al.(2018)Yadav, Shah, Xu, Jacobs, and
  Goldstein]{yadav_stabilizing_2017}
A.~Yadav, S.~Shah, Z.~Xu, D.~Jacobs, and T.~Goldstein.
\newblock Stabilizing adversarial nets with prediction methods.
\newblock In \emph{ICLR}, 2018.

\bibitem[Yaz{\i}c{\i} et~al.(2019)Yaz{\i}c{\i}, Foo, Winkler, Yap, Piliouras,
  and Chandrasekhar]{yazici2018unusual}
Y.~Yaz{\i}c{\i}, C.-S. Foo, S.~Winkler, K.-H. Yap, G.~Piliouras, and
  V.~Chandrasekhar.
\newblock The unusual effectiveness of averaging in {GAN} training.
\newblock In \emph{ICLR}, 2019.
\newblock To appear.

\bibitem[Yousefian et~al.(2014)Yousefian, Nedi{\'c}, and
  Shanbhag]{kannan_optimal_2014}
F.~Yousefian, A.~Nedi{\'c}, and U.~V. Shanbhag.
\newblock Optimal robust smoothing extragradient algorithms for stochastic
  variational inequality problems.
\newblock In \emph{CDC}. IEEE, 2014.

\bibitem[Zhu et~al.(2017)Zhu, Park, Isola, and Efros]{zhu2017unpaired}
J.-Y. Zhu, T.~Park, P.~Isola, and A.~A. Efros.
\newblock Unpaired image-to-image translation using cycle-consistent
  adversarial networks.
\newblock In \emph{ICCV}, 2017.

\end{thebibliography}
